\documentclass[11pt, a4paper, onecolumn, copyright, gdm]{google}

\usepackage[authoryear, sort&compress, round]{natbib}
\usepackage{algorithm} 
\usepackage{algpseudocode} 

\usepackage{hyperref}
\usepackage{float}
\usepackage{enumitem}

\everydisplay{\textstyle}
\usepackage{YK}
\usepackage{xcolor}

\hypersetup{
    colorlinks,
    linkcolor={red!50!black},
    citecolor={blue!50!black},
    urlcolor={blue!80!black}
}
\usepackage[utf8]{inputenc}
\usepackage{graphicx}
\usepackage{tcolorbox}
\tcbuselibrary{listings,breakable,skins}
\usepackage{array}
\usepackage{amsmath}
\usepackage{xltabular}
\usepackage{subcaption}  
\usepackage{wrapfig}
\usepackage{environ} 
\usepackage{booktabs}

\NewEnviron{fmpenumerate}{%
    \color{blue} 
    \begin{enumerate}
        \BODY 
    \end{enumerate}
}

\NewEnviron{vapenumerate}{%
    \color{gray} 
    \begin{enumerate}
        \BODY 
    \end{enumerate}
}

\NewEnviron{iaenumerate}{%
    \color{magenta} 
    \begin{enumerate}
        \BODY 
    \end{enumerate}
}

\usepackage{hyperref}
\usepackage{url}

\paperurl{https://arxiv.org/abs/2603.02029}

\uselogo{} 

\title{Rich Insights from Cheap Signals: Efficient Evaluations via Tensor Factorization}

\correspondingauthor{felipe.polo@mbzuai.ac.ae, isabelaa@google.com.\\ *This project was completed while Felipe Maia Polo was an intern at Google DeepMind, hosted by Isabela Albuquerque.\\Code: \url{https://github.com/google-deepmind/efficient_evals}.}

\reportnumber{} 

\author[1,*]{Felipe Maia Polo}
\author[2]{Aida Nematzadeh}
\author[2]{Virginia Aglietti}
\author[2]{Adam Fisch}
\author[2]{Isabela Albuquerque}

\affil[1]{Institute of Foundation Models, MBZUAI}
\affil[2]{\thepa{}{}}

\begin{abstract}
Moving beyond evaluations that collapse performance across heterogeneous prompts toward fine-grained evaluation at the prompt level, or within relatively homogeneous subsets, is necessary to diagnose generative models' strengths and weaknesses. Such fine-grained evaluations, however, suffer from a data bottleneck: human gold-standard labels are too costly at this scale, while automated ratings are often misaligned with human judgment. To resolve this challenge, we propose a novel statistical model based on tensor factorization that merges cheap autorater data with a limited set of human gold-standard labels. Specifically, our approach uses autorater scores to pretrain latent representations of prompts and generative models, and then aligns those pretrained representations to human preferences using a small calibration set. This sample-efficient methodology is robust to autorater quality, matches or improves on strong baselines in predicting per-prompt human preferences, and provides confidence intervals for key statistical parameters of interest. We also showcase the practical utility of our method by constructing granular leaderboards based on prompt qualities and by estimating model performance solely from autorater scores, eliminating the need for extra human annotations.
\end{abstract}

\begin{document}

\maketitle

\section{Introduction}

The widespread adoption and fast-paced development of generative artificial intelligence (AI) have made robust and insightful evaluation methodologies both more challenging and more relevant. As models become more capable, the limitations of traditional evaluation metrics, which typically aggregate performance into a single average score across a benchmark, are becoming increasingly apparent \citep{rodriguez2021evaluation,zhuangposition}. Unlike these overly coarse metrics, there is increasing interest in fine-grained evaluations, \ie, assessments designed to isolate subtle differences in model performance at the level of individual prompts or narrow, more homogeneous subsets of a benchmark\footnote{In our context, ``prompt'' is equivalent to an instance of a task. For example, in a summarization benchmark, each document forms a distinct prompt, and in a question answering benchmark, each question represents a separate prompt.}. These approaches, often grounded in psychometric frameworks such as Item Response Theory (IRT) \citep{cai2016item, rodriguez2021evaluation, polo2024tinybenchmarks, zhou2025lost, yuan2025benchmarking}, leverage fine-grained, prompt-level data to recover multidimensional skill profiles and identify the specific strengths and weaknesses of a model. Moreover, they enable the construction of prompt-level leaderboards, as well as leaderboards for relatively homogeneous groups of prompts \citep{frick2025prompt}. Such granular insights are crucial not only for model development but also for practical applications, such as dynamic model routing based on prompt difficulty or content \citep{frick2025prompt, song2025irt,somerstep2025carrot}.

The transition to fine-grained evaluation, \ie, at the prompt level or within relatively homogeneous subsets, introduces a significant bottleneck: data scarcity. Establishing reliable estimates of model capability at lower levels requires a vast number of labels. If these labels are to be generated by human annotators, the evaluation process becomes prohibitively expensive and slow. One possibility is to rely on automated rating systems, or ``autoraters'' (\eg, LLM-as-a-Judge), which offer a scalable and cost-effective alternative to derive fine-grained insights. 
Yet, these systems are not perfect: developing autoraters that are accurate across diverse prompts, not merely on average, is expensive and difficult to scale, as it requires large amounts of high-quality human data for development. Moreover, the challenge is amplified in settings involving subjective human preferences, where systematic biases often hinder reliable alignment with human judgments out of the box \citep{zheng2023judging,wang2023large,shi2024judging,dubois2024length, park2024offsetbias, thakur2024judging, ye2024justice, polo2025bridging}. The central challenge, therefore, is to reconcile the reliability of human evaluation with the scalability of automated systems for prompt-level assessment, without incurring the substantial costs of developing autoraters that are highly accurate at the prompt level.

In this work, we move beyond coarse leaderboard metrics to enable fine-grained characterization of generative model performance for individual prompts and relatively homogeneous prompt subsets. To enable fine-grained characterization of model capabilities despite the scarcity of human labels, we propose a novel statistical model rooted in tensor factorization. Our methodology treats autorater scores as auxiliary signals to learn rich representations for both prompts and models, which are subsequently aligned with human preferences using a small calibration set; this approach lets us combine the strengths of individual autoraters to produce accurate performance estimates, even when each autorater is weak on its own. Empirically, we validate our approach on both text-to-image and text-generation benchmarks, including the vision-language dataset Gecko \citep{wiles2024revisiting} and the language model benchmarks BigGen Bench \citep{kim2025biggen} and LMArena \citep{chiang2024chatbot}. With as little as 10\% of human annotations, our method recovers category-specific rankings and statistically significant prompt-level differences between models, and accurately predicts held-out models’ average scores and win-rate differences without observing any of their human labels. These results show that fine-grained, human-aligned evaluation can be performed reliably even under human annotation scarcity, enabling us to identify not only overall model quality but also the specific prompts and cohesive prompt categories where models excel or fall short.

In summary, our contributions span both methodological and empirical dimensions:
\begin{itemize}[leftmargin=*]
    
    \item \textbf{Methodological Framework:} We introduce a tensor factorization framework that unifies plentiful autorater data with sparse human labels to estimate models' performances on individual prompts and a homogeneous subset of prompts. Our approach leverages low-rank structure to handle varying autorater reliability while deriving confidence intervals for these estimates. Our framework gives a statistically grounded foundation for trustworthy evaluation, \ie, human-aligned and ensuring that uncertainty in model rankings is transparent and quantified.

    \item \textbf{Empirical Validation and Practical Applications:} We demonstrate on both text-to-image \citep{wiles2024revisiting} and text-generation benchmarks \citep{kim2025biggen,chiang2024chatbot} that our approach predicts human preferences better than methods that use human labels alone, even when individual autoraters are weak or misaligned. Moving beyond coarse aggregate metrics, our framework practically enables fine-grained leaderboards at the level of individual prompts and cohesive prompt groups that clarify when performance differences are meaningful and reveal where models systematically excel or underperform. 
\end{itemize}

\vspace{-.5cm}
\section{Problem statement}

We address the problem of evaluating a set of generative models $\cM=\{0,\dots,M-1\}$ over a large set of prompts $\cP=\{0,\dots,P-1\}$ in the regime where gold-standard labels, that is, human annotations are scarce, but autorater labels, potentially biased or noisy yet correlated with human judgments, are abundant. By leveraging these autoraters, our goal is to approximate the distribution of human annotations for any evaluation instance $(m,p)$. Here, $m$ denotes either a single model in $\cM$ under single-sided, or pointwise, evaluation, or a pair of models in $\cM$ under side-by-side, or pairwise, evaluation, and $p\in\cP$ denotes a prompt. In single-sided, or pointwise, evaluation, a rater, human or automated, assesses the output of one model in isolation, for example, by assigning a score on a fixed scale such as 0 to 10. In side-by-side, or pairwise, evaluation, the rater compares the outputs of two models and expresses a preference, such as ``Model A is better than Model B'' or ``Tie.'' Estimating this distribution enables us to diagnose fine-grained model strengths and weaknesses, construct model rankings at multiple levels of granularity, from individual prompts to clusters of related prompts, and estimate the performance of previously unseen generative models without requiring additional human supervision.

\section{Methodology}


\subsection{Statistical model}

\textbf{Background.} Let $\cR=\{0, \dots, R-1\}$ denote the complete set of raters, where $r=0$ represents the gold-standard human rater and $r \in\cR\setminus\{0\}$ represents cost-effective autoraters. Each rater $r\in \cR$ is defined by a specific backbone judge (\eg, human, Gemini 2.5 Flash-Lite) and an evaluation template specifying the assessment criteria and scale. We denote the evaluation outcome of rater $r$ for model(s) $m$ on prompt $p$ as $Y_{m,p,r} \in \mathcal{Y}_r$, where $\mathcal{Y}_r=\{0, \dots, C_r-1\}$ is a rater-specific finite set of ordered categories. The interpretation of $Y_{m,p,r}$ depends on the template type: for single-sided evaluations, $m$ is a single model and $Y_{m,p,r}$ represents a quality score; for side-by-side evaluations, $m=(m_0,m_1)$ is a model pair and $Y_{m,p,r}$ indicates the degree of preference for $m_1$ over $m_0$, which can include ties. In practice, we observe a few labels for humans ($r=0$), but many for autoraters ($r>0$), and different raters can evaluate outputs using different templates. To unify human and autorater labels, thus effectively utilizing the auxiliary data, we draw on foundational concepts from Item Response Theory (IRT) \citep{cai2016item} and Bradley-Terry (BT) models \citep{bradley1952rank}, and we propose a statistical model centered on a \textit{tensor of capabilities}.

\textbf{Tensor of capabilities.} We first introduce $\Psi \in \reals^{M\times P \times R}$, which we term the \emph{tensor of capabilities}, where the scalar entry $\Psi_{m,p,r}$ quantifies the capability of model $m$ on prompt $p$ as perceived by rater $r$. We assume that the observed outcome $Y_{m,p,r}$ is governed by a real scalar $\Delta_{m,p,r}$, representing the effective advantage derived from these latent capabilities. The relationship between $\Delta$ and $\Psi$ depends on the evaluation format: for a single-sided scoring template, the effective advantage corresponds directly to the model's capability, such that $\Delta_{m,p,r}=\Psi_{m,p,r}$. Conversely, if rater $r$ utilizes a side-by-side comparison template where $m$ denotes a pair $(m_0,m_1)$, we adopt the standard pairwise assumption that the outcome depends on the difference in capabilities: $\Delta_{m,p,r}=\Psi_{m_1,p,r}-\Psi_{m_0,p,r}$.

Crucially, we do not view the capability $\Psi_{m,p,r}$ as a monolithic attribute. Instead, we operate under the assumption that a model's performance on a specific problem instance is the composite outcome of low-dimensional fundamental abilities and rater criteria, making entries of $\Psi$ interrelated. This motivates a \emph{low-rank} assumption for the resulting tensor $\Psi$. We posit that the interaction between models, prompts, and raters can be factorized into $L$ latent dimensions, interpreted here as specific ``skills'' or ``factors'', such that $\Psi_{m,p,r} = \sum_{l=1}^L \Theta_{m,l} A_{p,l} \Gamma_{r,l}.$

This factorization is known as the CANDECOMP/PARAFAC (CP) tensor decomposition \citep{carroll1970analysis,harshman1970foundations,wang2020learning}. In the CP formulation, $L$ is relatively small, reflecting the finite set of skills required to characterize performance. The learnable factor matrices $\Theta\in \reals^{M\times L}$, $A\in \reals^{P\times L}$, and $\Gamma\in \reals^{R\times L}$ serve as rich representations: $\Theta_{m,l}$ represents model $m$'s proficiency in skill $l$; $A_{p,l}$ represents the demand of prompt $p$ for skill $l$; and $\Gamma_{r,l}$ captures the sensitivity or bias of rater $r$ toward skill $l$. Since the interpretation of the factor matrices can be non-trivial, our primary focus remains on the tensor $\Psi$. The utility of this decomposition lies in the structural constraints it imposes: by implicitly linking entries within $\Psi$, the model facilitates the estimation of the human slice, $\Psi_{m,p,0}$, with high data efficiency. In practice, the rank $L$ serves as a hyperparameter governing model expressiveness and is tuned using a validation set of human annotations.

\textbf{Distribution of $Y_{m,p,r}$.} Due to the ordinal nature of $Y_{m,p,r}$, as previously introduced, we rely on the ordinal logistic (ordered logit) regression framework \citep{wooldridge2010econometric}. Let $\sigma$ be the standard logistic function (\ie, sigmoid) and $\beta^{(r)}_{1}<\cdots<\beta^{(r)}_{C_r-1}$ be learnable ordered real cutoffs. Then, the probability of observing outcome $y \in \{0, \dots, C_r-1\}$ is assumed to be:
\begin{equation}\label{eq:model}
\Pr_\Lambda(Y_{m,p,r}=y) =
\begin{cases}
\sigma\left(\beta^{(r)}_{1} - \Delta_{m,p,r}\right), & \text{if } y = 0, \\
1 - \sigma\left(\beta^{(r)}_{C_r-1} - \Delta_{m,p,r}\right), & \text{if } y = C_r-1, \\
\sigma\left(\beta^{(r)}_{y+1} - \Delta_{m,p,r}\right)
- \sigma\left(\beta^{(r)}_{y} - \Delta_{m,p,r}\right), & \text{otherwise.}
\end{cases}
\end{equation}
where $\Lambda$ denotes the whole set of parameters, \ie, $\Lambda \triangleq (\beta^{(0)},\cdots, \beta^{(R-1)}, \Theta, A, \Gamma)$.

For convenience, we define the set of human parameters ($\beta^{(0)}$ and the first row of $\Gamma$) as $\Lambda^{(h)}$; the parameters that are in $\Lambda$ but not in $\Lambda^{(h)}$ are the autorater, model, and prompt parameters and denoted as $\Lambda^{(a)}$, \ie, $\Lambda=(\Lambda^{(h)},\Lambda^{(a)})$. This formulation is related to models routinely used in the literature. For example, when $P=R=C_r-1=1$ and we carry out side-by-side evaluations, we recover the BT model (with an order effect $\beta^{(0)}_1$, which is zero in the classic model) \citep{bradley1952rank}. When  $R=1$ and evaluations are single-sided, we recover an instance of the Graded Response Model (GRM) from the IRT literature with thresholds shared across items \citep{Reckase2009,samejima2016graded}.


\subsection{Model fitting}

We assume access to a dataset $\cD=\cup_{n=1}^N\{D_n\}$, where each $D_n \triangleq (m^{(n)},p^{(n)},r^{(n)},Y^{(n)})$ is a tuple containing the model(s) $m^{(n)}$, prompt $p^{(n)}$, rater $r^{(n)}$, and the observed label $Y^{(n)} \triangleq Y_{m^{(n)},p^{(n)},r^{(n)}}$. We partition this dataset into human-rated samples, $\cD^{(h)}\triangleq \cup_{n: r^{(n)}=0}\{D_n\}$, and autorater samples, $\cD^{(a)}\triangleq \cup_{n: r^{(n)}>0}\{D_n\}$. For convenience, we learn $\Lambda$ using both $\cD^{(h)}$ and $\cD^{(a)}$ using a two-stage maximum likelihood estimation (MLE) procedure; as we comment in the next paragraphs, this approach lets us identify human parameters by fixing model and prompt representations, and apply standard techniques to build confidence intervals for quantities of interest.

The negative log-likelihood (NLL) for a single data point $D_n$ under parameters $\Lambda$ is defined as:
\begin{equation*}
    \cL(\Lambda,D_n) \triangleq-\sum_{y=0}^{C_{r^{(n)}}-1} \ones[Y^{(n)}=y] \log \Pr_\Lambda(Y_{m^{(n)},p^{(n)},r^{(n)}}=y).
\end{equation*}
In the first stage, we estimate $\Lambda^{(a)}$, including the model embeddings, prompt embeddings, and autorater parameters, by minimizing the NLL over $\cD^{(a)}$:
\[
\widehat{\Lambda}^{(a)} \in \argmin_{\Lambda^{(a)}} \sum_{D_n \in \cD^{(a)}} \cL(\Lambda,D_n).
\]
In the second stage, we freeze the estimates $\widehat{\Lambda}^{(a)}$  and fit the human-specific parameters $\Lambda^{(h)}$ (the human rater embedding and cutoffs) by minimizing the NLL on the human-rated dataset $\cD^{(h)}$:
\[
\widehat{\Lambda}^{(h)} = \argmin_{\Lambda^{(h)}} \sum_{D_n \in \cD^{(h)}} \cL\big((\Lambda^{(h)},\widehat{\Lambda}^{(a)}),D_n\big).
\]

In practice, we perform the first stage optimization using the Adam optimizer \citep{kingma2014adam} (Algorithm \ref{alg:stage1}). We run this stage with multiple random initializations and learning rates, selecting the configuration that yields the lowest training loss. The second stage is conducted using the BFGS algorithm \citep{nocedal2006numerical}.

An important observation is that the estimator $\widehat{\Lambda}^{(a)}$ is not unique\footnote{Therefore, we use ``$\in$'' instead of ``$=$''.} because the parameterization of $\Lambda^{(a)}$ is not identifiable. However, conditioned on a fixed $\widehat{\Lambda}^{(a)}$, the estimator $\widehat{\Lambda}^{(h)}$ is typically unique, as the optimization reduces to standard maximum likelihood estimation for ordinal logistic regression (Appendix \ref{sec:unique}). While identifiability of the full set of parameters can be achieved by constraining\footnote{Please check Appendix \ref{sec:ident} for a concrete set of constraints.} the parameter space during the first stage, enforcing it offers no practical benefit, as we are interested in the predictive performance of the capability tensor rather than in the interpretation of the specific latent factors. Pragmatically, we interpret the first stage as a representation learning phase, and the second stage as fitting a conventional ordinal regression model on top of those learned representations. In Appendix \ref{sec:fit}, we connect our approach to transfer learning and comment on how to fine-tune models/prompt representations to boost our approach's performance; this optional third stage, ``Ours FT'' in Figure \ref{fig:varying_B}, fine-tunes all parameters on the human labels, which improves predictions when prompts have several human labels but invalidates the confidence intervals below.

\subsection{Fine-grained AI evaluation}\label{sec:fine-grained}


In this subsection, we demonstrate how the fitted model can be used to construct leaderboards based on a single prompt or a homogeneous subset of prompts. We treat the first fitting stage as a representation learning phase; accordingly, all model parameters (except those associated with humans) are held fixed in the second stage and used to construct the ordinal regression features. Conceptually, the first stage learns a shared representation, while the second stage focuses on statistical inference and uncertainty quantification.


\textbf{Prompt-specific evaluation.} We describe how to estimate model capabilities with uncertainty quantification for prompt-specific leaderboards. Denote by $\gamma_r$ the transpose of the $r$-th row of $\Gamma$ (rater representations) and by $v_{m,p}$ the transpose of the element-wise product of the $m$-th row of $\Theta$ (model representations) and the $p$-th row of $A$ (prompt representations). Now, to rank models according to a prompt $p$ and the human rater ($r=0$), we use an estimate for the capability scalar $\Psi_{m,p,0} = v_{m,p}^\top \gamma_{0}$, swapping $\gamma_{0}$ for $\hat{\gamma}_{0}$ (obtaining $\hat{\Psi}_{m,p,0}$), where $\hat{\gamma}_{0}$ is obtained in the second stage of the fitting algorithm, using $|\cD^{(h)}|$ human labels. We derive the confidence interval for $\Psi_{m,p,0}$ as 

\begin{equation*}
    \text{CI}_\lambda =\hat{\Psi}_{m,p,0}\pm \Phi^{-1}\left(\frac{1+\lambda}{2}\right)\sqrt{\frac{v_{m,p}^\top\hat{\Sigma} v_{m,p}}{|\cD^{(h)}|}},
\end{equation*}

where $\hat{\Sigma}$ is a matrix computed from the data, $\lambda\in(0,1)$ is the target confidence level and $\Phi^{-1}$ the standard Gaussian\footnote{The use of Gaussian quantiles is motivated in Appendix \ref{sec:ci}.} quantile function. We report these marginal intervals throughout the main text, mainly for simplicity. They treat the first-stage representations as fixed and are therefore valid conditionally on them. In Appendix \ref{sec:ci}, we introduce a parametric bootstrap that also propagates first-stage uncertainty, extend both approaches to simultaneous coverage, and study in simulation when each is accurate.

\textbf{Category-specific evaluation.}
To better understand model proficiency in specific skills, it is often more insightful to aggregate capability across a subset of related prompts, denoted by ${\mathcal{P}_s}$, rather than relying on individual rankings. Since individual prompts often capture secondary, unrelated skills alongside the target capability, aggregating allows us to isolate the dominant common skill. To achieve this, we leverage the concept of the \textit{reference composite} \citep{Reckase2009}. Let $\alpha_{p}$ denote the transpose of the $p$-th row of $A$. The reference composite for the set $\{\alpha_{p}\}_{p\in{\mathcal{P}_s}}$ is defined as the unit vector $\alpha_{{\mathcal{P}_s}}$ that best summarizes the prompts in ${\mathcal{P}_s}$ by solving 
\[
\alpha_{{\mathcal{P}_s}} = \operatorname*{arg\,min}_{u: \| u \|=1} \sum_{p\in {\mathcal{P}_s}} \left\| \alpha_{p}-(u^\top \alpha_{p})u \right\|^2.
\]
In words, $\alpha_{{\mathcal{P}_s}}$ represents the direction of the line that best approximates the set $\{\alpha_{p}\}_{p\in{\mathcal{P}_s}}$. By minimizing the orthogonal projection errors, $\alpha_{{\mathcal{P}_s}}$ aligns with the principal direction of the category, effectively averaging out the noise introduced by secondary skills. Operationally, $\alpha_{{\mathcal{P}_s}}$ corresponds to the leading eigenvector of the matrix $\sum_{p\in {\mathcal{P}_s}} \alpha_{p}\alpha_{p}^\top$. This observation allows us to quantify how well the reference composite captures the information in ${\mathcal{P}_s}$ by calculating the ratio of the first eigenvalue to the sum of all eigenvalues; this metric provides a notion of \emph{cohesion}, indicating the extent to which the set ${\mathcal{P}_s}$ is driven by a single dominant skill. We define $v_{m,{\mathcal{P}_s}}$ as the entry-wise product of the (transposed) $m$-th row of $\Theta$ and $\alpha_{{\mathcal{P}_s}}$, while the construction of $\Psi_{m,{\cP_s},0}$ (or $\hat{\Psi}_{m,{\cP_s},0}$) proceeds as before by substituting $v_{m,p}$ with $v_{m,{\mathcal{P}_s}}$.

\textbf{Comparing two generative models.} 
In certain scenarios, \eg, when assessing the relative strengths of one model versus another, it is more informative to directly compare two models of interest, $m=(m_0,m_1)$, rather than constructing a leaderboard containing all models. In this context, our attention shifts to the quantity $\Delta_{m,p,0} = \Psi_{m_1,p,0} - \Psi_{m_0,p,0} = (v_{m_1,p} - v_{m_0,p})^\top \gamma_{0}.$ Analogous to $\Psi_{m,p,0}$, the estimator $\hat{\Delta}_{m,p,0}$ is obtained by substituting $\gamma_{0}$ with $\hat{\gamma}_{0}$. Similarly, confidence intervals are derived by replacing the estimate $\hat{\Psi}_{m,p,0}$ with $\hat{\Delta}_{m,p,0}$ and the vector $v_{m,p}$ with the difference vector $v_{m_1,p} - v_{m_0,p}$.

\textbf{Comparing results across prompts $p$.} A critical observation is that when using side-by-side templates for human annotation, only the differences $\Psi_{m_1,p,0} - \Psi_{m_0,p,0}$ are identifiable (see Appendix \ref{sec:ident}). Consequently, the values of $\Psi_{m,p,0}$ are not comparable across different values of $p$. In contrast, such comparisons are entirely valid when the human evaluation template is single-sided. Moreover, when different sets ${\cP_s}$ are considered, it is hard to compare results across groups of prompts (in all cases) since $\alpha_{\cP_s}$'s are all normalized. 

\section{Experiments}

In this section, we apply our method to real-world data. Specifically, we: (i) compare the predictive power of our approach against baselines; (ii) demonstrate that insightful category-specific rankings can be derived using only a fraction of human annotations; and (iii) analyze the relative strengths and weaknesses of models across benchmarks. In Appendix \ref{sec:extra_exp}, we include extra experiments with real and synthetic data that (iv) predict the performance of models that received no human annotations; (v) provide evidence that prompt representations can be leveraged to extract useful insights about benchmarks; and (vi) provide extra insights on how our method performs under correlated and biased autoraters. For all experiments and benchmarks, we set $L=10$ as this number produces reasonable performance across benchmarks (Figure \ref{fig:varying_rank}; Figure \ref{fig:first_stage_losses} reports the corresponding first-stage losses). All experiments were run on a CPU\footnote{Memory peaked at 6GB while the running time for a single job, \ie, running the first and second stages once, would take a few hours on a laptop.} cluster.

\subsection{Data}

We test our methodology on the following three established benchmarks for generative models: the first focuses on image generation, while the latter two target the text modality. For all benchmarks, we utilize the model generations and human annotations provided by their authors.

\begin{itemize}[leftmargin=*]
    \item \textbf{Gecko2K (Gecko)} \citep{wiles2024revisiting} is a comprehensive evaluation suite for text-to-image (T2I) alignment. We utilize the Gecko(S) subset, which comprises approximately 1,000 prompts selected via a hierarchical, semi-automatic method to ensure diverse coverage of fine-grained skills. The diagnostic design of Gecko(S) provides a controlled and objective measure of model performance against specific generation challenges. We rely on the provided side-by-side human annotations, utilizing approximately 18k pairwise human annotations involving four different T2I models.
    \item \textbf{BigGen Bench (BGB)} \citep{kim2025biggen} evaluates language model outputs based on detailed rubrics across five satisfaction levels. We map the original $1$--$5$ scale to a $0$--$4$ range and focus on the subset containing English-language human annotations. After excluding a small number of invalid entries, this results in 695 instances across 77 tasks and nine evaluated capabilities (\eg, planning, tool usage). Each instance includes responses from four different models, totaling 2,780 human-annotated data points.
    \item \textbf{LMArena:} We use a dataset\footnote{Retrieved from \href{https://huggingface.co/datasets/lmarena-ai/arena-human-preference-140k}{https://huggingface.co/datasets/lmarena-ai/arena-human-preference-140k}.} derived from LMArena (also known as Chatbot Arena) \citep{chiang2024chatbot}. The original dataset consists of 140k queries with side-by-side responses from anonymous models. User preferences are annotated as either a clear choice or a tie (labeled as ``good'' or ``bad'' ties); we treat both tie types equivalently. The dataset we use is obtained by filtering matches with one round where both competing models are among ten pre-specified state-of-the-art LLMs\footnote{The chosen models are: claude-3-5-haiku-20241022, claude-3-5-sonnet-20241022, gpt-4o-mini-2024-07-18, llama-3.3-70b-instruct, amazon.nova-pro-v1:0, gemini-2.5-flash-lite-preview-06-17-thinking, gemini-2.5-pro, gpt-4.1-2025-04-14, claude-opus-4-20250514, and claude-sonnet-4-20250514.}, resulting in nearly 5k human-annotated matches. 
\end{itemize}

\textbf{Autorater labels.} We adopted different strategies for collecting autorater labels across these three benchmarks. For BGB, we aggregated pre-existing ratings provided by \citet{kim2025biggen} for five models\footnote{The columns in their data are given by the names: claude\_score, gpt4\_04\_turbo\_score, gpt4\_score, prometheus\_8x7b\_score, prometheus\_8x7b\_bgb\_score.} and by \citet{polo2025bridging} for seven models\footnote{Model names are: atlaAI/Selene-1-Mini-Llama-3.1-8B, meta-llama/Llama-3.1-8B-Instruct, gpt-4-turbo, gpt-4.1-nano, gpt-4.1, gpt-4o-mini-2024-07-18, prometheus-eval/prometheus-8x7b-v2.0.}, including distinct variations; for each one of the open models, the authors applied two distinct prompting strategies. This resulted in a total of 15 autoraters for BGB. For both Gecko and LMArena, we developed custom autoraters based on Gemini 2.5 Flash-Lite\footnote{We rely on a lightweight model to enable faster iteration and large-scale annotation collection, reaching hundreds of thousands of scores. While upgrading to more capable models would likely improve baseline performance, it would also enhance the performance of our method, since it depends directly on autorater labels. In summary, we expect our empirical findings to extend to stronger models, as our experiments already include GPT-4.1 in the BGB setting.} \citep{comanici2025gemini}. To induce diverse evaluative behaviors, we varied both the evaluation template (single-sided vs. side-by-side) and the specific persona or criteria used. For autoraters using the single-sided template, we employed a two-step process: the model first generates a checklist of important factors required for each prompt, and second, it evaluates each item on the checklist as present (1) or absent (0); we use the same checklists for different generations for a given prompt. We calculate the proportion of present factors and bin this fraction into 10 ordered categories. We employed 8 autoraters for Gecko and 24 for LMArena, with an even split between template types. We set the autorater temperature to 1, sampled 8 ratings per input, and utilized all replicas when fitting the statistical model. In simulation, judges that share a systematic bias still recover human preferences as long as the panel is diverse; recovery degrades only when the judges are near-copies of one another and misaligned with humans (Appendix \ref{sec:shared_bias}).

In Appendix \ref{sec:auto}, we provide examples of the prompts used and describe the associated personas.

\subsection{Predictive power}\label{sec:predictive}

In this subsection, we assess the goodness of fit of our method relative to baseline approaches. These results serve primarily as a sanity check, demonstrating that for different human annotation budgets: (i) prompt-specific modeling is essential; and (ii) auxiliary autorater data is highly beneficial for fitting our statistical model when human annotations are scarce.

The first baseline method, denoted as ``Constant'', represents a simplification of our approach where: (i) the rank is set to $L=1$; (ii) all entries of $A$ are fixed to the same constant; and (iii) $\Gamma$ consists of a single row representing the human rater. Without loss of generality, the entries of $A$ and $\Gamma$ are fixed to $1$. Consequently, the ``Constant'' baseline disregards prompt heterogeneity and ignores auxiliary data from autoraters. Note that for side-by-side evaluations (where $C_0=2$), this baseline reduces to the classic Bradley-Terry model \citep{bradley1952rank}. The second baseline, ``Prompt-specific'', is a variant of our main method that excludes auxiliary data but allows for prompt heterogeneity; this approach can be seen as a variation of the Item Response Theory model \citep{Reckase2009}. Here, we let $L>1$ and assign each prompt a distinct, trainable representation in $A$, while $\Gamma$ remains restricted to a single row fixed to a vector of ones. The final baseline is the Prompt-to-Leaderboard (P2L) method \citep{frick2025prompt}, applied only to LMArena. In short, P2L parameterizes $A$ using embeddings extracted from a language model fine-tuned to predict the demands of generic task descriptions. We initialize $A$ using embeddings from the model \texttt{lmarena-ai/p2l-3b-rk-01132025} and train the remaining model parameters from scratch using human annotations (except for $\Gamma$, which is again fixed since no autoraters are considered). 

\begin{figure}[t]
    \centering
    \begin{subfigure}[b]{0.285\textwidth}
        \centering
        \includegraphics[width=1.065\linewidth]{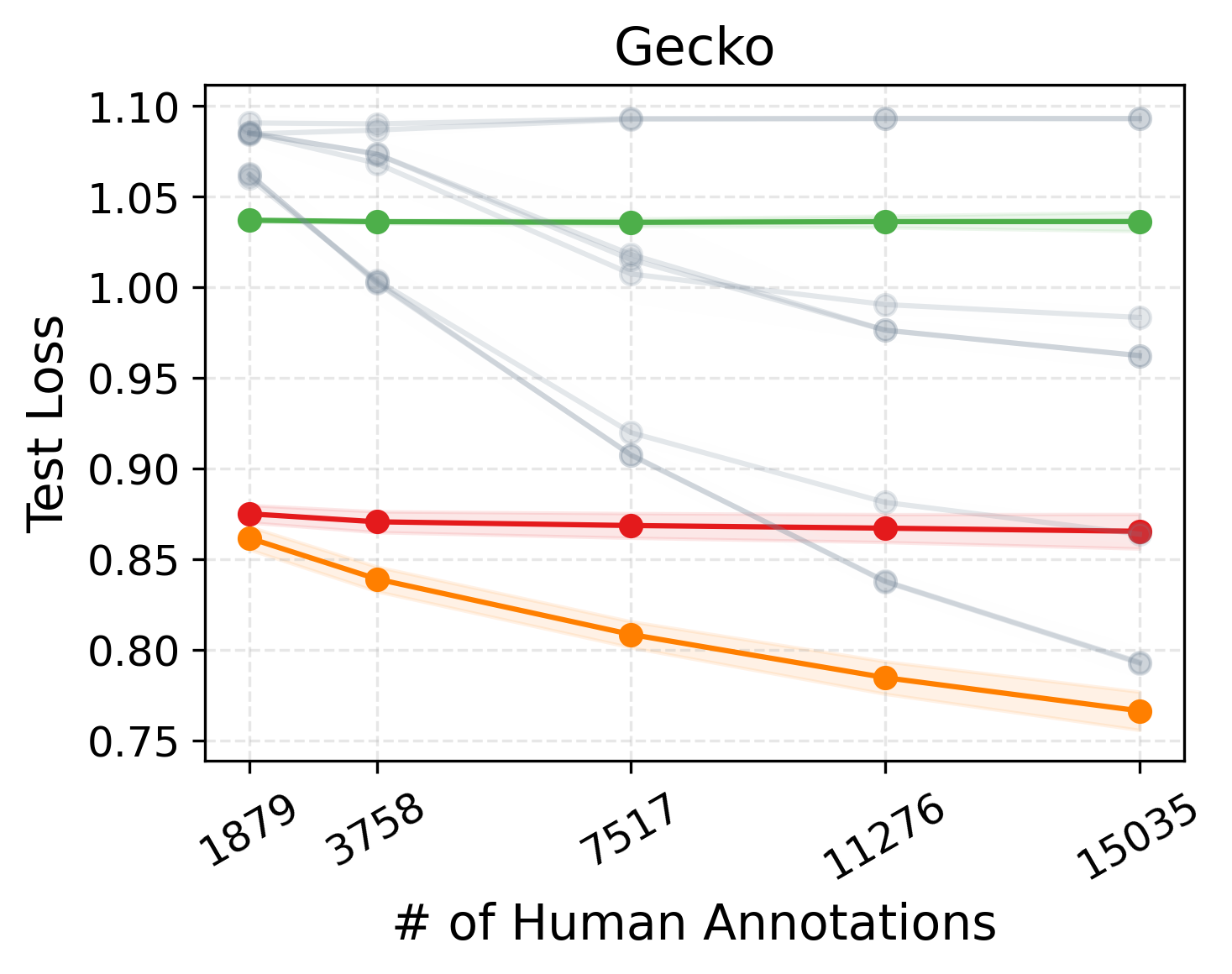}
    \end{subfigure}
    \begin{subfigure}[b]{0.285\textwidth}
        \centering
        \includegraphics[width=1.065\linewidth]{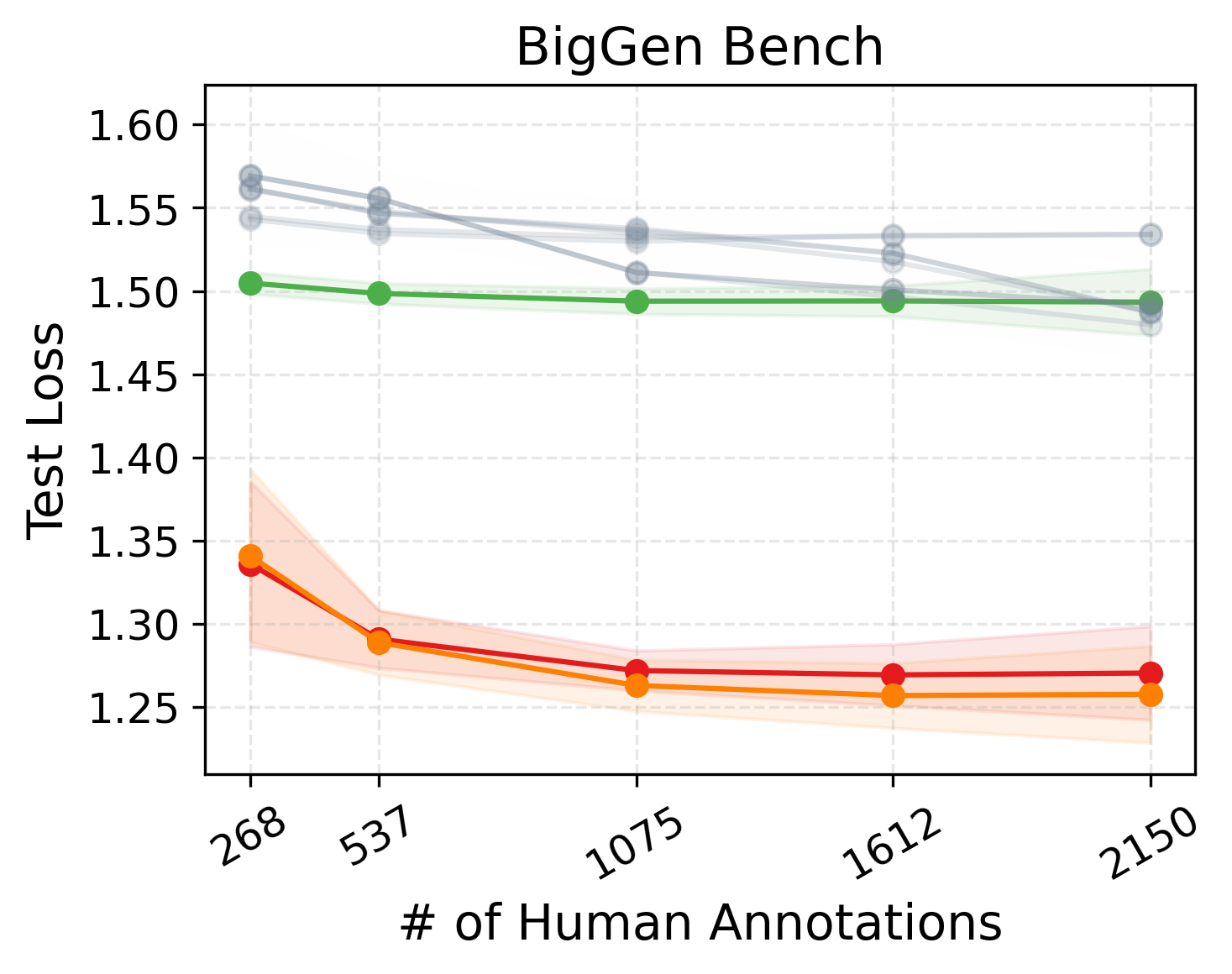}
    \end{subfigure}
    \begin{subfigure}[b]{0.37\textwidth}
        \centering
        \includegraphics[width=1.075\linewidth]{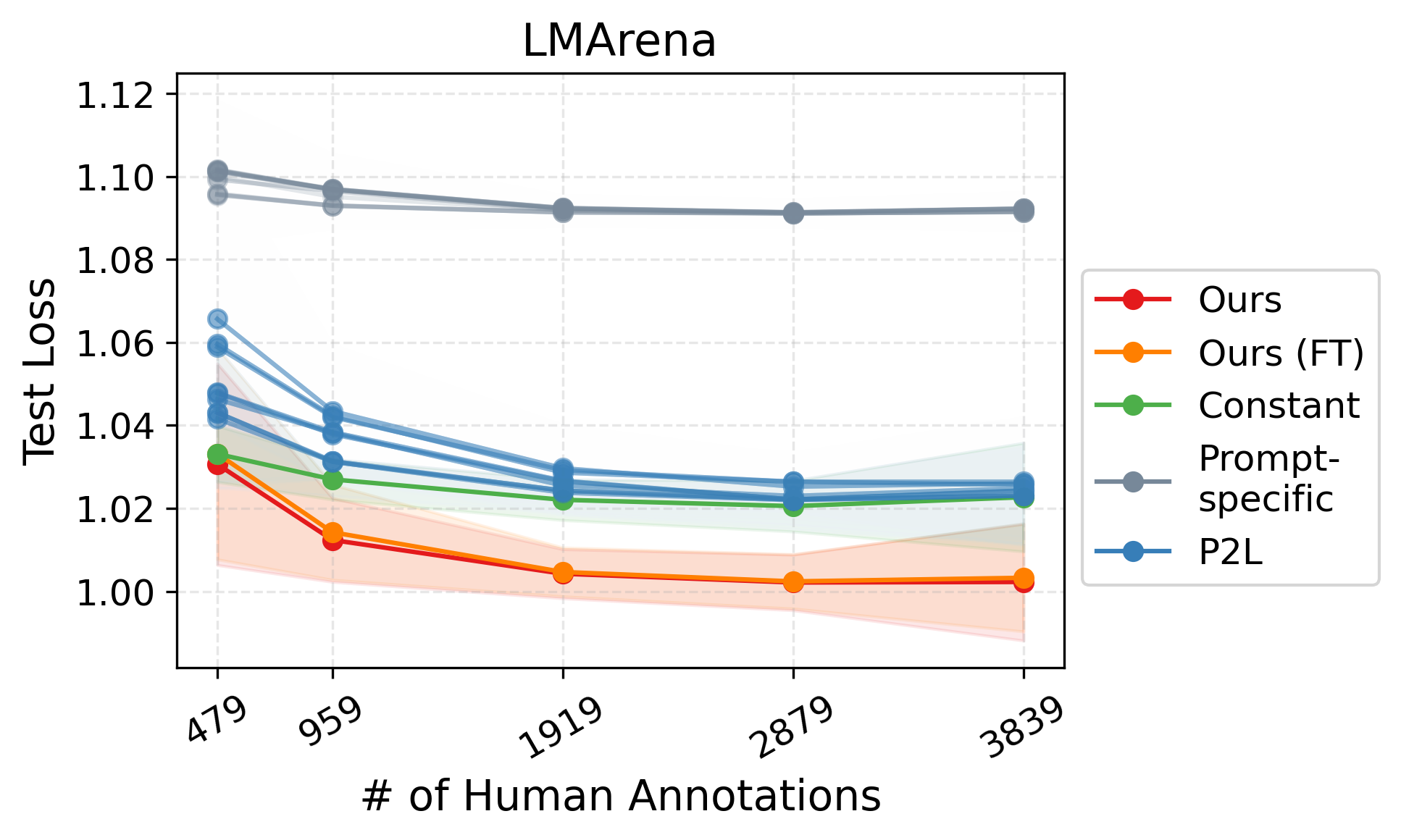}
    \end{subfigure}
    
    \caption{\small Test cross-entropy loss comparison between our proposed method (default and fine-tuned) and baseline approaches (Constant, Prompt-specific, P2L) across three benchmarks: Gecko (left), BigGen Bench (center), and LMArena (right). Our methods consistently achieve lower losses for different human annotation budgets, demonstrating the benefits of prompt-specific modeling and auxiliary autorater data. Gray curves: Prompt-specific baseline over the swept ranks $L$ and $\ell_2$ strengths.}

\label{fig:varying_B}
\end{figure}

\begin{figure*}[t]
    \centering
    \begin{subfigure}[b]{0.49\textwidth}
        \centering
        \includegraphics[width=1.095\linewidth]{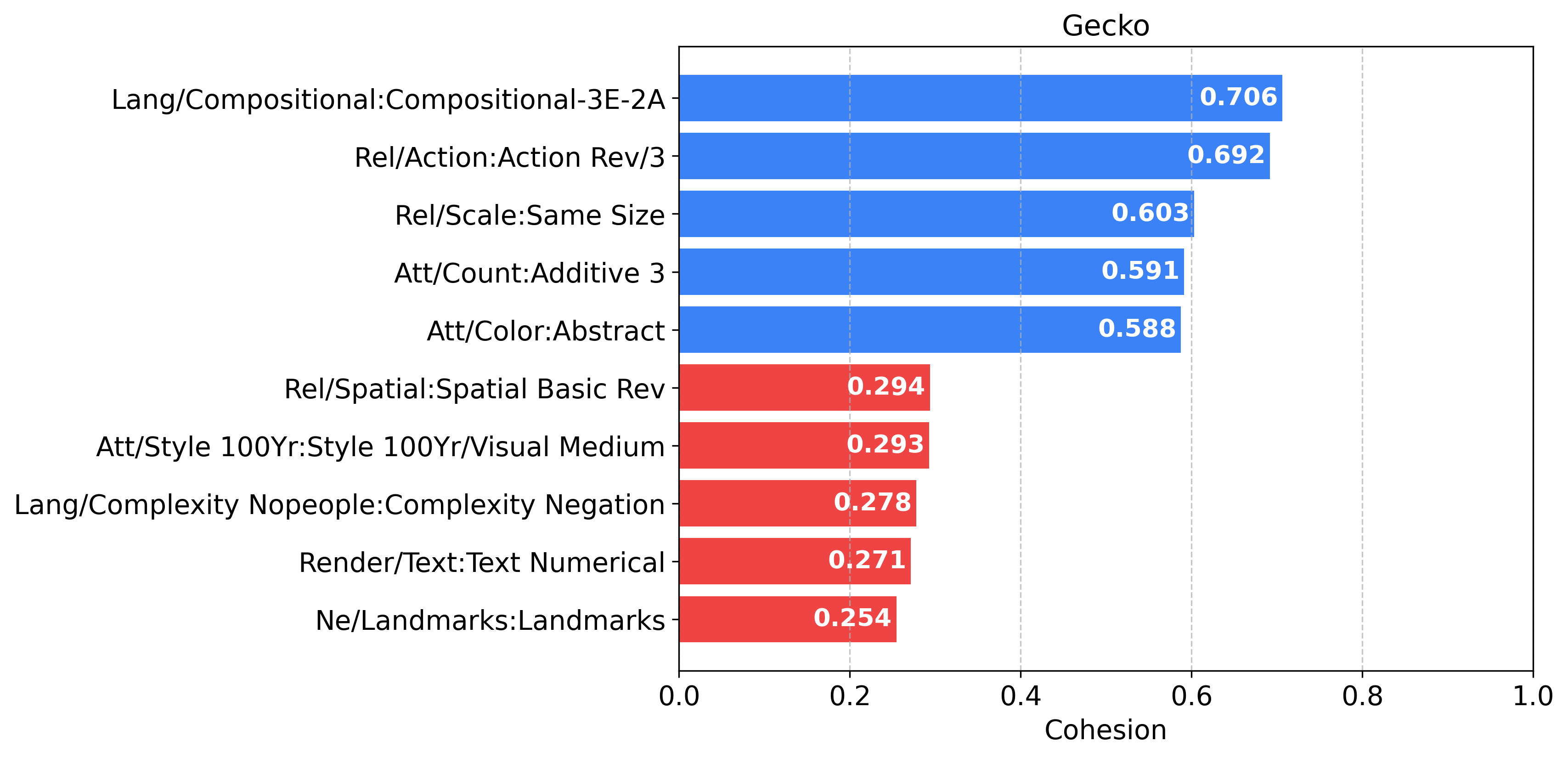}
    \end{subfigure}
    \begin{subfigure}[b]{0.49\textwidth}
        \centering
        \includegraphics[width=.89\linewidth]{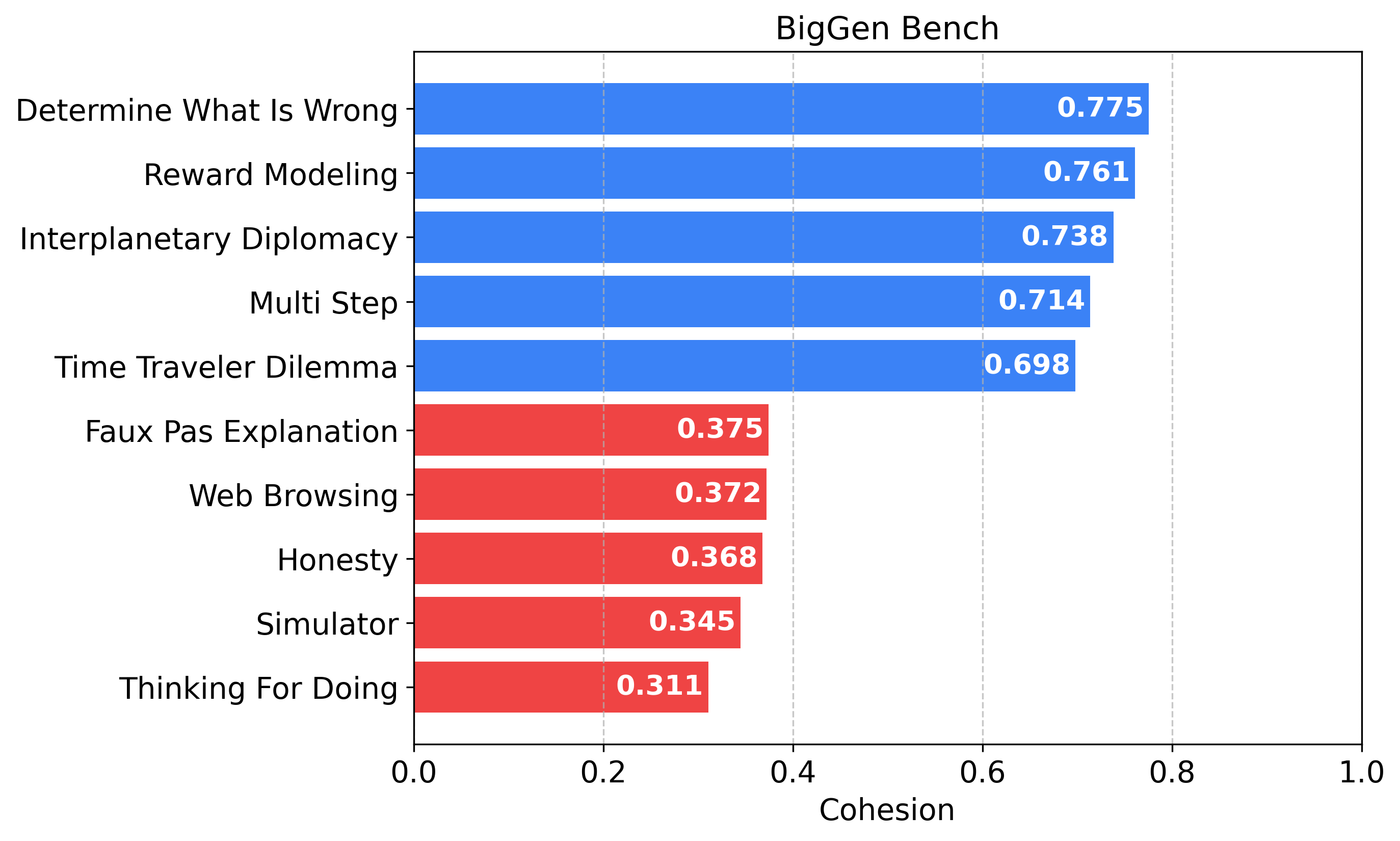}
    \end{subfigure}  
    \caption{\small Category cohesion rankings for Gecko (left) and BigGen Bench (right). Categories are ordered from most to least cohesive based on the metric described in \S \ref{sec:fine-grained}.}
\label{fig:cohesion}
\vspace{-.7cm}
\end{figure*}
Figure \ref{fig:varying_B} illustrates the test cross-entropy loss for both the default and fine-tuned versions of our approach compared to all baselines. For each budget, we randomly allocate a portion of human annotations for training and the remainder for testing. This process is repeated 30 times for each benchmark, and we report the average results $\pm$ one standard deviation. For the ``Prompt-specific'' baseline, we sweep over a wide range of ranks $L$ and $\ell_2$ regularization strengths, whereas for P2L we vary the $\ell_2$ regularization strength and learning rates. Overall, our methods outperform the baselines that use human labels alone across all benchmarks. Thus, the low-rank structure that enables the leaderboards below comes at no cost in predictive accuracy. Appendix baselines on individual autoraters (Figure \ref{fig:varying_B_autoraters}) and a sweep over the fraction of autoraters used (Figure \ref{fig:varying_autorater_fraction}) complete this comparison. Notably, both our fine-tuned model and the ``Prompt-specific'' baseline perform well on Gecko given sufficient human labels. We attribute this to the multiple annotations available per prompt in Gecko, which allow for better fitting of prompt representations compared to the sparse annotations found in other benchmarks.

\subsection{Category-specific rankings}

In this subsection, we focus on Gecko and BGB, as these benchmarks provide a pre-specified set of fine-grained prompt categories. We first report the cohesion of these categories, as measured by the metric described in \S \ref{sec:fine-grained}. Figure \ref{fig:cohesion} shows the ranking of categories from each benchmark from most to least cohesive; permutation tests in Appendix \ref{sec:cohesion} indicate that the cohesion of the most cohesive categories is not due to chance. From the most cohesive categories, we estimated the human score, $\Psi_{m,\cP_s,0}$, for two categories per benchmark. 

\begin{wrapfigure}{r}{0.5\textwidth}

    \centering

    \begin{subfigure}[b]{0.235\textwidth}
        \centering
        \includegraphics[width=1.025\linewidth]{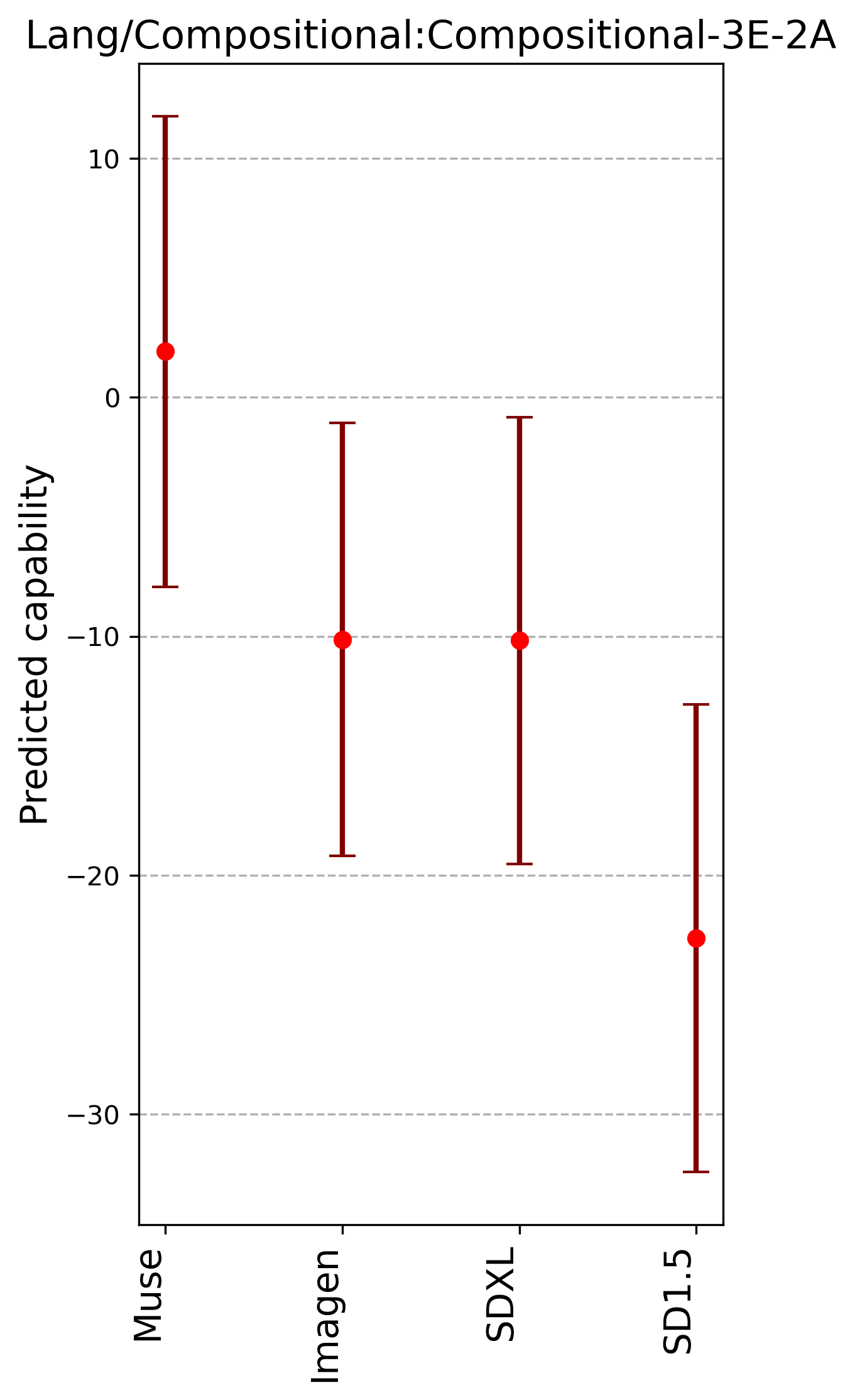}
    \end{subfigure}
    \begin{subfigure}[b]{0.23\textwidth}
        \centering
        \includegraphics[width=.9\linewidth]{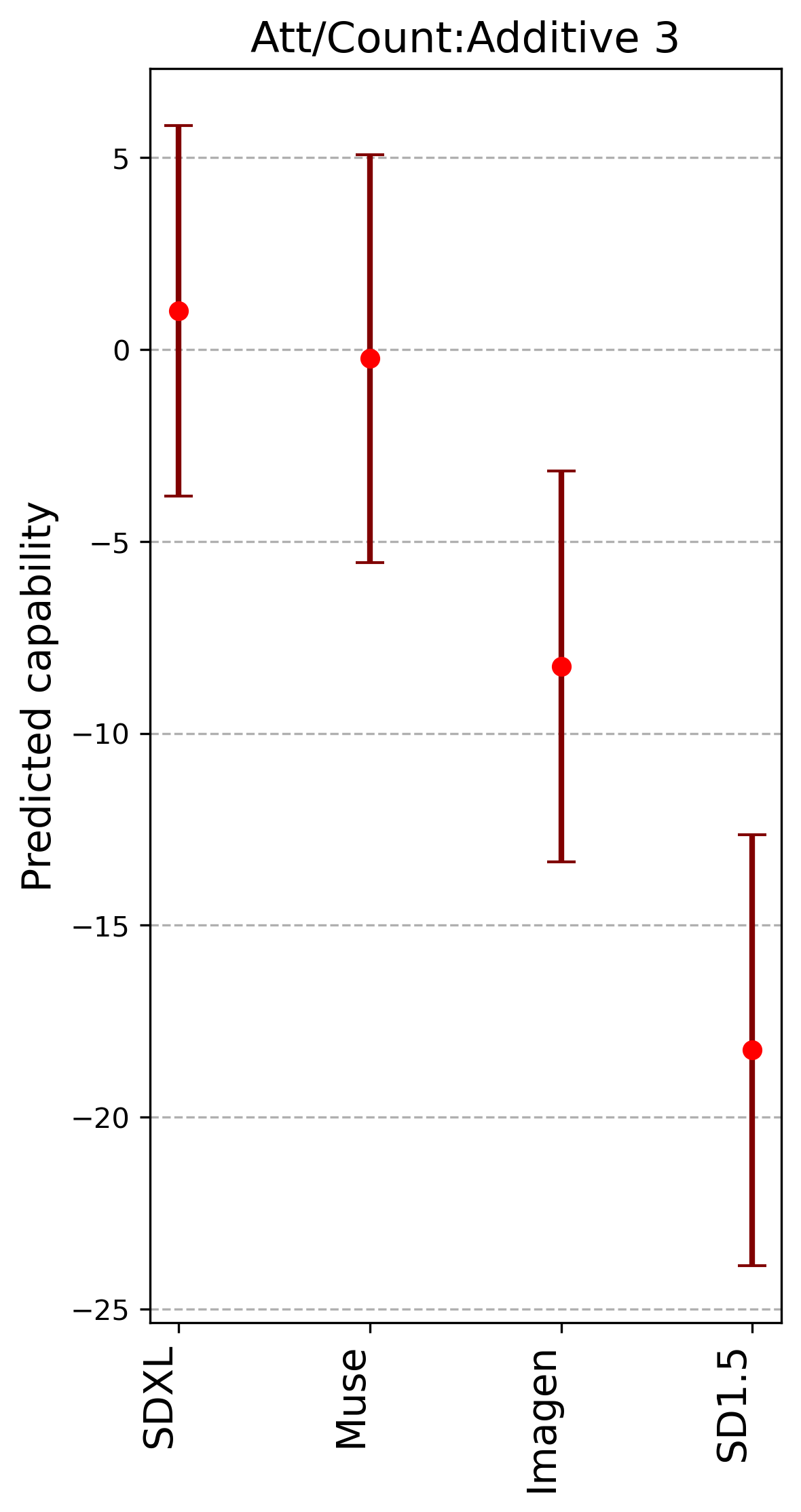}
    \end{subfigure}
    \caption{\small Model rankings with $95\%$ confidence intervals for two Gecko categories: ``Lang/Compositional'' (left) and ``Additive'' (right), estimated using only $10\%$ of human annotations. The plots reveal performance discrepancies across different skills; for instance, Imagen and SDXL are statistically indistinguishable in compositional tasks but Imagen underperforms in additive tasks.}
    \vspace{-.5cm}
\label{fig:ranking_cis_10_gecko}
\end{wrapfigure}
\begin{wrapfigure}{r}{0.5\textwidth}
    \centering

    \begin{subfigure}[b]{0.23\textwidth}
        \centering
        \includegraphics[width=.9\linewidth]{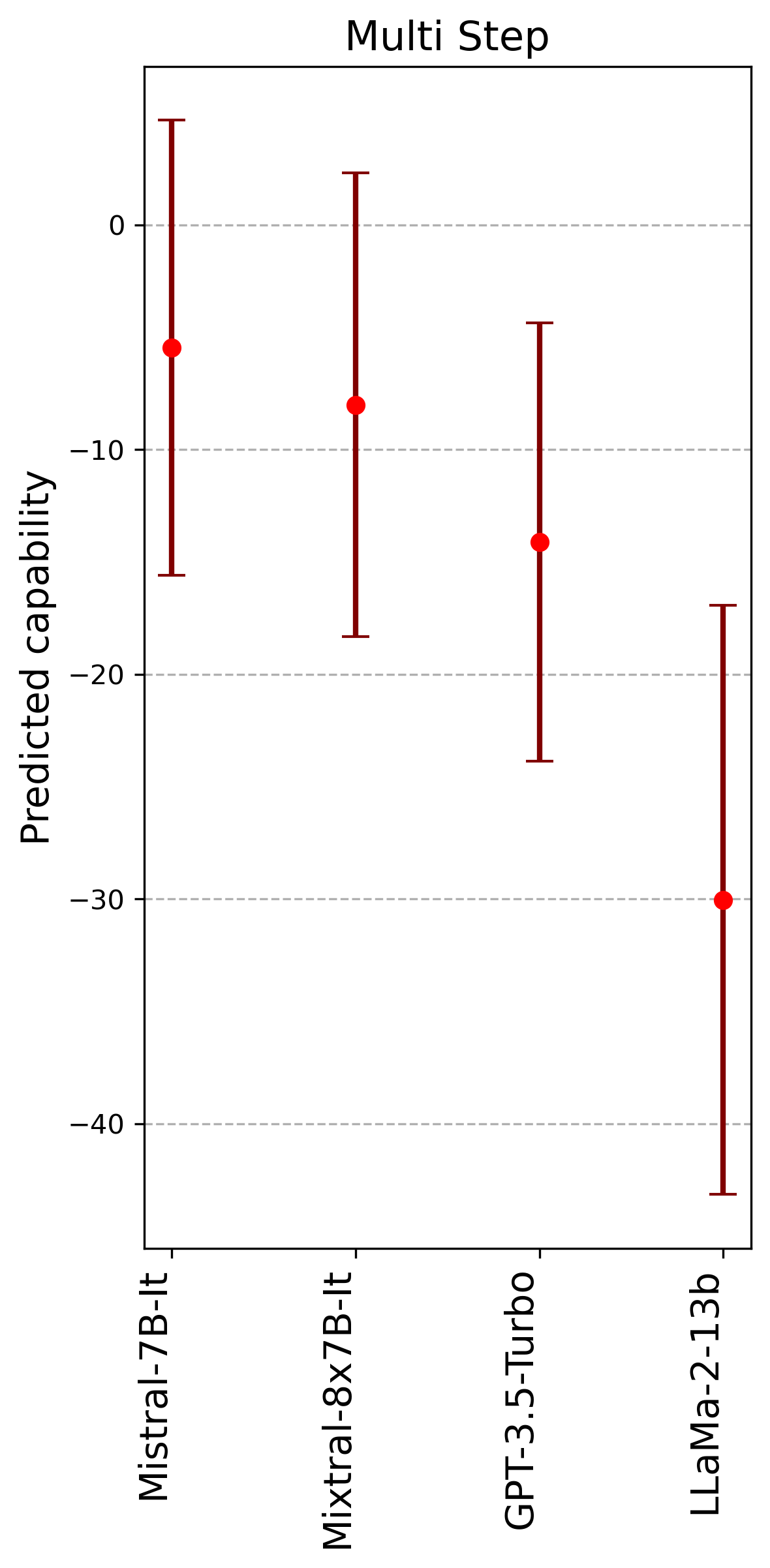}
    \end{subfigure}
    \begin{subfigure}[b]{0.23\textwidth}
        \centering
        \includegraphics[width=.9\linewidth]{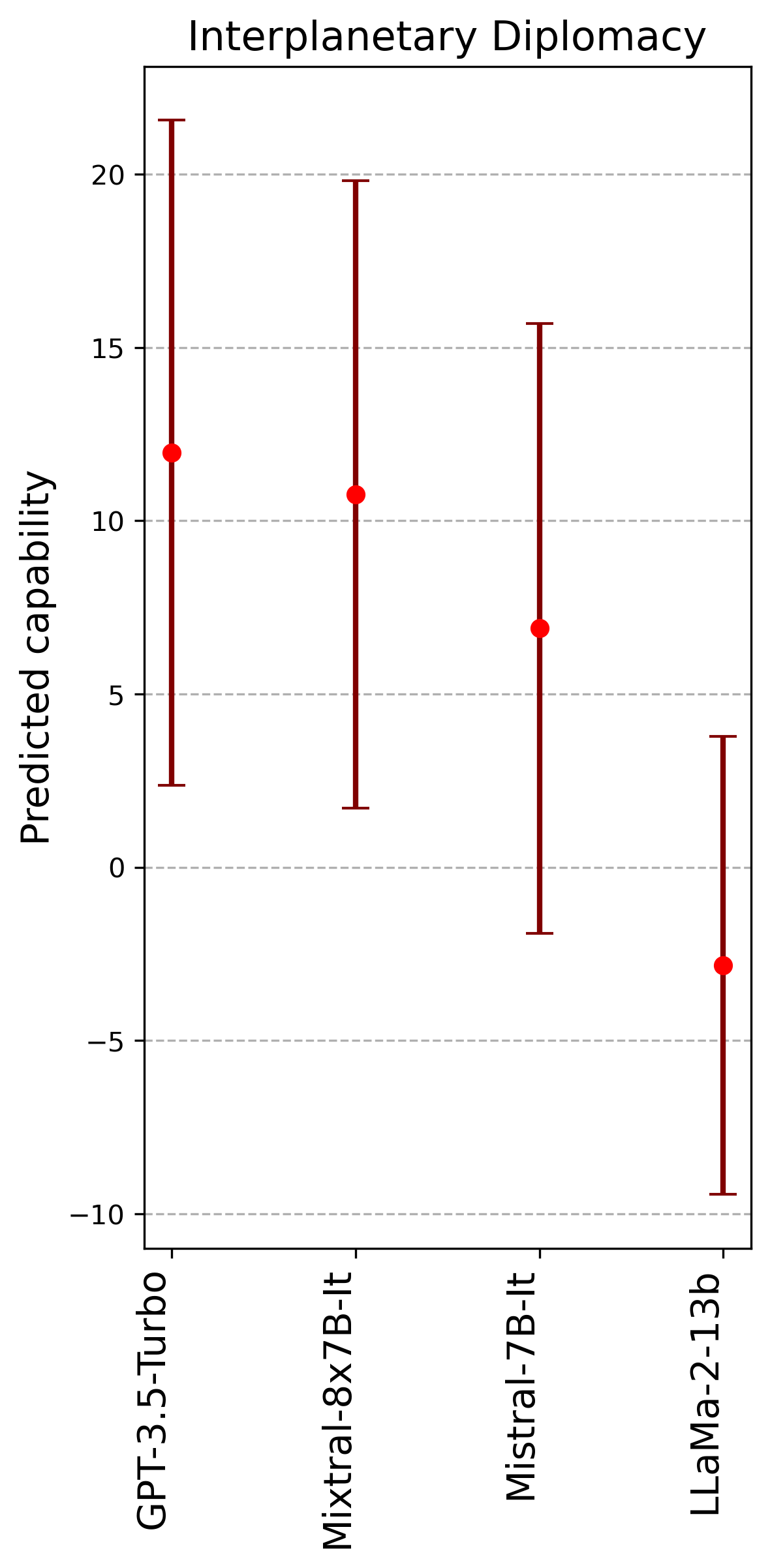}
    \end{subfigure}
    \caption{\small Model rankings with $95\%$ confidence intervals for two BigGen Bench tasks: ``Multi-step'' (left) and ``Interplanetary Diplomacy'' (right), estimated using a $10\%$ sample of human annotations.}
\label{fig:ranking_cis_10_bgb}
\vspace{-.4cm}
\end{wrapfigure}

We used only $10\%$ of the total human annotations and $95\%$ confidence intervals. As shown in Figures \ref{fig:ranking_cis_10_gecko} and \ref{fig:ranking_cis_10_bgb}, using only 10\% of the annotations\footnote{This is equivalent to $<2$ (resp. $<1/2$) human labels per prompt on Gecko (resp. BGB), on average.} across all prompts suffices to reliably distinguish model performance across individual categories. We further see that models may excel or underperform depending on the target category. For instance, the data do not distinguish Imagen from SDXL on the ``Lang/Compositional'' category, but Imagen is significantly worse at generating for prompts for the ``Additive'' category.

In the appendix, Figures \ref{fig:ranking_cis_10_gecko_full} and \ref{fig:ranking_cis_10_bgb_full} expand the results of this section for all top cohesive categories in Figure \ref{fig:cohesion}. Moreover, Figures \ref{fig:ranking_cis_100_gecko} and \ref{fig:ranking_cis_100_bgb} demonstrate that increasing the sample size improves our ability to distinguish between models, particularly when performance differences are marginal. In practice, researchers can continue data collection until the confidence intervals are sufficiently narrow; their length shrinks as $1/\sqrt{|\cD^{(h)}|}$ while second-stage error dominates (Appendix \ref{sec:ci}). Finally, it is important to note that a smaller rank $L$ yields narrower intervals by reducing the variance of $\hat{\gamma}_0$; consequently, one must balance predictive performance with different uncertainty levels.

\subsection{Exploring strengths and weaknesses of different models}

\begin{figure}[t]
    \centering
     \begin{subfigure}[b]{.95\textwidth}
        \centering
        \includegraphics[width=\linewidth]{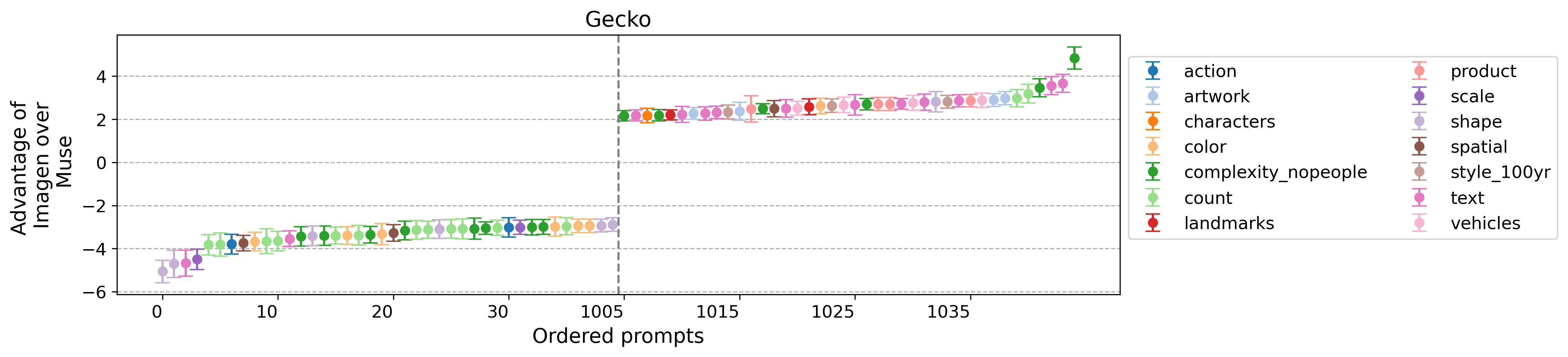}
    \end{subfigure} \\
    \begin{subfigure}[b]{\textwidth}
        \centering
        \includegraphics[width=.975\linewidth]{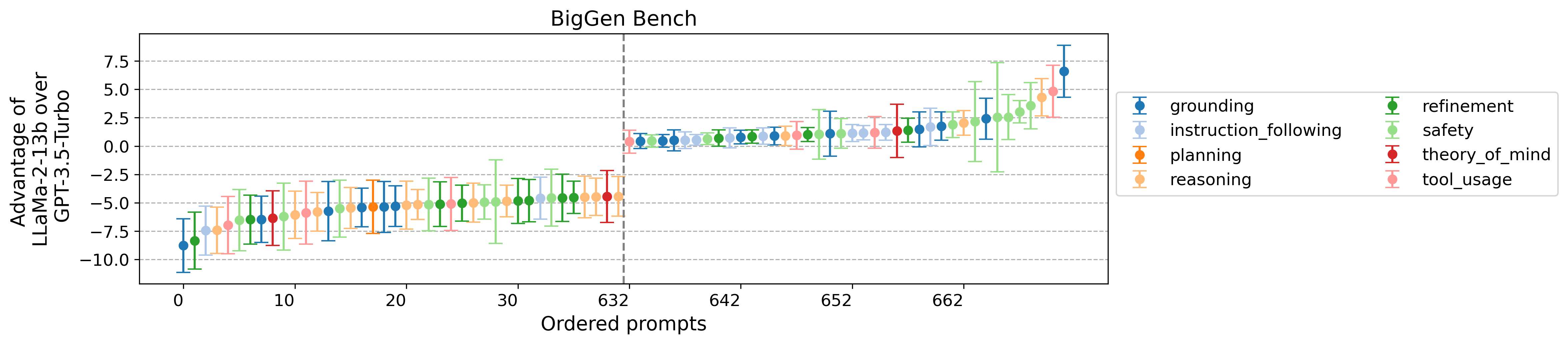}
    \end{subfigure}
    \caption{\small Fine-grained comparison of model capabilities using $10\%$ of human annotations. Top: Difference in estimated scores between Imagen and Muse on Gecko prompts, colored by category. Imagen excels in text rendering, while Muse shows advantages in object counting. Bottom: Difference between LLaMa-2-13b and GPT-3.5-Turbo on BigGen Bench prompts. GPT-3.5-Turbo demonstrates a significant advantage in reasoning-related prompts.}
\label{fig:deltas_cis_10}
\end{figure}

 For each benchmark, we compare the capabilities of two models across all prompts using the methodology introduced in \S \ref{sec:fine-grained}. For both Gecko and BGB, we plot the estimates for $\Psi_{m_1,p,0}-\Psi_{m_0,p,0}$ (along with $95\%$ confidence intervals with marginal coverage over all prompts)\footnote{These intervals are marginal: even for two equally capable models, about $5\%$ of prompts would exclude zero by chance.}, coloring them using a coarser set of categories for improved interpretability. Figure \ref{fig:deltas_cis_10} displays the results using only $10\%$ of the available human annotations. In the Gecko panel, we observe that many of the prompts where Imagen holds a significant advantage over Muse are related to text rendering; conversely, we see few such prompts where Muse holds a big advantage. Additionally, a substantial portion of the prompts where Muse outperforms Imagen involves object counting, a finding consistent with the conclusions drawn from Figure \ref{fig:ranking_cis_10_gecko}. Regarding BGB, we observe that the set of prompts where LLaMa-2-13b holds any advantage over GPT-3.5-Turbo is very limited; however, the data do not distinguish them for many instruction-following and safety prompts. Moreover, among the prompts where GPT-3.5-Turbo has a significant advantage over LLaMa-2-13b, a large proportion relates to reasoning capabilities. Figure \ref{fig:deltas_cis_100} conveys a similar message but utilizes a larger set of human annotations, thereby decreasing uncertainty and reducing the length of the confidence intervals.

\begin{figure}[t]
    \centering
    \begin{subfigure}[b]{\textwidth}
        \centering
        \includegraphics[width=.7\linewidth]{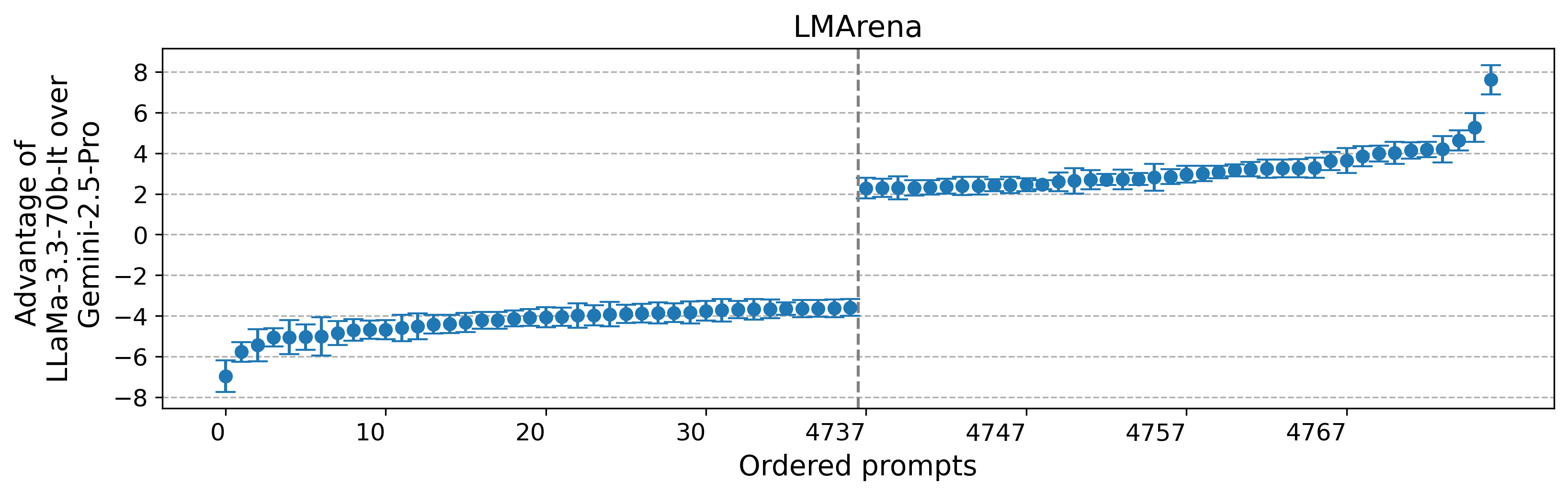}
    \end{subfigure}
    \caption{\small In LMArena, we estimate that LLaMa-3.3-70b-it is statistically better than Gemini-2.5-Pro in $\approx 7.8\%$ of the prompts, and that the data do not distinguish the two in another $\approx 25\%$. With all human labels, plug-in intervals can under-cover (Appendix \ref{sec:ci-marg}).}
\label{fig:deltas_cis_100_lmarena}
\vspace{-.5cm}
\end{figure}

Separately, we show the results for LMArena in Figure \ref{fig:deltas_cis_100_lmarena} using the full set of human annotations (one per prompt). In this plot, we compare LLaMa-3.3-70b-it with the state-of-the-art model Gemini-2.5-Pro. For this set of results, we verified that LLaMa is statistically better than Gemini in $\approx 7.8\%$ of the prompts\footnote{Confidence interval lower bound is greater than zero.} and that the data do not distinguish the two models in another $\approx 25\%$\footnote{Confidence interval includes zero.}. Failure to reject a difference is not evidence of equivalence, so these prompts are not ties. In this case, most of the prompts were never used to evaluate the two chosen models. In the appendix, Table \ref{tab:ols_capability} tries to explain the gaps in performance between the two models as a function of different tags provided in the LMArena dataset.


\subsection{Other interpretable insights}

\begin{wrapfigure}{r}{0.5\textwidth}
    \centering
    \vspace{-.5cm}
    \includegraphics[width=.85\linewidth]{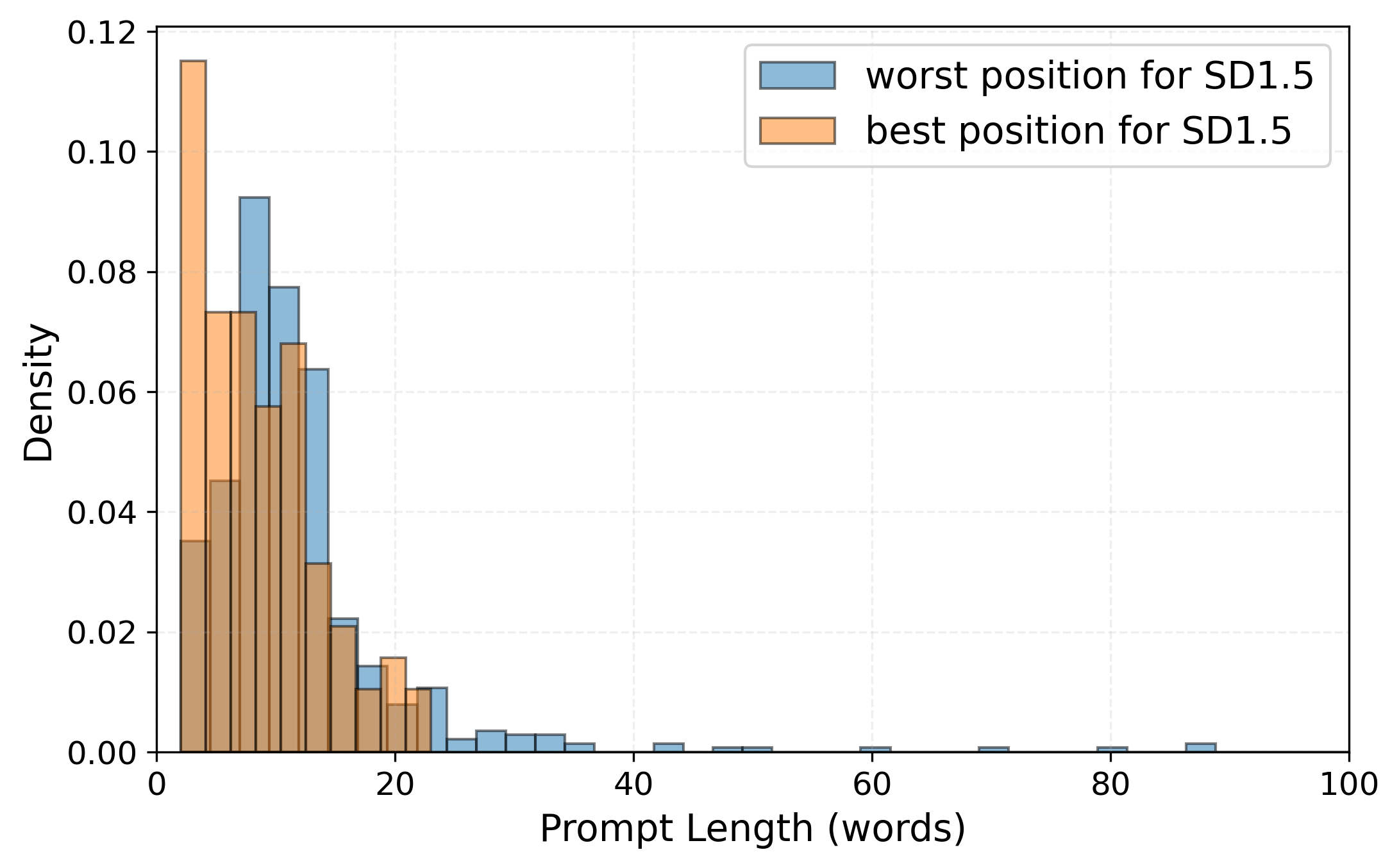}
    \caption{\small Prompt length for SD1.5 on the Gecko benchmark, comparing densities when it achieves its best versus worst rank. The shift shows SD1.5 performs relatively better on shorter prompts.}
    \label{fig:sd}
\end{wrapfigure}

In this subsection, we present an additional example of the interpretable insights achievable with our approach. While this result is specific to Gecko, similar analyses can be derived for other benchmarks. We characterize the properties of prompts where SD1.5 performs relatively well versus poorly. Figure \ref{fig:sd} compares the conditional distributions of prompt length, a proxy for prompt complexity in Gecko, given that SD1.5 achieves the best versus the worst rank; in this case, we use the full data to estimate parameters with the fine-tuning approach for more accurate results. The data reveal a clear trend: SD1.5 performs relatively better on less complex (shorter) prompts.

\section{Related work}\label{sec:related}

\textbf{Autoraters.} Autoraters are automated systems that score or rank the outputs of generative models given a task description; with the emergence of LLMs and multimodal LLMs, they are mostly instances of the LLM-as-a-Judge paradigm \citep{gu2024survey, li2024generation, li2024llms, chen2024mllm}. These systems often exhibit biases \citep{zheng2023judging,wang2023large,shi2024judging,dubois2024length, park2024offsetbias, thakur2024judging, ye2024justice, polo2025bridging} and do not align with human raters out of the box, typically requiring prompt refinement, in-context examples, fine-tuning, or post-hoc correction. As with human evaluators, they use single-sided or side-by-side templates. We employ autoraters not as final raters, but to generate auxiliary data from which we learn representations of prompts and generative models.

\textbf{Statistical and prompt-specific AI evaluation.} Statistical models make AI evaluation more data-efficient and insightful. The best known is the Bradley-Terry (BT) model and its variations \citep{bradley1952rank, rao1967ties, bohnet2024long, yuan2025benchmarking, frick2025prompt}, popularized by LMArena \citep{chiang2024chatbot}, which ranks models with a single scalar skill and assumes outcomes are independent of the prompt. Factor models, such as Gaussian Factor Analysis (FA) \citep{lawley1962factor, burnell2023revealing} and Item Response Theory (IRT) \citep{cai2016item, rodriguez2021evaluation, polo2024tinybenchmarks, zhou2025lost, yuan2025benchmarking}, allow several skills, in which case no global ranking exists and performance is prompt-dependent. Classical FA/IRT treats some parameters as random effects and integrates them out, which prevents consistent estimation of the prompt-level quantities we target; we therefore treat all parameters as fixed \citep{chen2019joint}. Closest to our goal, the Prompt-to-Leaderboard (P2L) framework \citep{frick2025prompt} builds prompt-specific leaderboards by fine-tuning a language model on human preferences; we instead leverage autoraters, which makes the approach feasible when human annotations are scarce.

\textbf{Efficient evaluation.} Strategies to reduce evaluation costs either evaluate on a subset of prompts and predict performance on the rest \citep{vivek2024anchor, perlitz2024efficient, polo2024tinybenchmarks, polo2024efficient, zhang2025benchmark}, select data points adaptively \citep{kossen2021active,liu2024survey,berrada2025scaling,angelopoulos2025cost}, or use label proxies to reduce the variance of estimates \citep{boyeau2024autoeval}. Our approach belongs to the last group: autorater data reduce the variance of our estimates.

\section{Conclusion}

The transition to fine-grained evaluation is essential for characterizing modern generative models, but is significantly bottlenecked by the high cost of human annotation. To address this, we presented a statistical framework rooted in tensor factorization that treats cost-effective autorater scores as auxiliary signals to learn shared latent representations of prompts and models. This approach effectively bridges the gap between scalable automated systems and high-quality human judgment, allowing for robust estimation of model capabilities even in data-scarce regimes where standard metrics often fail. Empirically, our method significantly outperforms baselines like Bradley-Terry and IRT by capturing complex model-prompt interactions while using auxiliary data to solve cold-start issues. We demonstrated practical utility by constructing granular leaderboards that reveal trade-offs hidden by aggregate metrics, such as identifying that models statistically indistinguishable on some prompts or sub-tasks can excel or fail on other instances, and by accurately predicting the performance of unseen models without requiring new human labels. We comment on limitations, future directions, and broader impacts in Appendix \ref{sec:lim-fut}.

\section*{Acknowledgements}
During the development of this project, we had several helpful discussions with Ashkan Khakzar, Pinelopi Papalampidi, Hannah Rashkin, Aleksandar Stani\'c, Reinald Kim Amplayo, and Junwei Huang. We are grateful for their contribution.

\bibliography{refs}

\appendix
\newpage
\section{Limitations, future directions, and broader impacts}\label{sec:lim-fut}

\textbf{Limitations.} First, our approach relies on a low-rank capability tensor and an ordinal logit model with rater-specific cutoffs. Low-rank structure has been observed across evaluation settings \citep{polo2024tinybenchmarks, fisch2026collabeval}, and in our benchmarks predictive performance saturates quickly in the rank (Figure \ref{fig:varying_rank}); moreover, $L$ is a hyperparameter that can be increased at the cost of more human labels. Second, with side-by-side human templates, only relative capabilities are identifiable, limiting cross-prompt comparability. Third, autorater data help only if some combination of autoraters carries information about human preferences. Leniency, severity, and position biases are absorbed by the rater-specific cutoffs, and biases in what the judges measure matter only when the panel is both redundant and misaligned with humans (Appendix \ref{sec:shared_bias}); in that case, the remedy is adding judges with different criteria (Figure \ref{fig:varying_autorater_fraction}). Relatedly, the custom autoraters for Gecko and LMArena share one backbone, although BGB, whose autoraters span several model families, yields the same conclusions. Finally, our plug-in confidence intervals condition on the first-stage representations and can under-cover when human labels are plentiful relative to autorater labels; the bootstrap of Appendix \ref{sec:ci} mitigates this at a higher computational cost, and neither applies after the optional fine-tuning stage.

\textbf{When not to use it.} When correctness is mechanically verifiable (\eg, code with an executable test suite), the verifier is effectively an oracle and should be used directly; our framework is most useful when judgments are graded or subjective and gold labels are expensive.

\textbf{Future directions.} While this work establishes a foundation for granular evaluation, several research avenues remain. Integrating active learning could optimize calibration by adaptively selecting model-prompt pairs for human labeling. Additionally, the estimated latent capabilities could serve as dense, human-aligned reward signals for reinforcement learning from human feedback (RLHF). Finally, it is possible to extend this framework to complex modalities like video and code, and to the evaluation of autonomous agents. This would involve capturing the multi-turn dynamics and environmental interactions inherent in agentic workflows, moving beyond static prompt-response pairs.

\textbf{Broader impacts.} We acknowledge that our work may have broader societal implications. On the one hand, reducing the need for human annotations could displace workers who rely on this activity. On the other hand, organizations may be able to do more with the same annotation budget, potentially increasing the aggregate value generated by such work. The net effect of these forces is ultimately an empirical question that can only be determined in practice.


\section{Confidence intervals}\label{sec:ci}

We consider two constructions: plug-in intervals, which treat the first-stage representations as fixed, and a parametric bootstrap, which refits both stages and thus also propagates first-stage uncertainty. Section \ref{sec:ci-marg} covers marginal intervals (\ie, with point-wise guarantees), used in the main text, and Section \ref{sec:ci-sim} covers intervals with simultaneous coverage.

\subsection{Marginal confidence intervals}\label{sec:ci-marg}

\textbf{Plug-in intervals.} Based on standard results in maximum likelihood estimation theory \citep[\eg,][]{wooldridge2010econometric}, the asymptotic distribution of $\hat{\gamma}_{0}$ is Gaussian. Specifically,
\[
\sqrt{|\cD^{(h)}|}(\hat{\gamma}_{0}-\gamma_{0}) \overset{d}{\rightarrow} \cN(0,\Sigma) \text{ as }|\cD^{(h)}|\to\infty,
\]
where ``$\overset{d}{\rightarrow}$'' denotes convergence in distribution and $|\cD^{(h)}|$ is the number of human annotations. The matrix $\Sigma$ can be estimated from the data; a consistent estimate $\hat{\Sigma}$ is typically derived from the inverse of the Hessian of the normalized negative log-likelihood loss function. In our methodology, we compute $\hat{\Sigma}$ as $|\cD^{(h)}|$ times the covariance matrix given by the \href{https://www.statsmodels.org/devel/generated/statsmodels.miscmodels.ordinal_model.OrderedResults.cov_params.html#statsmodels.miscmodels.ordinal_model.OrderedResults.cov_params}{statsmodels} implementation, which estimates the covariance of $\hat{\gamma}_{0}$ itself, \ie, $\Sigma/|\cD^{(h)}|$. 

Consequently, for a given pair $(m,p)$, we have
\[
\sqrt{|\cD^{(h)}|}(\hat{\Psi}_{m,p,0}-\Psi_{m,p,0}) \overset{d}{\rightarrow} \cN(0,v_{m,p}^\top\Sigma v_{m,p}),
\]
or, equivalently, after standardization,
\[
\sqrt{\frac{|\cD^{(h)}|}{v_{m,p}^\top\hat{\Sigma} v_{m,p}}}(\hat{\Psi}_{m,p,0}-\Psi_{m,p,0}) \overset{d}{\rightarrow} \cN(0,1).
\]

It follows that the inequality
\[
-\Phi^{-1}\left(\frac{1+\lambda}{2}\right)\leq\sqrt{\frac{|\cD^{(h)}|}{v_{m,p}^\top\hat{\Sigma} v_{m,p}}}(\hat{\Psi}_{m,p,0}-\Psi_{m,p,0})\leq \Phi^{-1}\left(\frac{1+\lambda}{2}\right)\Rightarrow
\]
\[
\Rightarrow \hat{\Psi}_{m,p,0}-\Phi^{-1}\left(\frac{1+\lambda}{2}\right)\sqrt{\frac{v_{m,p}^\top\hat{\Sigma} v_{m,p}}{|\cD^{(h)}|}}\leq\Psi_{m,p,0}\leq \hat{\Psi}_{m,p,0}+\Phi^{-1}\left(\frac{1+\lambda}{2}\right)\sqrt{\frac{v_{m,p}^\top\hat{\Sigma} v_{m,p}}{|\cD^{(h)}|}}
\]

holds with probability approaching $\lambda\in(0,1)$. Therefore, we define the pointwise asymptotic confidence interval as
\[
\text{CI}_\lambda =\hat{\Psi}_{m,p,0}\pm\Phi^{-1}\left(\frac{1+\lambda}{2}\right)\sqrt{\frac{v_{m,p}^\top\hat{\Sigma} v_{m,p}}{|\cD^{(h)}|}}.
\]

Here, $\gamma_0$ denotes the population value of the second-stage parameter given the first-stage representations $\widehat{\Lambda}^{(a)}$; it coincides with the true human embedding only when those representations are exact, which is why these intervals are valid conditionally on them.

\textbf{Bootstrap intervals.} Let $\Omega$ be a finite set indexing the quantities for which we want intervals. Each index $\omega\in\Omega$ is either a model-prompt pair $(m,p)$, a pair of models and a prompt $((m_0,m_1),p)$, or a model and a category $(m,\cP_s)$, and we write $\tau_\omega(\Lambda)$ for the corresponding quantity computed from parameters $\Lambda$: $\Psi_{m,p,0}$, $\Delta_{m,p,0}$, or $\Psi_{m,\cP_s,0}$, respectively. The target is $\tau_\omega\triangleq\tau_\omega(\Lambda)$ and its estimate is $\hat{\tau}_\omega\triangleq\tau_\omega(\widehat{\Lambda})$. For categories, $\alpha_{\cP_s}$ is recomputed from each fit; since a leading eigenvector is defined only up to sign, we align its sign with that of the original fit. Because the fit does not impose the conditions of Appendix \ref{sec:ident}, two fits that induce the same label distribution can return human capabilities that differ by a shift (permutations and rescalings of the latent factors leave $\Psi$ unchanged and do not matter here). With a single-sided human template, adding a constant to every $\Psi_{m,p,0}$ and to every human cutoff $\beta^{(0)}_y$ leaves Equation \eqref{eq:model} unchanged; with a side-by-side template, only differences $\Psi_{m_1,p,0}-\Psi_{m_0,p,0}$ enter it, so each prompt $p$ admits its own shift. Rather than constraining the fit, we remove the shift after fitting, choosing the representative singled out by Conditions \ref{cond:shifts} and \ref{cond:sparsity}: when computing $\tau_\omega(\Lambda)$, we replace $\Psi_{m,p,0}$ by $\Psi_{m,p,0}-\beta^{(0)}_1$ for single-sided human templates (so that $\beta^{(0)}_1=0$) and by $\Psi_{m,p,0}-\Psi_{0,p,0}$ for side-by-side ones (so that model $0$ is the reference), and analogously for $\Psi_{m,\cP_s,0}$. This normalization changes neither $\Delta_{m,p,0}$ nor the ranking of models within a prompt or category. Algorithm \ref{alg:boot} is a parametric bootstrap: it draws $K$ new datasets from the fitted model, keeping the observed designs (the model(s), prompt, and rater of every label) and resampling only the labels, refits both stages on each dataset, and recomputes the targets, giving replicates $\hat{\tau}_{\omega,k}$, $k=1,\dots,K$. The bootstrap standard error $\widehat{\text{se}}_\omega$, the standard deviation of $\{\hat{\tau}_{\omega,k}\}_{k=1}^K$, replaces the plug-in one and gives the marginal interval $\hat{\tau}_\omega\pm\Phi^{-1}\left(\frac{1+\lambda}{2}\right)\widehat{\text{se}}_\omega$.

\begin{algorithm}[h]
 \textbf{Input:} (i) Datasets $\cD^{(a)}$ and $\cD^{(h)}$, (ii) estimates $\widehat{\Lambda}=(\widehat{\Lambda}^{(h)},\widehat{\Lambda}^{(a)})$, (iii) index set $\Omega$ of the targets, (iv) confidence level $\lambda$, (v) number of replicates $K$;

 \textbf{Output:} Marginal and simultaneous confidence intervals for the targets $\{\tau_\omega\}_{\omega\in\Omega}$;
 \smallskip

 \begin{algorithmic}[1]
 \State \textit{Compute the targets on the original fit:} $\hat{\tau}_\omega \leftarrow \tau_\omega(\widehat{\Lambda})$ for all $\omega\in\Omega$
 \For{$k = 1 \dots K$ (in parallel)}
    \State \textbf{1. Resample Labels:}
    \State \textit{Keep the observed designs and draw new outcomes from Equation \eqref{eq:model}:}
    \State $\displaystyle Y^{(n)}_k \sim \Pr_{\widehat{\Lambda}}\big(Y_{m^{(n)},p^{(n)},r^{(n)}}=\cdot\,\big)$ for all $D_n \in \cD^{(a)}\cup\cD^{(h)}$
    \State \textbf{2. Refit Both Stages:}
    \State $\widehat{\Lambda}^{(a)}_k \leftarrow$ Algorithm \ref{alg:stage1} on $\cD^{(a)}_k$, initialized at $\widehat{\Lambda}^{(a)}$
    \State $\widehat{\Lambda}^{(h)}_k \leftarrow$ second stage on $\cD^{(h)}_k$, with $\widehat{\Lambda}^{(a)}_k$ frozen
    \State \textbf{3. Recompute and Normalize Targets:}
    \State \textit{For categories, align the sign of $\alpha_{\cP_s}$ with the original fit}
    \State $\displaystyle \hat{\tau}_{\omega,k} \leftarrow \tau_\omega\big(\widehat{\Lambda}^{(h)}_k,\widehat{\Lambda}^{(a)}_k\big)$ for all $\omega\in\Omega$
 \EndFor
 \State \textbf{4. Standard Errors and Critical Values:}
 \State $\displaystyle \widehat{\text{se}}_\omega \leftarrow \text{sd}_k\big(\hat{\tau}_{\omega,k}\big), \quad t_k \leftarrow \max_{\omega\in\Omega} \frac{|\hat{\tau}_{\omega,k}-\hat{\tau}_\omega|}{\widehat{\text{se}}_\omega}$
 \State $\displaystyle c_{\text{marg}} \leftarrow \Phi^{-1}\left(\frac{1+\lambda}{2}\right), \quad c_{\text{sim}} \leftarrow \lambda\text{-quantile of } \{t_k\}_{k=1}^K$
 \State \textbf{return} $\hat{\tau}_\omega\pm c_{\text{marg}}\,\widehat{\text{se}}_\omega$ (marginal) and $\hat{\tau}_\omega\pm c_{\text{sim}}\,\widehat{\text{se}}_\omega$ (simultaneous)
 \end{algorithmic}
\caption{Parametric bootstrap for confidence intervals}
\label{alg:boot}
\end{algorithm}

\textbf{Simulation instance.} In our simulations, all raters use binary single-sided templates ($C_r=2$, $\Delta_{m,p,r}=\Psi_{m,p,r}$) with cutoffs fixed at zero, so $\Psi$ is identified and the normalization does nothing. Step 1 then draws $Y^{(n)}_k\sim\text{Bernoulli}\big(\sigma(\hat{\Psi}_{m^{(n)},p^{(n)},r^{(n)}})\big)$, the second stage is a logistic regression of the human labels on $v_{m,p}$ without intercept, and the targets are $\Psi_{m,p,0}$ for all $(m,p)$. We use $K=100$, and warm-started refits run for $2{,}000$ epochs with learning rate $0.02$.

\textbf{Simulation setup.} We draw data from the model with $M=10$, $P\in\{100,500\}$, $R=21$ (one human and 20 autoraters), $L=10$, $\Theta_{m,l},A_{p,l}\sim\cN(0,1)$, and $\Gamma_{r,l}\sim\cN(0,1/L)$. Each autorater labels every cell $n_{\text{reps}}\in\{8,16,32,64\}$ times, and the human provides $|\cD^{(h)}|\in\{100,250,500\}$ labels on uniformly drawn cells; for $P=500$ and $|\cD^{(h)}|=100$, this is $0.2$ labels per prompt. For each configuration, we fix the parameters, draw $100$ independent datasets, and record whether each $95\%$ interval contains the true $\Psi_{m,p,0}$. As a control, an oracle fits only the second stage, using the true $\Theta$ and $A$.

\begin{figure}[H]
    \centering
    \begin{subfigure}[b]{0.32\textwidth}
        \centering
        \includegraphics[width=\linewidth]{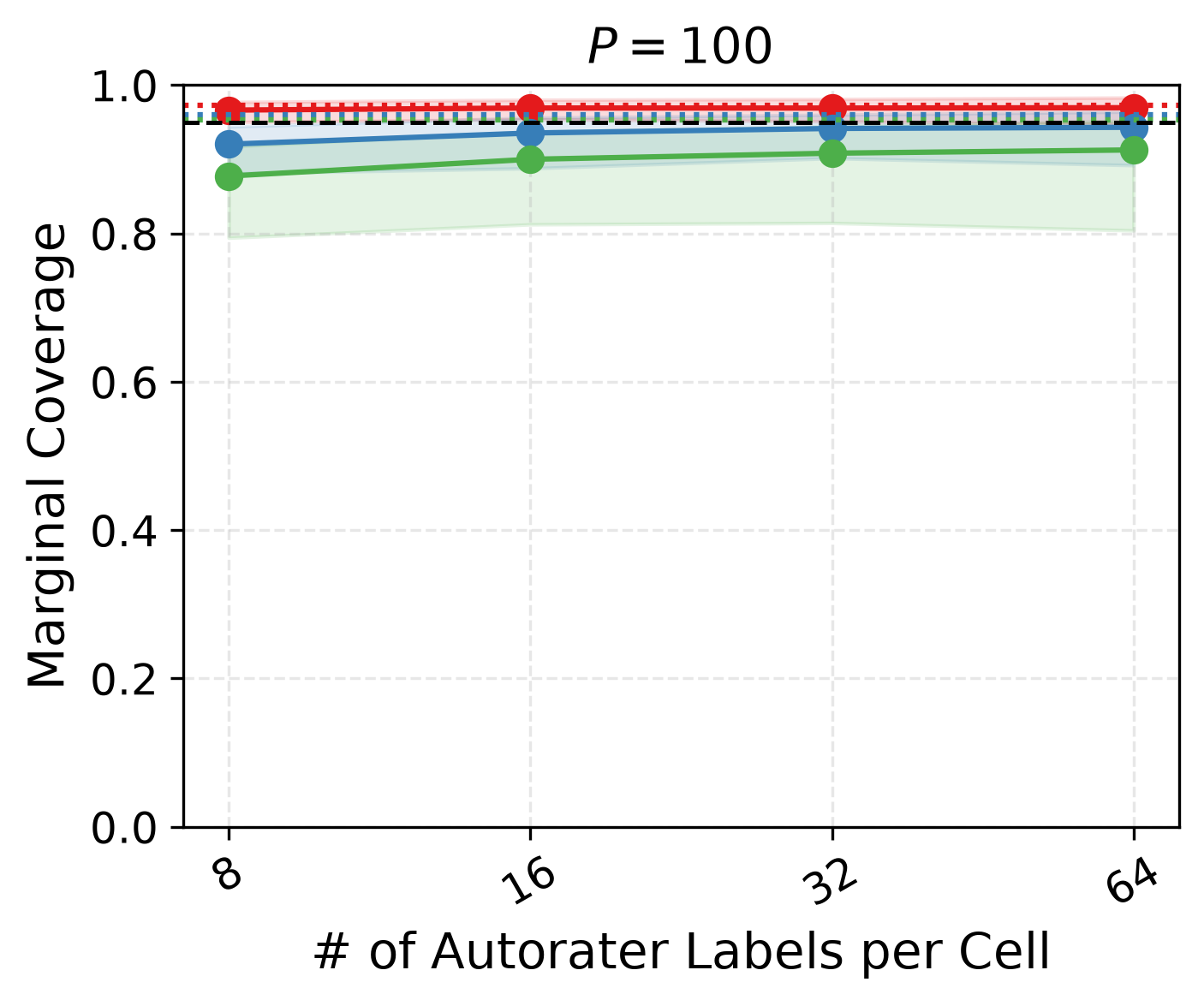}
    \end{subfigure}
    \begin{subfigure}[b]{0.44\textwidth}
        \centering
        \includegraphics[width=\linewidth]{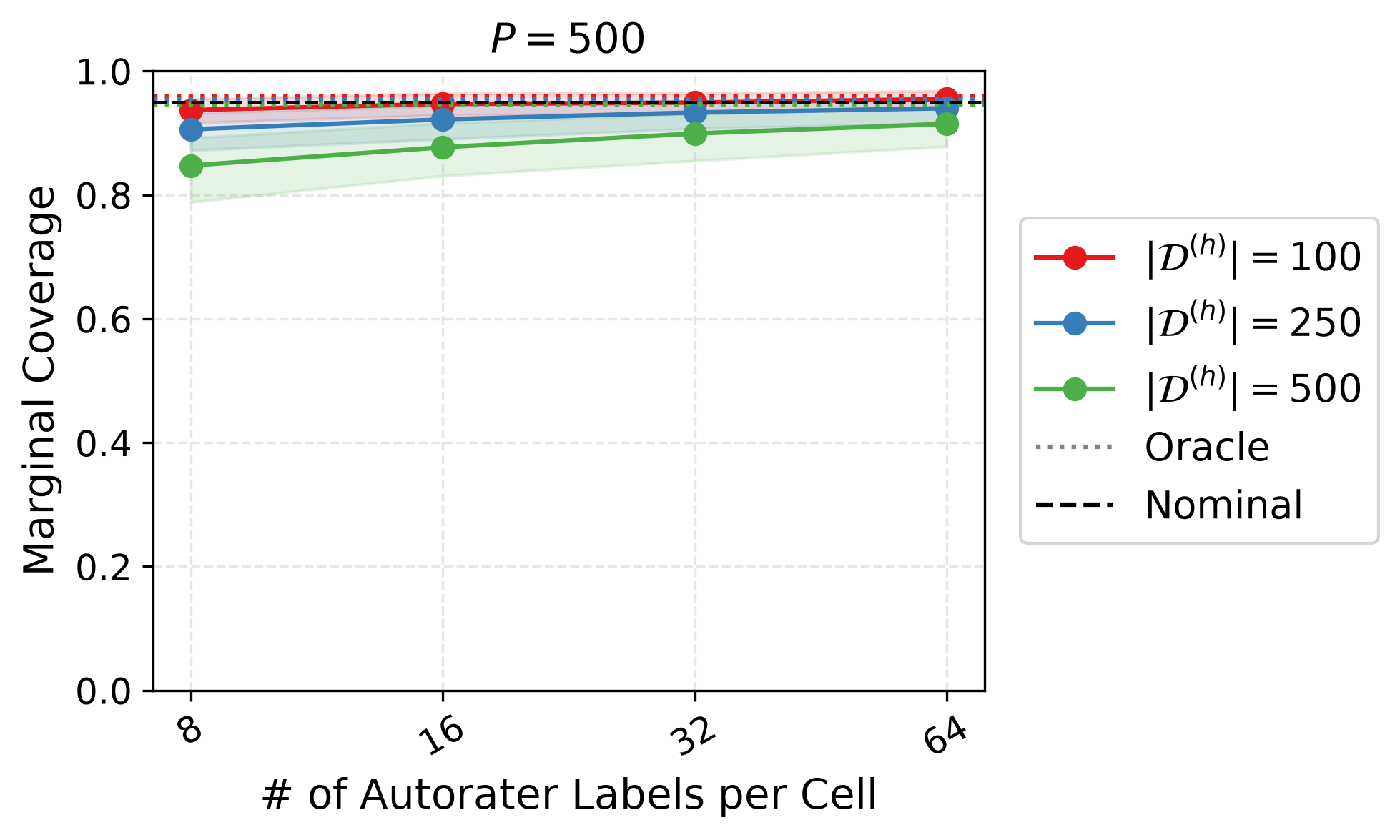}
    \end{subfigure}
    \caption{\small Marginal coverage of $95\%$ plug-in intervals against the number of autorater labels per cell, for $P=100$ (left) and $P=500$ (right). Shaded bands span 5--95\% of prompts; dotted lines show the oracle.}
    \label{fig:sim-marg}
\end{figure}

\begin{figure}[H]
    \centering
    \begin{subfigure}[b]{0.3\textwidth}
        \centering
        \includegraphics[width=\linewidth]{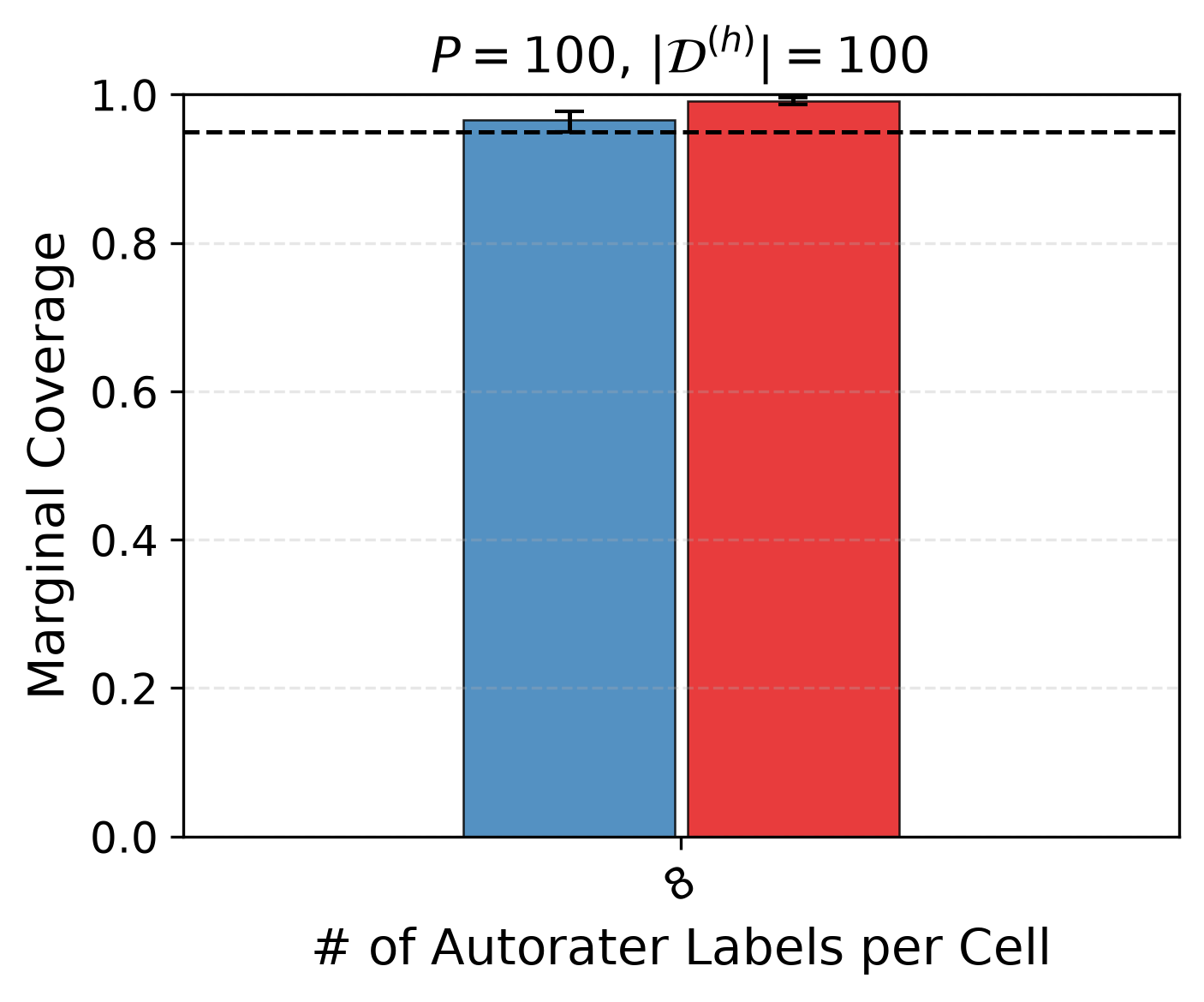}
    \end{subfigure}
    \begin{subfigure}[b]{0.3\textwidth}
        \centering
        \includegraphics[width=\linewidth]{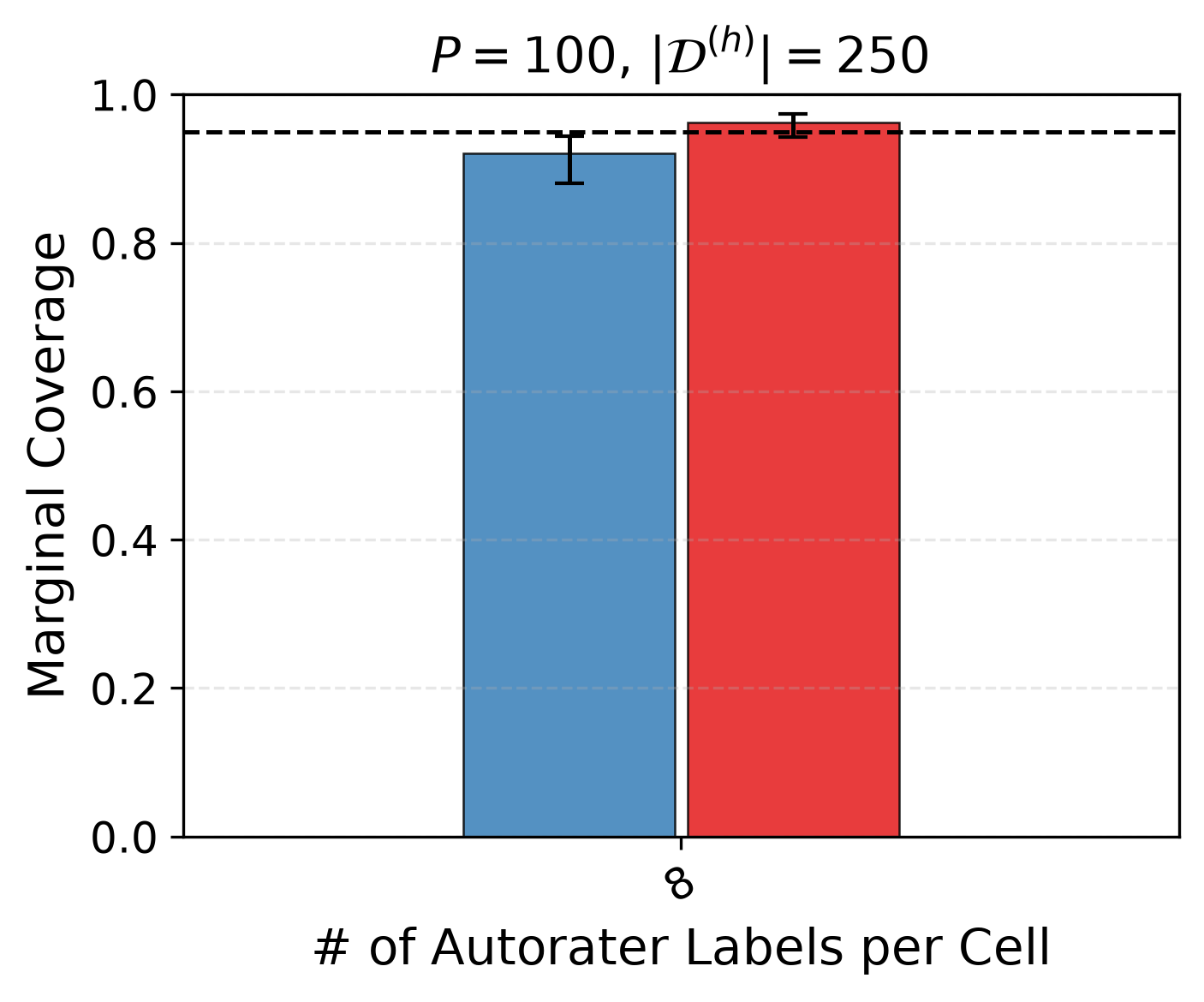}
    \end{subfigure}
    \begin{subfigure}[b]{0.3\textwidth}
        \centering
        \includegraphics[width=\linewidth]{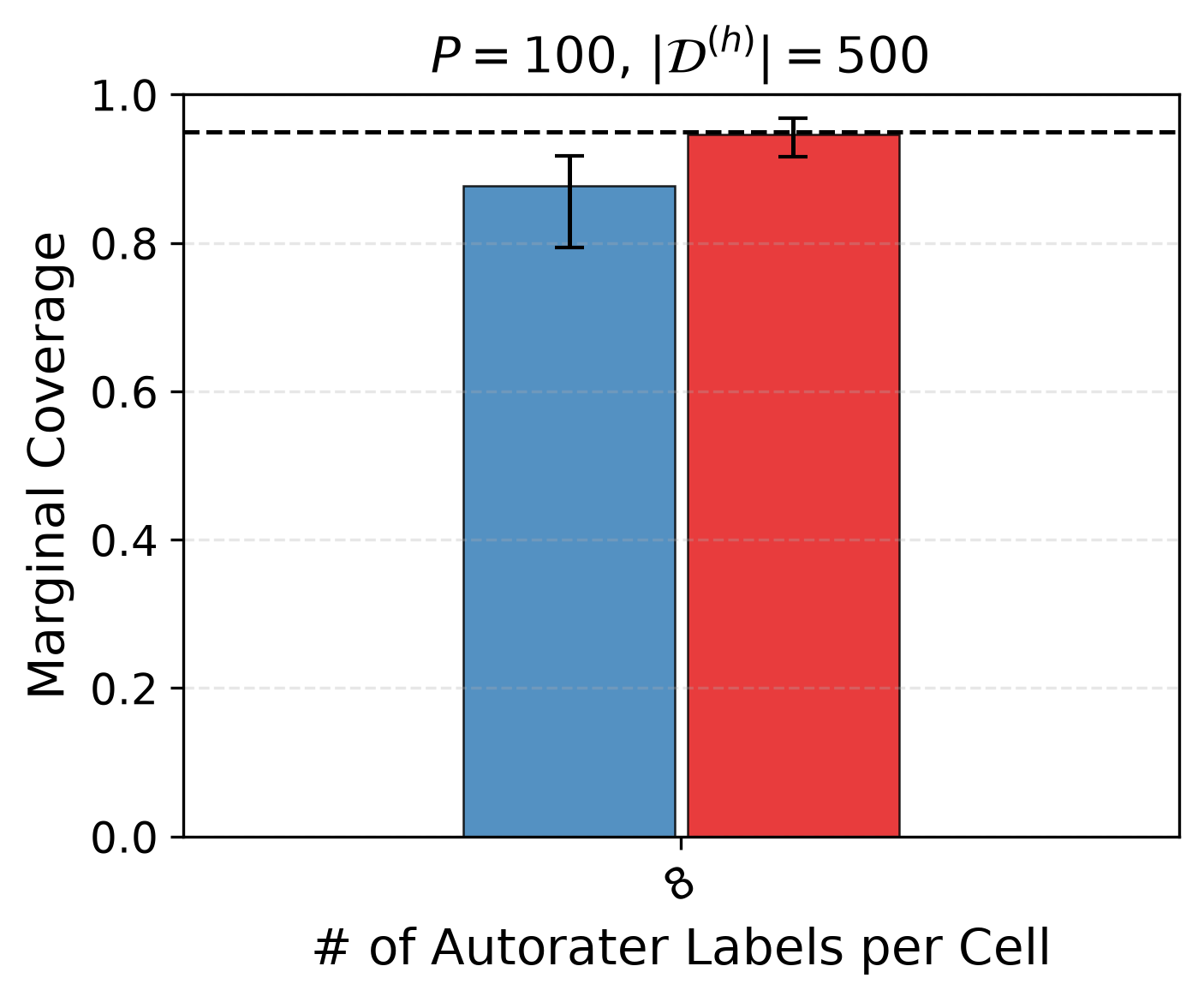}
    \end{subfigure}\\
    \begin{subfigure}[b]{0.27\textwidth}
        \centering
        \includegraphics[width=\linewidth]{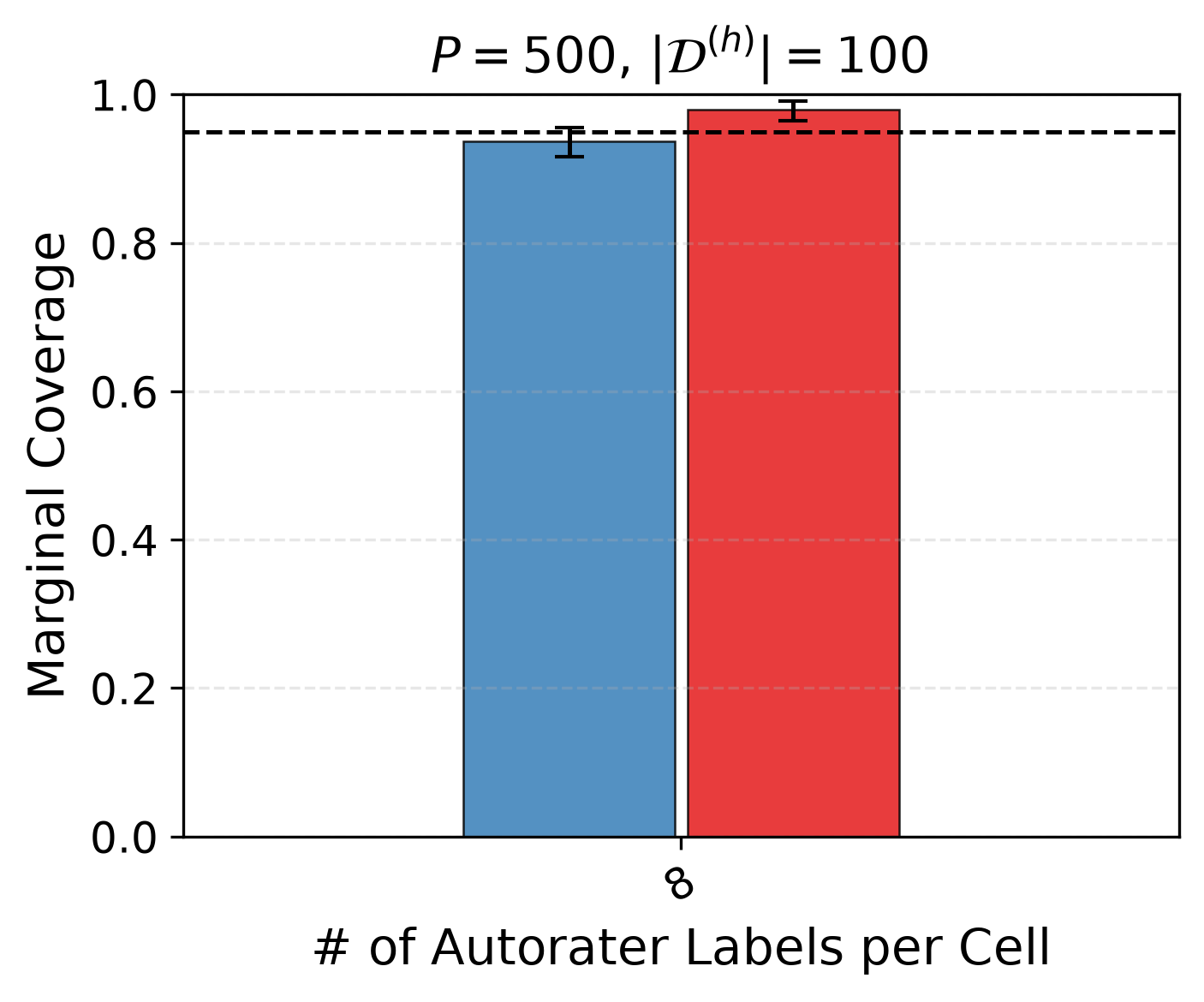}
    \end{subfigure}
    \begin{subfigure}[b]{0.27\textwidth}
        \centering
        \includegraphics[width=\linewidth]{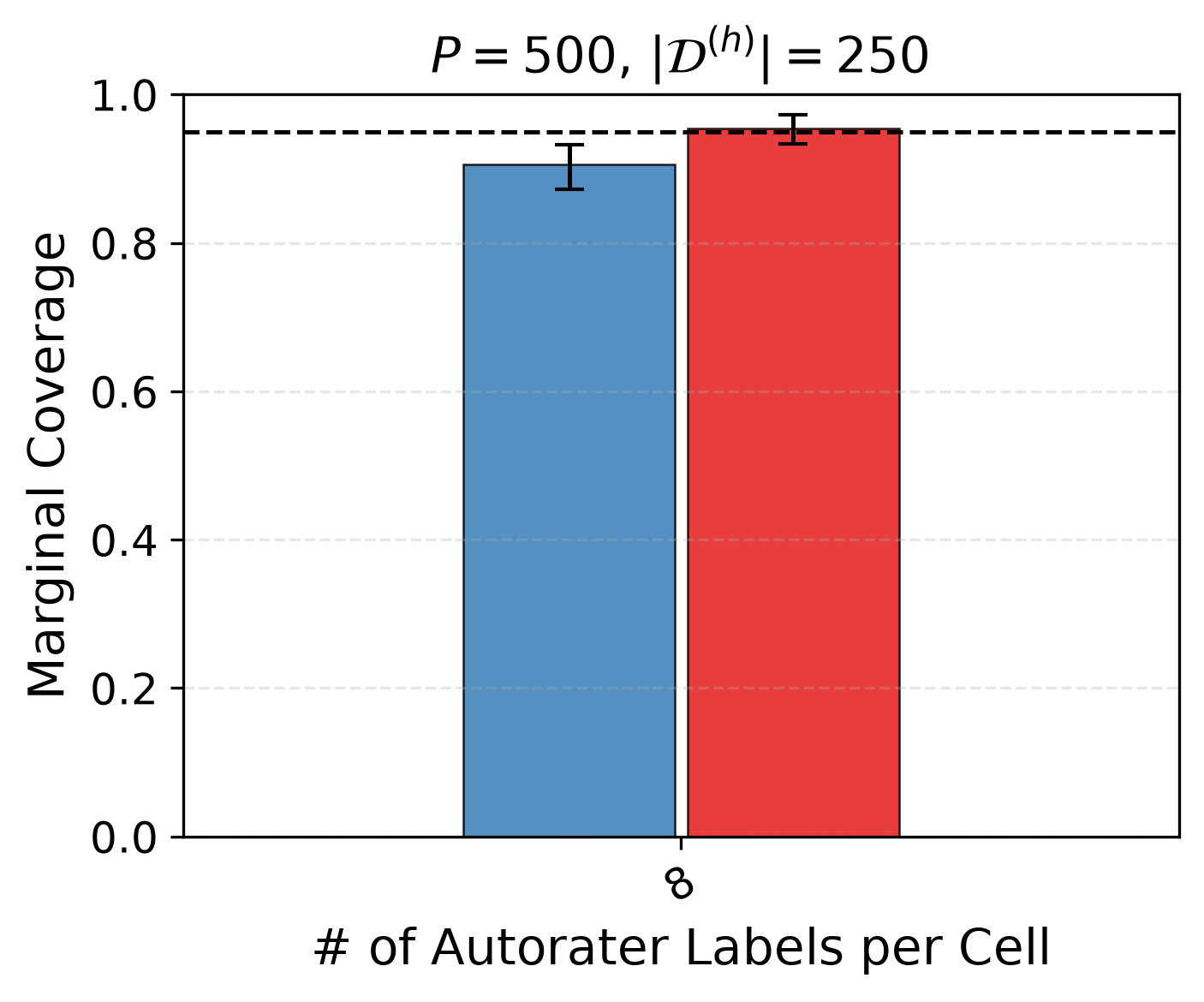}
    \end{subfigure}
    \begin{subfigure}[b]{0.38\textwidth}
        \centering
        \includegraphics[width=\linewidth]{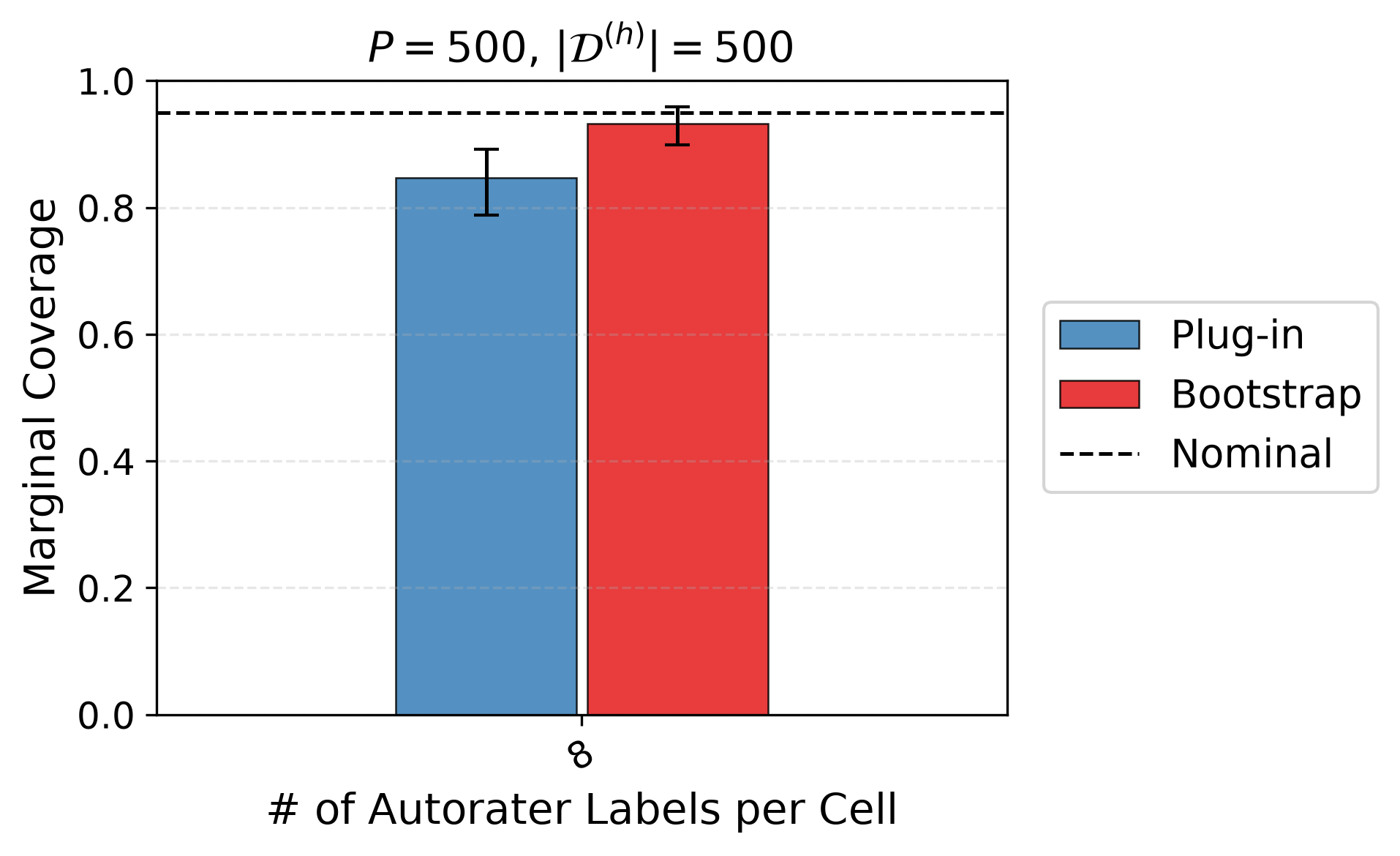}
    \end{subfigure}
    \caption{\small Marginal coverage of plug-in and bootstrap intervals on the same fitted models, one panel per $(P, |\cD^{(h)}|)$, with $n_{\text{reps}}=8$. Error bars span 5--95\% of prompts.}
    \label{fig:sim-boot-marg}
\end{figure}

\textbf{Results.} Figure \ref{fig:sim-marg} shows that plug-in marginal intervals are close to nominal when human labels are few relative to autorater labels ($0.97$ for $P=100$ and $|\cD^{(h)}|=100$), but under-cover as $|\cD^{(h)}|$ grows ($0.85$ to $0.92$ for $P=500$ and $|\cD^{(h)}|=500$); coverage improves with $n_{\text{reps}}$, and the oracle stays at $0.95$ to $0.97$. The reason is that, with more human labels, the intervals concentrate around the best readout of the \emph{estimated} representations rather than around $\Psi_{m,p,0}$. Bootstrap marginal intervals cover $0.93$ to $0.99$ in all settings (Figure \ref{fig:sim-boot-marg}), at the cost of intervals $1.15$ to $1.24\times$ wider; pooling all settings, a least-squares fit through the origin of the bootstrap width on the plug-in width gives a multiplier of $1.19$ (Figure \ref{fig:sim-widths}, left).

\textbf{Which interval to use.} The plug-in intervals suffice when the autorater data carry much more information than the human data, \ie, many autorater labels per cell relative to the number of human labels; otherwise, and whenever computation allows $K$ refits, the bootstrap is the safer choice.

\subsection{Simultaneous confidence intervals}\label{sec:ci-sim}

\textbf{Plug-in intervals.} We fix the confidence level $\lambda\in(0,1)$ for the remainder of this section and let $\Omega \subseteq \cM \times \cP$.

Our objective is to derive random intervals $R_{m,p}$ providing simultaneous coverage such that
\[
\Pr\left(\bigcap_{(m,p)\in\Omega} \{\Psi_{m,p,0} \in R_{m,p}\}  \right) \to \lambda
\]
as $|\cD^{(h)}|\to\infty$. To this end, let us define a matrix $V$ such that its rows correspond to the vectors $v_{m,p}$ for all $(m,p)\in\Omega$. We further define the vectors $\Psi_{\Omega,0}\triangleq V\gamma_0$ and $\hat{\Psi}_{\Omega,0}\triangleq V\hat{\gamma}_0$. Note that the entries of $\Psi_{\Omega,0}$ (resp. $\hat{\Psi}_{\Omega,0}$) correspond to $\Psi_{m,p,0}$ (resp. $\hat{\Psi}_{m,p,0}$) for indices in $\Omega$. By the properties of linear transformations of Gaussian vectors,
\[
\sqrt{|\cD^{(h)}|}(\hat{\Psi}_{\Omega,0}-\Psi_{\Omega,0}) \overset{d}{\rightarrow} \cN(0,\Xi),
\]
where $\Xi=V\Sigma V^\top$. We define the estimator $\hat{\Xi}=V\hat{\Sigma} V^\top$. We then construct the intervals as
\[
R_{m,p}=\hat{\Psi}_{m,p,0}\pm c\sqrt{\frac{\widehat{\Var}(\hat{\Psi}_{m,p,0})}{|\cD^{(h)}|}},
\]
where $\widehat{\Var}(\hat{\Psi}_{m,p,0})$ denotes the estimate of the asymptotic variance of $\hat{\Psi}_{m,p,0}$, corresponding to the diagonal entry of $\hat{\Xi}$ associated with $(m,p)$. Here, $c$ is a positive real constant that requires calibration. 

To determine the value of $c$, observe that
\[
\Pr\left(\bigcap_{(m,p)\in\Omega} \{\Psi_{m,p,0} \in R_{m,p}\}  \right)=\Pr\left(\bigcap_{(m,p)\in \Omega} \left\{\frac{\left|\hat{\Psi}_{m,p,0}-\Psi_{m,p,0}\right|}{\sqrt{\widehat{\Var}(\hat{\Psi}_{m,p,0})/|\cD^{(h)}|}}\leq c\right\}  \right)=\Pr\left(\max_{(m,p)\in \Omega} \frac{\left|\hat{\Psi}_{m,p,0}-\Psi_{m,p,0}\right|}{\sqrt{\widehat{\Var}(\hat{\Psi}_{m,p,0})/|\cD^{(h)}|}}\leq c  \right).  
\]

Consequently, we must select the constant $c$ satisfying
\[
\Pr\left(\max_{(m,p)\in \Omega} \frac{\left|\hat{\Psi}_{m,p,0}-\Psi_{m,p,0}\right|}{\sqrt{\widehat{\Var}(\hat{\Psi}_{m,p,0})/|\cD^{(h)}|}}\leq c  \right)= \lambda.
\]
The exact finite-sample distribution of the maximum statistic is unknown. In practice, however, we estimate $c$ via Monte Carlo integration, relying on the fact that the joint distribution of the standardized terms is asymptotically Gaussian with zero means and unit marginal variances. As $|\cD^{(h)}|$ grows, the quantiles of the maximum statistic converge to the quantiles of a limiting distribution defined by the maximum absolute value of a multivariate Gaussian vector with covariance matrix $D^{-1/2}\Xi D^{-1/2}$, where $D=\diag(\Xi)$. 

The ideas discussed in this section follow the general framework for simultaneous inference in parametric models described by \citet{hothorn2008simultaneous}.

\textbf{Bootstrap intervals.} Algorithm \ref{alg:boot} also returns a simultaneous band, with critical value $c_{\text{sim}}$ given by the $\lambda$-quantile of the maximum standardized deviation.

\begin{figure}[H]
    \centering
    \begin{subfigure}[b]{0.32\textwidth}
        \centering
        \includegraphics[width=\linewidth]{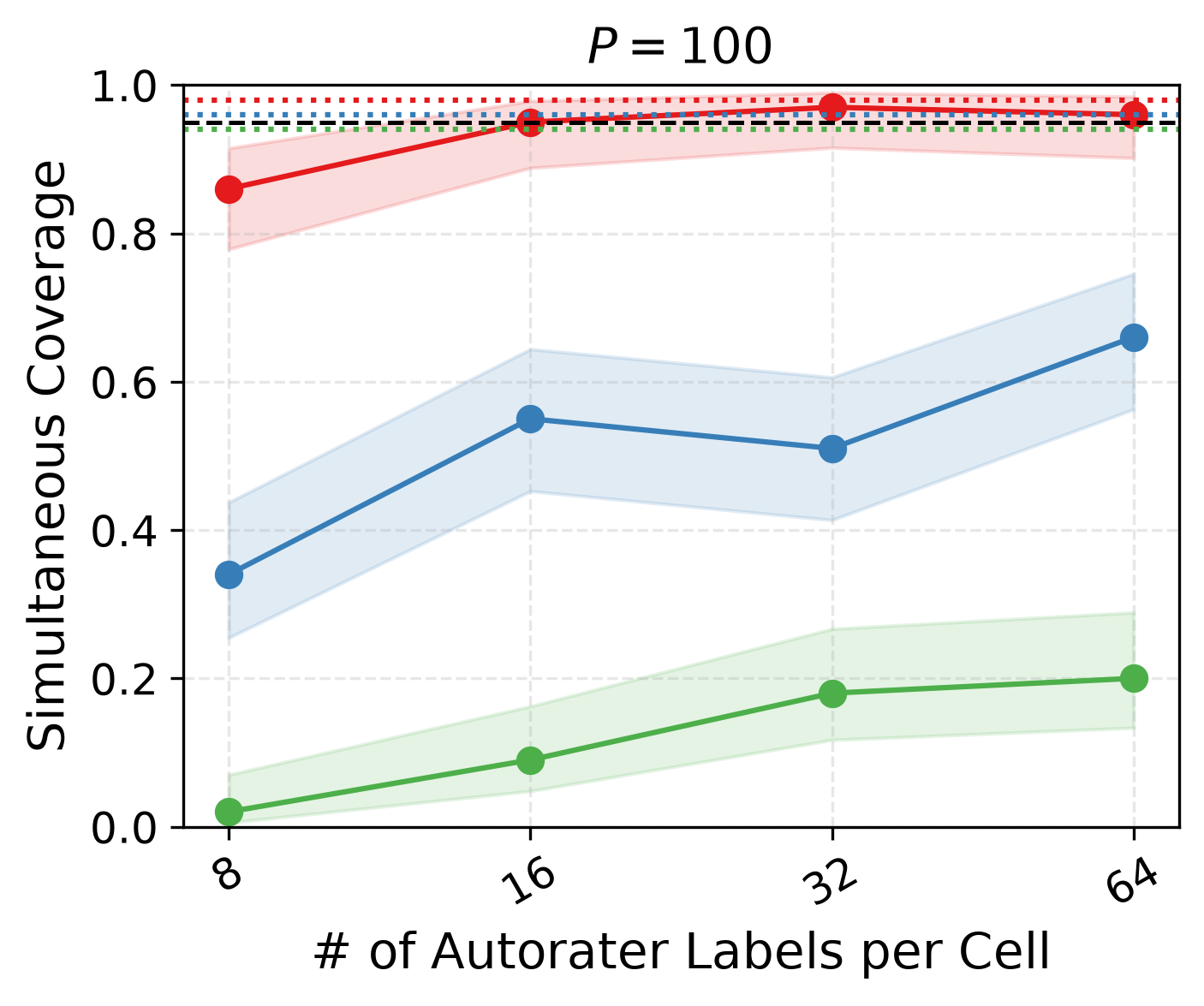}
    \end{subfigure}
    \begin{subfigure}[b]{0.44\textwidth}
        \centering
        \includegraphics[width=\linewidth]{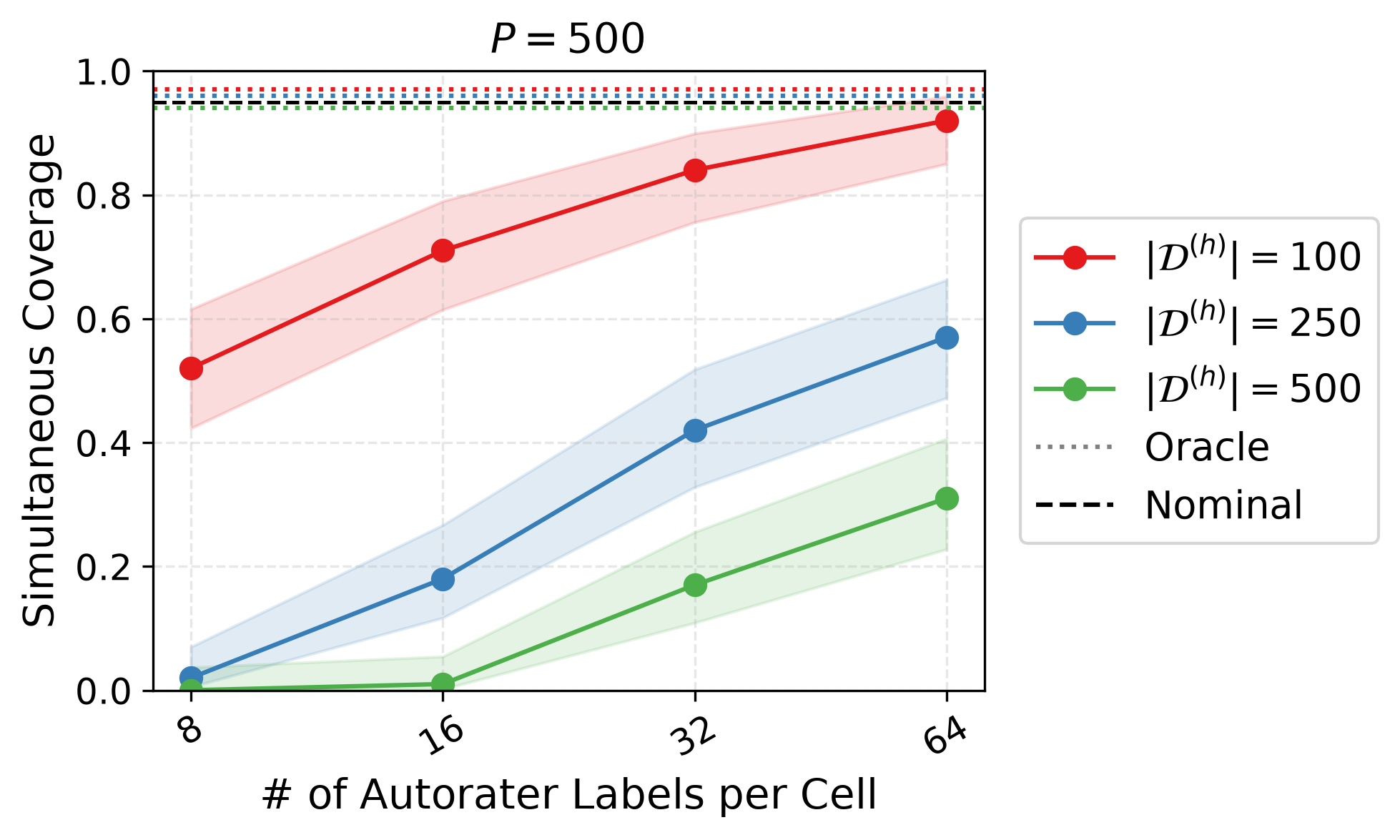}
    \end{subfigure}
    \caption{\small Simultaneous coverage of $95\%$ plug-in intervals, for $P=100$ (left) and $P=500$ (right), with $95\%$ Wilson bands; dotted lines show the oracle.}
    \label{fig:sim-sim}
\end{figure}

\begin{figure}[H]
    \centering
    \begin{subfigure}[b]{0.3\textwidth}
        \centering
        \includegraphics[width=\linewidth]{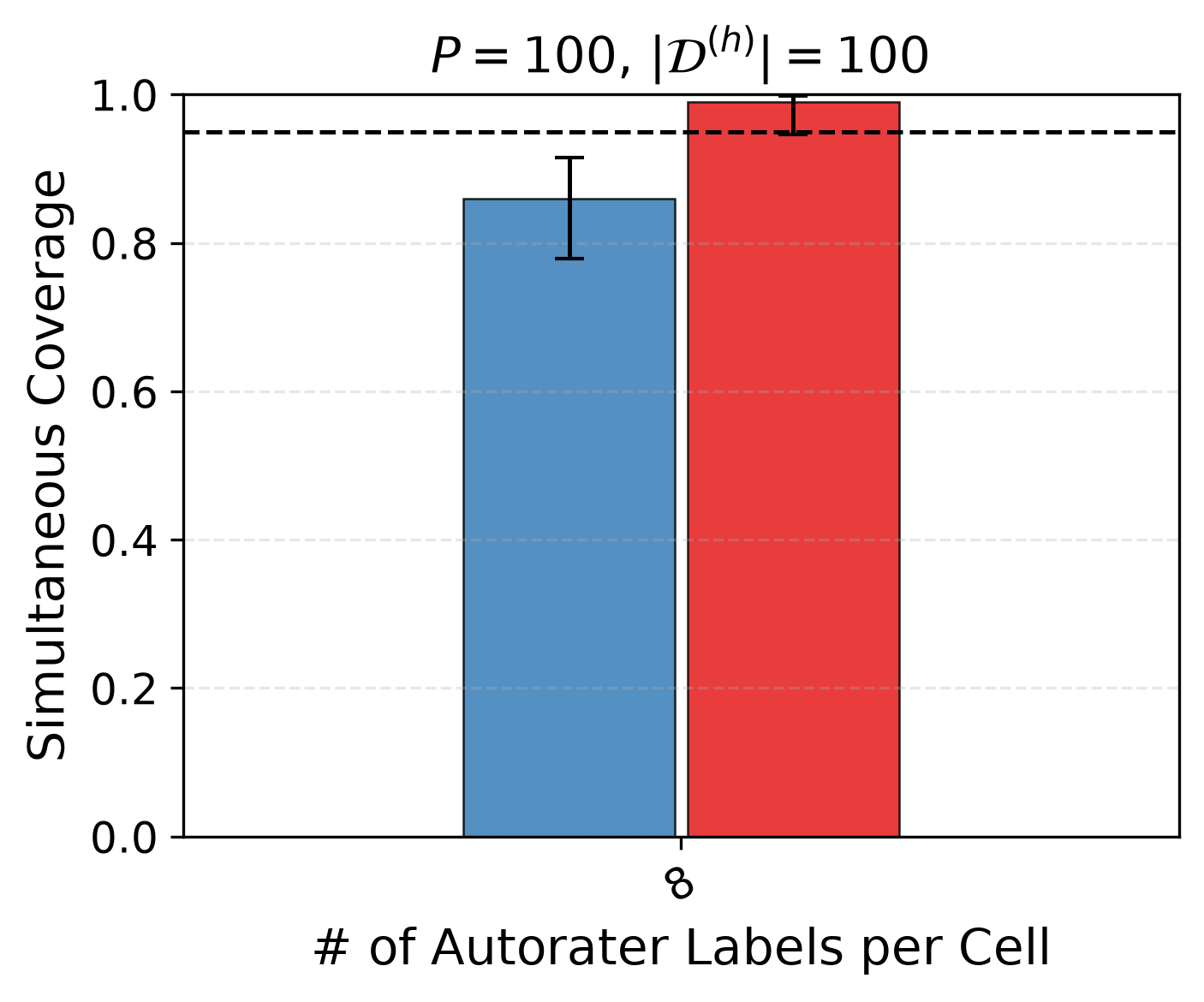}
    \end{subfigure}
    \begin{subfigure}[b]{0.3\textwidth}
        \centering
        \includegraphics[width=\linewidth]{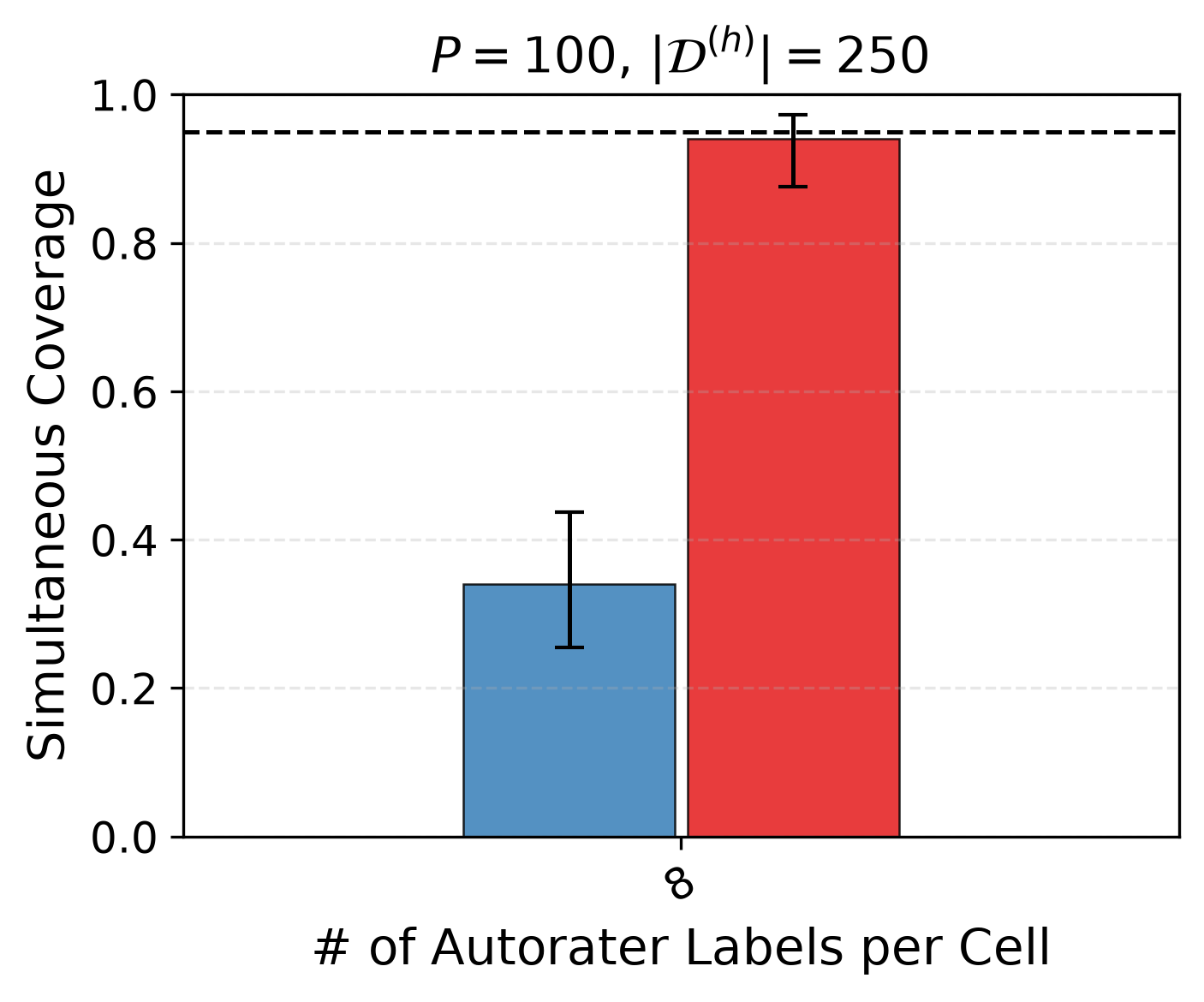}
    \end{subfigure}
    \begin{subfigure}[b]{0.3\textwidth}
        \centering
        \includegraphics[width=\linewidth]{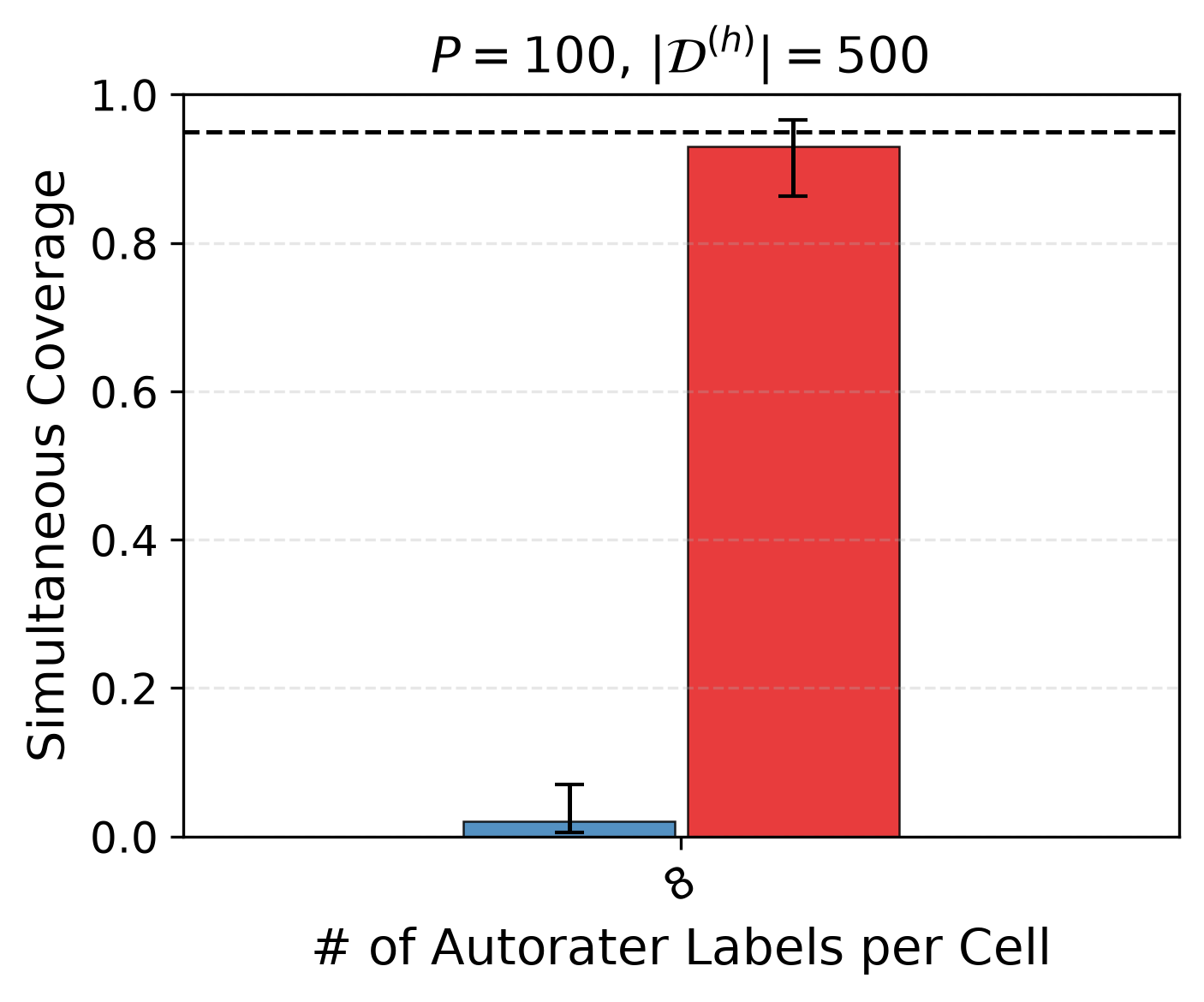}
    \end{subfigure}\\
    \begin{subfigure}[b]{0.27\textwidth}
        \centering
        \includegraphics[width=\linewidth]{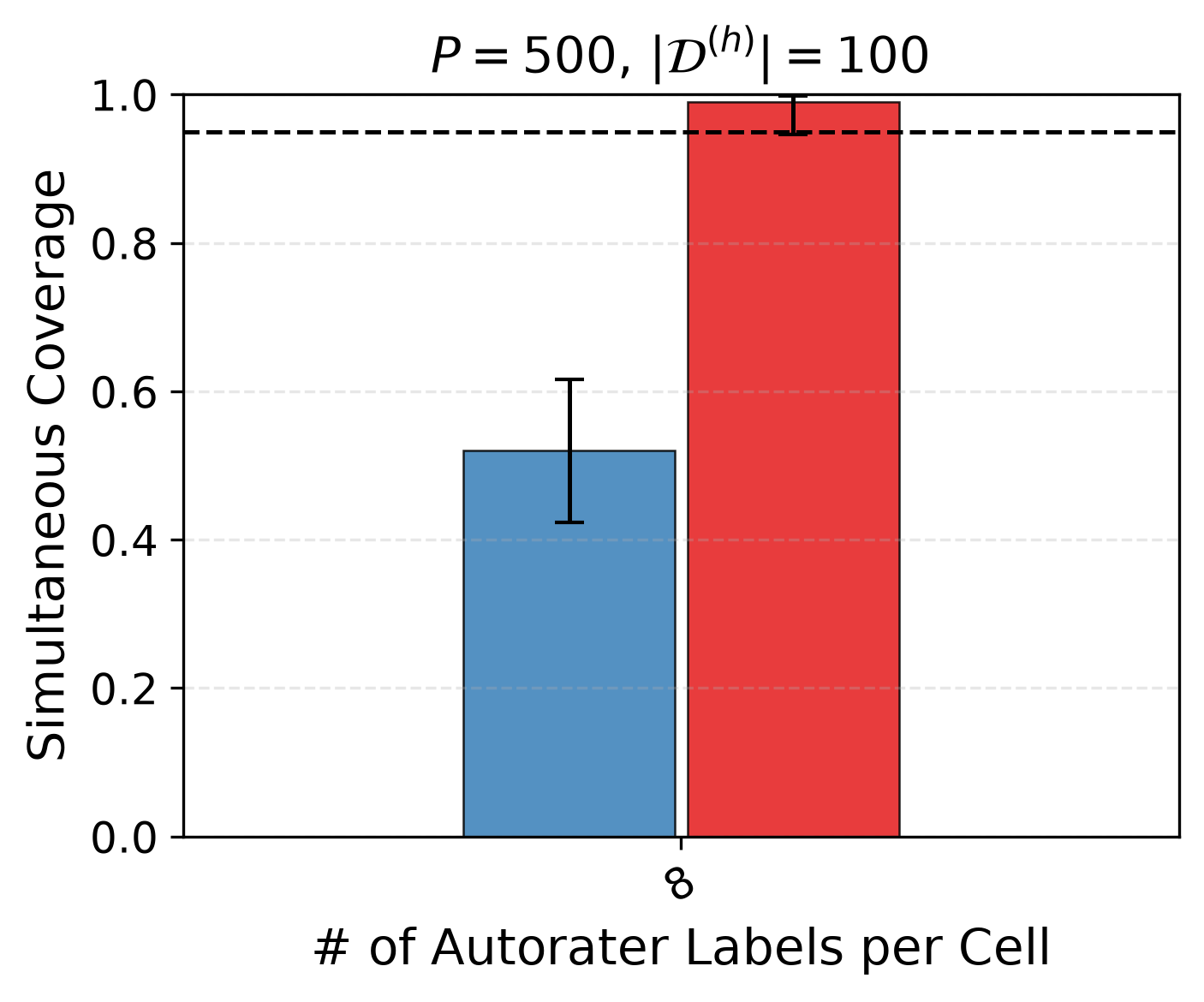}
    \end{subfigure}
    \begin{subfigure}[b]{0.27\textwidth}
        \centering
        \includegraphics[width=\linewidth]{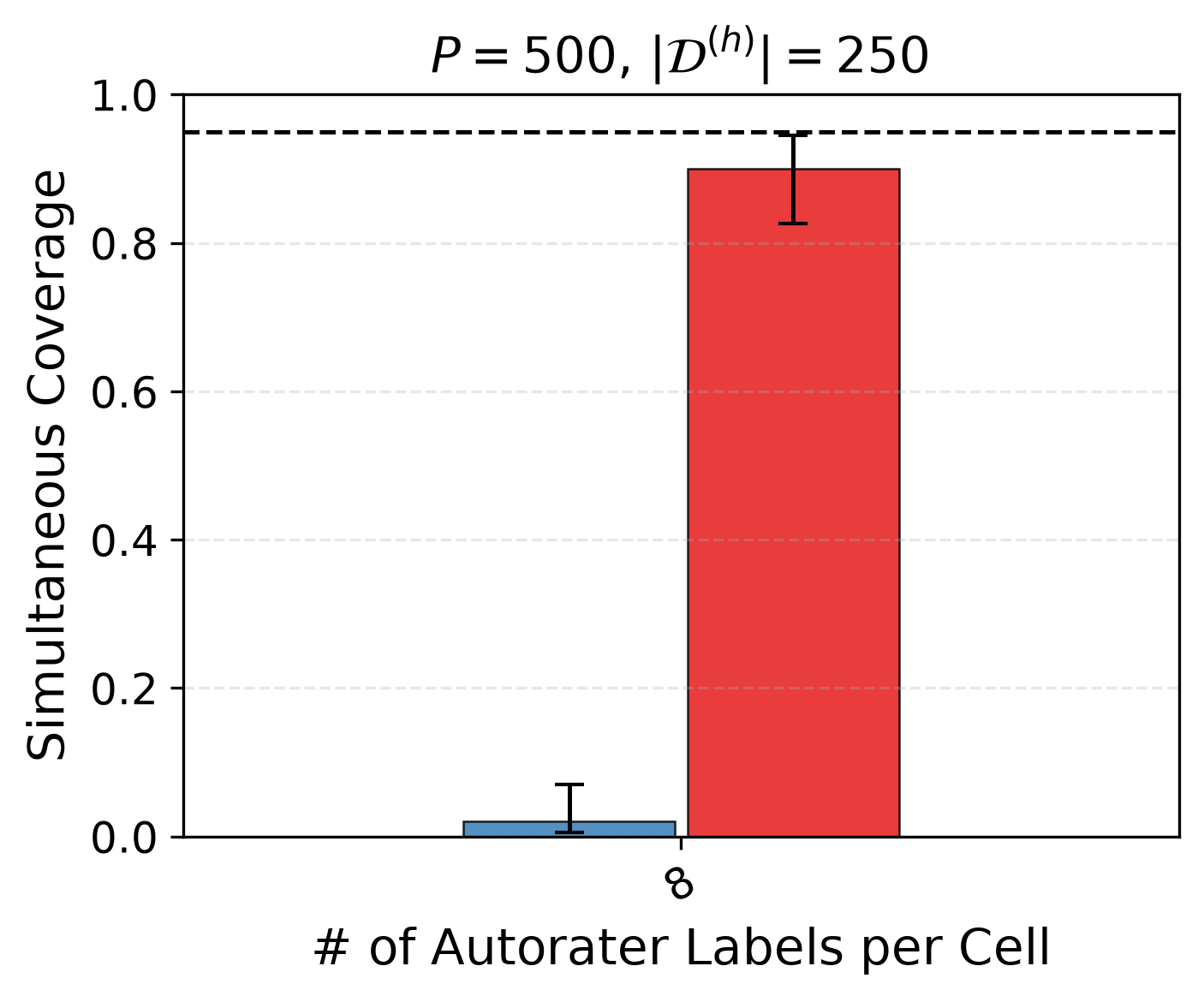}
    \end{subfigure}
    \begin{subfigure}[b]{0.38\textwidth}
        \centering
        \includegraphics[width=\linewidth]{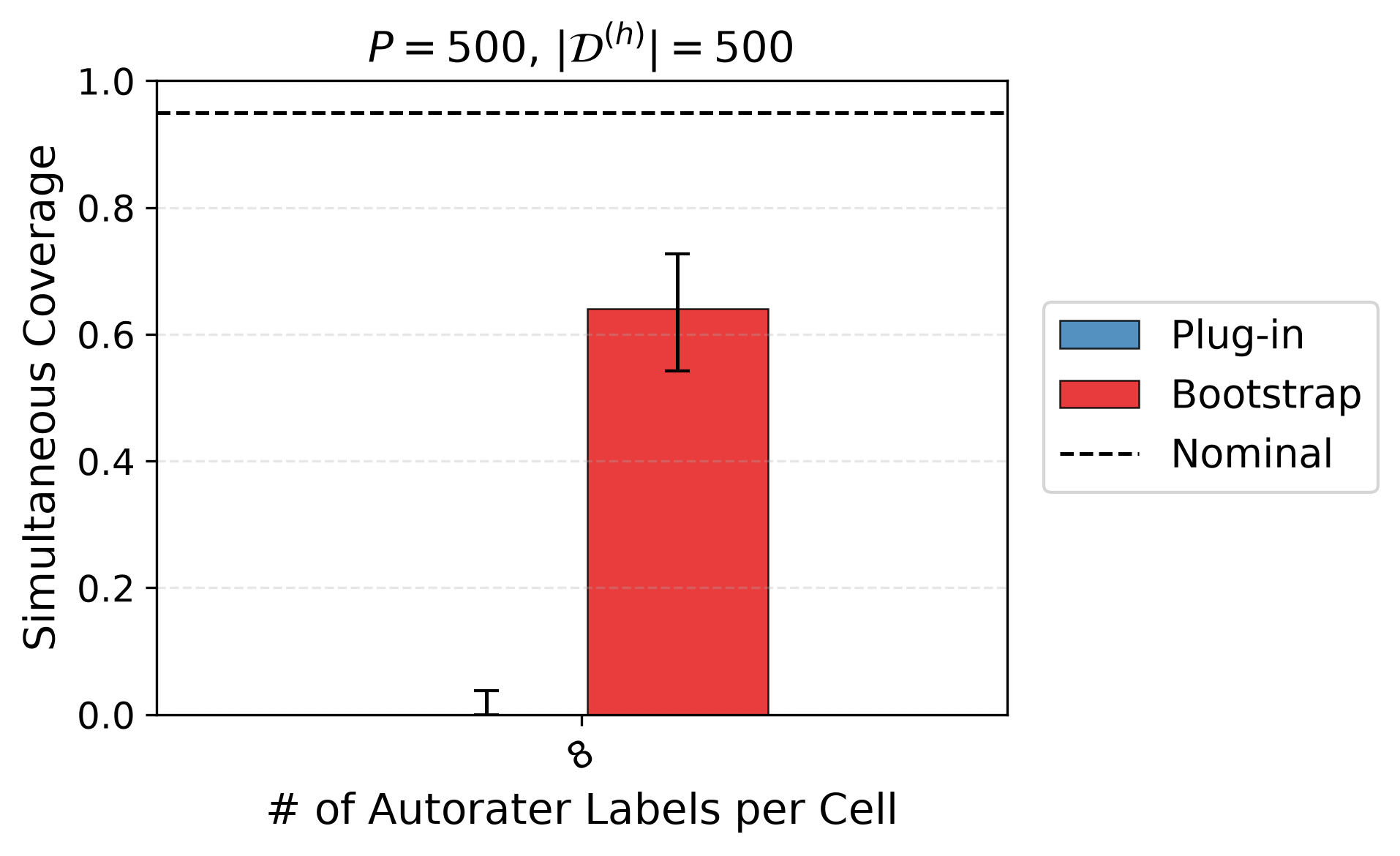}
    \end{subfigure}
    \caption{\small Simultaneous coverage of plug-in intervals and bootstrap bands on the same fitted models, one panel per $(P, |\cD^{(h)}|)$, with $n_{\text{reps}}=8$ and $95\%$ Wilson intervals.}
    \label{fig:sim-boot-sim}
\end{figure}

\begin{figure}[H]
    \centering
    \begin{subfigure}[b]{0.35\textwidth}
        \centering
        \includegraphics[width=\linewidth]{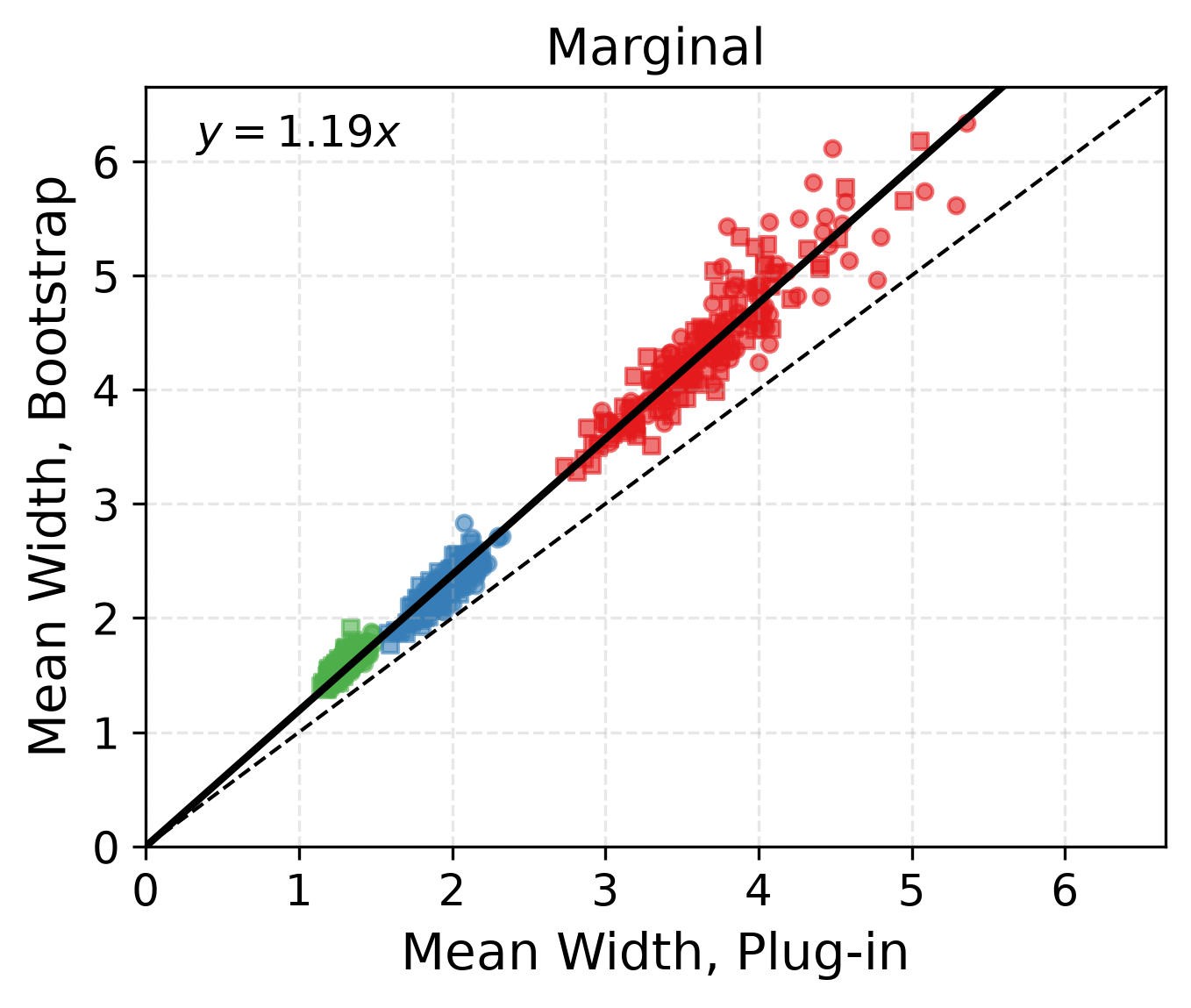}
    \end{subfigure}
    \begin{subfigure}[b]{0.53\textwidth}
        \centering
        \includegraphics[width=\linewidth]{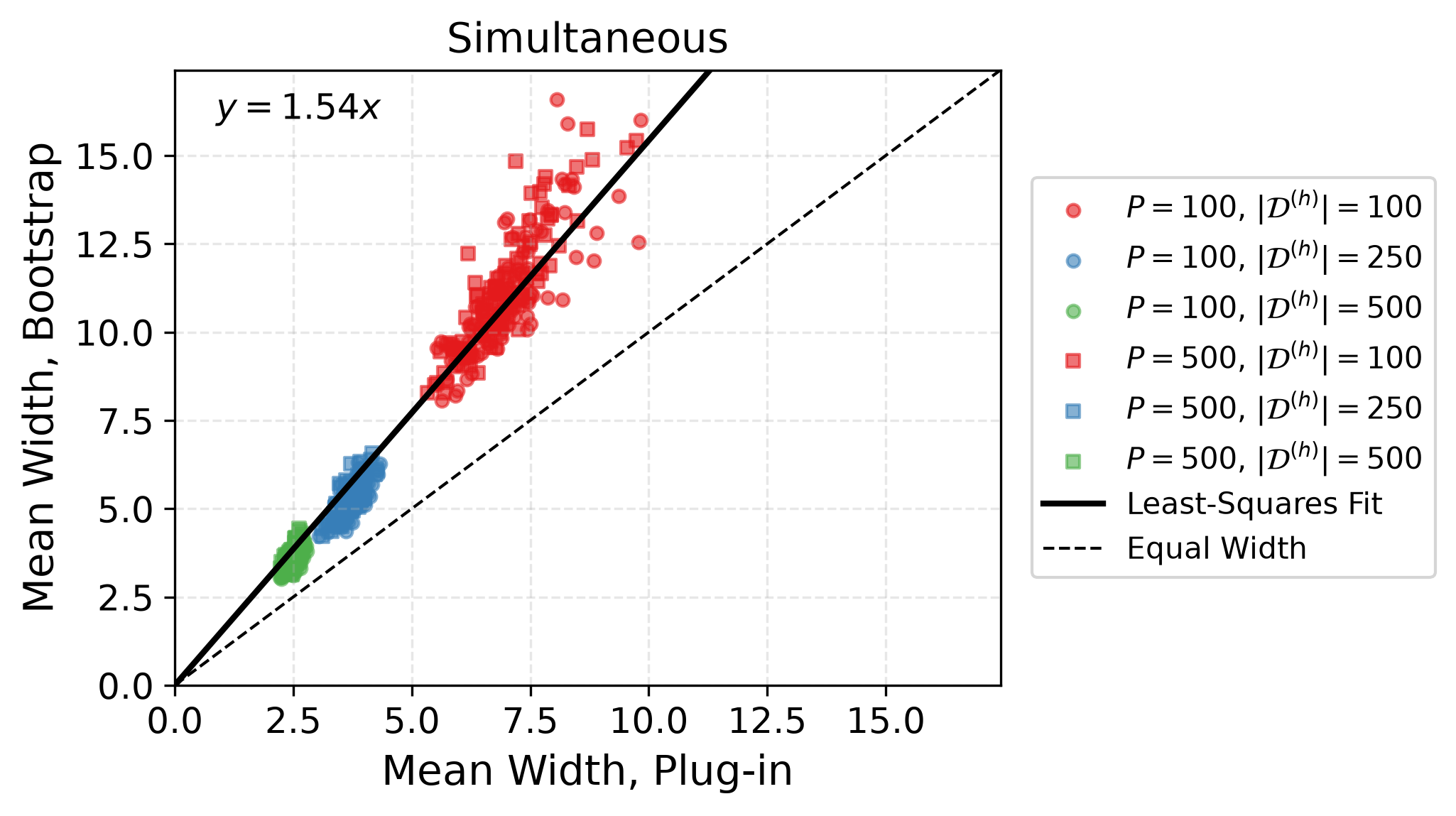}
    \end{subfigure}
    \caption{\small Width of $95\%$ bootstrap intervals against the width of plug-in intervals on the same fitted models, for marginal intervals (left) and simultaneous bands (right), with $n_{\text{reps}}=8$. Each point is one simulated dataset, and its coordinates are the widths (logit scale) averaged over all pairs $(m,p)$; colors give $|\cD^{(h)}|$ and markers give $P$. The solid line is a least-squares fit through the origin over all datasets and settings, and its slope ($1.19$ marginal, $1.54$ simultaneous) is the typical factor by which the bootstrap widens the plug-in interval. The dashed line marks equal widths.}
    \label{fig:sim-widths}
\end{figure}

\textbf{Results.} Simultaneous coverage is far more demanding (Figure \ref{fig:sim-sim}): plug-in simultaneous intervals reach the nominal level only for $P=100$ and $|\cD^{(h)}|=100$, and degrade as $|\cD^{(h)}|$ grows, while the oracle stays at $0.94$ to $0.98$. With $n_{\text{reps}}=8$, the bootstrap band raises coverage from $0.00$--$0.86$ to $0.90$--$0.99$ in five of six settings (Figure \ref{fig:sim-boot-sim}), at the cost of bands $1.4$ to $1.6\times$ wider (pooled multiplier $1.54$; Figure \ref{fig:sim-widths}, right), but remains short ($0.64$) for $P=500$ and $|\cD^{(h)}|=500$.

\section{Model fitting}\label{sec:fit}

\subsection{More details about model fitting}

\textbf{Connections to transfer learning.} The first fitting stage is analogous to pretraining, where we learn robust representations for models ($\Theta$) and prompts ($A$) by exploiting the large volume of autorater data. In the second stage, we transfer these rich features to the downstream task of predicting human labels. By freezing the latent factors, we reduce the optimization problem to learning a simple linear alignment, defined by the human rater embedding $\Gamma_{0,\cdot}$ and cutoffs $\beta^{(0)}$, on top of the pretrained representations. This effectively treats the scarce human data as a supervision signal for calibrating the shared latent space, dramatically reducing sample complexity compared to training from scratch.

\textbf{Fine-tuning (``Ours FT'').} One possible downside of the two-stage fitting procedure is that model/prompt representations ($\Theta$ and $A$) are not fitted using human observations. Therefore, in some cases, an optional third stage can be beneficial for improving empirical accuracy. Starting from the combined estimates $\widehat{\Lambda}=(\widehat{\Lambda}^{(h)},\widehat{\Lambda}^{(a)})$, we further fine-tune all parameters $\Lambda$ by performing gradient descent on the human-rated data objective, $\sum_{D_n \in \cD^{(h)}} \cL(\Lambda,D_n)$. To prevent overfitting, we use a small learning rate and hold out a part of $\cD^{(h)}$ as a validation set to apply early stopping. We perform this step using a variant of Algorithm \ref{alg:stage1}: starting from the parameter estimates obtained in earlier stages, we fine-tune the model on the human-labeled data by updating all weights, with early stopping to limit overfitting. In practice, this refinement rarely harms predictive performance, but it invalidates the standard confidence intervals of Appendix \ref{sec:ci}. Gains are typically larger when at least a few human labels per prompt are available\footnote{See the case of Gecko in Figure~\ref{fig:varying_B}.}, enabling meaningful updates to prompt representations. Accordingly, we recommend fine-tuning when human annotations cover a non-negligible fraction of prompts and point prediction accuracy is the primary objective, and we discourage it when prompt-level human data are extremely sparse or when valid, easily interpretable uncertainty quantification is required for downstream inference.

\textbf{Uniqueness of the second stage.}\label{sec:unique} Conditional on $\widehat{\Lambda}^{(a)}$, the second stage is an ordinal logistic regression with covariates $v_{m,p}$ (or $v_{m_1,p}-v_{m_0,p}$ for side-by-side templates) and parameters $\Lambda^{(h)}=(\gamma_0,\beta^{(0)})$. Its log-likelihood is concave, and strictly concave when the design matrix built from $\cD^{(h)}$, augmented with a constant column absorbed by the cutoffs, has full column rank; $\widehat{\Lambda}^{(h)}$ is then unique whenever it exists, and existence fails only under separation of the human labels or when a response category is never observed. Both conditions are easy to check in practice, mirror conditions for uniqueness in ordinary logistic regression models, and arise only in pathological cases.

\textbf{Choice of decomposition.} We use the CP decomposition because it is the most parsimonious three-way factorization and has the strongest uniqueness theory (Appendix \ref{sec:ident}); alternative parameterizations are compatible with our two-stage procedure, at the cost of more parameters and weaker identifiability.

\subsection{Fitting algorithm}

Algorithm \ref{alg:stage1} gives a description for the first stage fitting process. We parameterize the cutoffs $\beta$ using their differences ($\delta$) to ensure monotonicity. Moreover, while it does reduce the model's flexibility, we found that normalizing the columns of $\Theta$ and $A$ during optimization gives better results.

\begin{algorithm}[h]
 \textbf{Input:} (i) Autorater dataset $\mathcal{D}^{(a)}$, (ii) Adam hyperparameters, (iii) batch size $B$, (iv) epochs $T_a$, (v) positivity threshold $\epsilon > 0$;
 
 \textbf{Output:} Estimated autorater parameters $\widehat{\Lambda}^{(a)}$ (containing weights and cutoff gaps $\delta$);
 \smallskip

 \begin{algorithmic}[1]
 \State Initialize parameters $\Lambda^{(a)}$ randomly.
 \State Initialize \text{Adam\_State}.
 \For{$t = 1 \dots T_a$}
    \State Shuffle $\mathcal{D}^{(a)}$.
    \For{each mini-batch $\mathcal{B} \subset \mathcal{D}^{(a)}$ of size $B$}
        \State \textbf{1. Forward Pass (Construct Cutoffs):}
        \State \textit{Construct ordered cutoffs $\beta$ from learnable gaps $\delta$:}
        \State $\displaystyle \beta^{(r)}_y \leftarrow \beta^{(r)}_{1} + \sum_{k=2}^y \delta^{(r)}_k$
        
        \State \textbf{2. Optimization Step (Adam):}
        \State Compute gradient of NLL w.r.t $\Lambda^{(a)}$:
        \State $\displaystyle g \leftarrow \nabla_{\Lambda^{(a)}} \sum_{D_n \in \mathcal{B}} \mathcal{L}(\Lambda, D_n)$
        \State Update parameters using Adam rule:
        \State $\displaystyle (\Lambda^{(a)}, \text{Adam\_State}) \leftarrow \text{AdamStep}(\Lambda^{(a)},g,\text{Adam\_State}, \text{Adam\_Hyperparameters})$
        
        \State \textbf{3. Enforce Constraints \& Projections:}
        \State \textit{Positivity for gaps (ensures monotonicity $\beta_y < \beta_{y+1}$):}
        \State $\displaystyle \delta \leftarrow \max(\delta, \epsilon)$
        \State \textit{Column-wise $L_2$ Normalization:}
        \State $\displaystyle A \leftarrow A \cdot \text{diag}(\|A_{\cdot l}\|_2^{-1}), \quad \Theta \leftarrow \Theta \cdot \text{diag}(\|\Theta_{\cdot l}\|_2^{-1})$
    \EndFor
 \EndFor
 \State \textbf{return} $\widehat{\Lambda}^{(a)} \leftarrow \Lambda^{(a)}$
 \end{algorithmic}
\caption{Stage 1 fitting via Projected Adam}
\label{alg:stage1}
\end{algorithm}

\section{Identifiability}\label{sec:ident}

In this section, we study sufficient conditions for parameter identifiability. Because this type of result is general, we do not need to make a distinction between human raters and autoraters. To establish the identifiability of the model, we show that the map connecting model parameters and data distribution is injective. This is true if $\Lambda$ can be uniquely recovered from the observed ordinal probabilities.

To that end, we start with a practical assumption about the data collection process. It simply states that we have a chance of observing every combination of model(s), prompt, and rater. This condition ensures that asymptotically we have access to the full set of choice probabilities for all model-prompt-rater combinations, allowing us to form the complete system of equations necessary to uniquely solve for the unknown parameters without dealing with missing data.
\begin{condition}[Sampling]\label{cond:samp}
   Every triple $(m,p,r)$ is sampled with positive probability; for side-by-side raters, this includes both orderings $(m_0,m_1)$ and $(m_1,m_0)$ of every pair.
\end{condition}

See that for single-sided raters, whose set we denote $\cR_{\text{point}}$, simultaneous shifts in $\beta^{(r)}_{y}$ and $\Psi_{m,p,r}$ have no effect on the probability distribution. Setting the first cutoff $\beta^{(r)}_{1}=0$ resolves the shift ambiguity inherent in ordinal models.
\begin{condition}[Shifts]\label{cond:shifts}
   Assume $\beta^{(r)}_{1}=0$ if $r\in\cR_{\text{point}}$.
\end{condition}

Now, we propose a block-structured sparsity constraint. This will be useful for dealing with side-by-side raters. Let us define the set of side-by-side (pairwise) raters as $\cR_\text{pair}\subseteq\cR$. 

\begin{condition}[Block-Structured Sparsity]\label{cond:sparsity}
The tensor rank $L$ is partitioned into two sets $\mathcal{L}_1 = \{1, \dots, L_1\}$ and $\mathcal{L}_2 = \{L_1+1, \dots, L\}$. The factor matrices satisfy:
\begin{enumerate}
    \item $\Theta_{0,l} = 0$ for all $l \in \mathcal{L}_1$.
    \item $\Gamma_{r,l} = 0$ for all $l \in \mathcal{L}_2$ and $r \in \cR_\text{pair}$.
\end{enumerate}
\end{condition}
When all raters are side-by-side, this condition is a normalization (it sets model $0$ as the reference); when single-sided and side-by-side raters coexist, it restricts the parameter space, so Theorem \ref{thm:id} gives identifiability within this restricted model.

We introduce a classic condition from tensor analysis \citep{rhodes2010concise,kruskal1977three,kruskal1989rank} that guarantees the uniqueness of the CP decomposition up to permutation of its rank-one components and the scaling of the factor vectors within each component. It ensures that the factor matrices are sufficiently diverse, preventing multiple different decompositions from yielding the same capability tensor $\Psi$.

\begin{definition}[Kruskal rank \citep{rhodes2010concise,kruskal1977three,kruskal1989rank}] The Kruskal rank of a matrix $B$, denoted $R_B$, is the largest number $r$ such that every set of $r$ columns of $B$ is linearly independent.
\end{definition}

\begin{condition}\label{cond:krank}
   The Kruskal ranks of the factor matrices satisfy $R_\Theta+R_A+R_\Gamma \geq 2L+2$.
\end{condition}

The final set of constraints resolves the remaining scaling and permutation ambiguities left over after applying Kruskal's theorem \citep{rhodes2010concise,kruskal1977three,kruskal1989rank}.

\begin{condition}\label{cond:columns}
   The columns of $\Theta$ and $A$ have unit Euclidean norms. Moreover, the columns of $\Gamma$, denoted $\Gamma_{\cdot,l}$, have distinct norms and are ordered such that $\|\Gamma_{\cdot,l}\|_2>\|\Gamma_{\cdot,l+1}\|_2$. Finally, the last elements $\Theta_{M-1,l}$ and $A_{P-1,l}$ are positive for all $l=1, \dots, L$.
\end{condition}

We are ready to present our main result.
\begin{theorem}\label{thm:id}
Under Conditions \ref{cond:samp}-\ref{cond:columns}, the cutoffs $\beta^{(r)}_{y}$, the capability tensor $\Psi$, and the factor matrices $\Theta, A, \Gamma$ are identifiable.
\end{theorem}

\subsection{Proof of Theorem \ref{thm:id}}

We split the proof into two steps, where in the first step we show that the capability tensor $\Psi$ and the cutoffs $\beta^{(r)}$ are uniquely determined by the observed probabilities, and in the second step we demonstrate that the factor matrices $\Theta, A,$ and $\Gamma$ are uniquely recovered from $\Psi$ given the rank and sparsity constraints.

\textbf{Step 1: Identification of $\Psi$ and $\beta$ from categorical probabilities.}
We define $\mu_{m,p,r,y} \triangleq \Pr(Y_{m,p,r} \leq y) = \sigma(\beta^{(r)}_{y+1} - \Delta_{m,p,r})$. Since the logistic CDF $\sigma$ is strictly increasing, its inverse $\sigma^{-1}$ is well-defined.

\begin{itemize}
    \item \textbf{Single-Sided Templates:} Here, $\Delta_{m,p,r} = \Psi_{m,p,r}$. By Condition \ref{cond:shifts}, $\beta^{(r)}_1 = 0$ for these templates. For the first category ($y=0$), we have:
    \[ \sigma^{-1}(\mu_{m,p,r,0}) = \beta^{(r)}_1 - \Psi_{m,p,r} = -\Psi_{m,p,r} \implies \Psi_{m,p,r} = -\sigma^{-1}(\mu_{m,p,r,0}). \]
    With $\Psi_{m,p,r}$ identified, the remaining cutoffs are identified via:
    \[ \beta^{(r)}_{y+1} = \sigma^{-1}(\mu_{m,p,r,y}) + \Psi_{m,p,r} \quad \text{for } y \in \{1, \dots, C_{r}-2\}. \]

    \item \textbf{Side-by-Side Templates ($r\in\cR_\text{pair}$):} Here, $\Delta_{m,p,r} = \Psi_{m_1,p,r} - \Psi_{m_0,p,r}$. Let $m=(m_0, m_1)$ and $\bar{m}=(m_1, m_0)$. Inverting the logistic function, we get:
    \[ \sigma^{-1}(\mu_{m,p,r,y}) = \beta^{(r)}_{y+1} - (\Psi_{m_1,p,r} - \Psi_{m_0,p,r}) \]
    \[ \sigma^{-1}(\mu_{\bar{m},p,r,y}) = \beta^{(r)}_{y+1} - (\Psi_{m_0,p,r} - \Psi_{m_1,p,r}). \]
    Summing these yields:
    \[ \beta^{(r)}_{y+1} = \frac{\sigma^{-1}(\mu_{m,p,r,y}) + \sigma^{-1}(\mu_{\bar{m},p,r,y})}{2}. \]
    Subtracting them identifies the difference $\Delta_{m,p,r} = \Psi_{m_1,p,r} - \Psi_{m_0,p,r}$. To identify the absolute values of $\Psi$, we note that for $r\in\cR_\text{pair}$, the reference model $m=0$ satisfies:
    \[ \Psi_{0,p,r} = \sum_{l=1}^{L_1} \Theta_{0,l} A_{p,l} \Gamma_{r,l} + \sum_{l=L_1+1}^{L} \Theta_{0,l} A_{p,l} \Gamma_{r,l}=0+0=0 \]
    by Condition \ref{cond:sparsity}. Consequently, $\Psi_{m_1,p,r} =$ $\Delta_{(0,m_1), p, r}$ is identified.
\end{itemize}

\textbf{Step 2: Identification of Factor Matrices via Tensor Uniqueness.}
Having identified the tensor $\Psi$, we apply Kruskal's Theorem (Theorem 4a in \cite{kruskal1989rank}). Condition \ref{cond:krank} ($R_\Theta + R_A + R_\Gamma \geq 2L + 2$) guarantees that $\Psi$ has a unique CP decomposition up to a permutation matrix $M_P$ and diagonal scaling matrices $D_1, D_2, D_3$.

\begin{itemize}
    \item \textbf{Eliminating Scaling Ambiguity:} Kruskal's theorem gives $\tilde{\Theta}=\Theta M_P D_1$, $\tilde{A}=A M_P D_2$, and $\tilde{\Gamma}=\Gamma M_P D_3$ with $D_1D_2D_3=I$. Since the columns of $\Theta$ and $\tilde{\Theta}$ have unit norm, the diagonal entries of $D_1$ are $\pm1$, and since the last entries of their columns are positive, $D_1=I$. The same argument gives $D_2=I$, so $D_3=I$.
    \item \textbf{Eliminating Permutation Ambiguity:} Then $\tilde{\Gamma}=\Gamma M_P$. Since the columns of both $\Gamma$ and $\tilde{\Gamma}$ have distinct norms in strictly decreasing order (Condition \ref{cond:columns}), $M_P=I$.
\end{itemize}

Since the unique set of factor matrices $\{\Theta, A, \Gamma\}$ must satisfy the zero patterns defined in Condition \ref{cond:sparsity}, the parameters are uniquely identified. \qed

\section{Experiments}\label{sec:extra_exp}

\subsection{Extra experiments}

\subsubsection{Average score and win-rate prediction}

In this subsection, we demonstrate the predictive power of our methodology by estimating the performance of a generative model without observing any of its human annotations. Specifically, we aim to approximate one of two metrics for a held-out model $m$: either the average score (for single-sided evaluations), defined as $\text{Average score}= \frac{1}{P}\sum_{p=0}^{P-1} Y_{m,p,0}$, or the win rate difference (for side-by-side evaluations) against each other model, defined as $\text{Win rate difference}= \frac{1}{P}\sum_{p=0}^{P-1} (\ones[Y_{m,p,0}=2] -\ones[Y_{m,p,0}=0]),$ where $Y_{m,p,0}=2$ denotes a victory for the held-out model and $Y_{m,p,0}=0$ denotes a victory for the opponent. To approximate these metrics, we replace the observed outcomes with the expected values predicted by our method. In practice, the held-out model's row of $\Theta$ is fitted from its autorater labels while everything else is kept frozen.

\begin{figure}[H]
    \centering
    \begin{subfigure}[b]{0.3\textwidth}
        \centering
        \includegraphics[width=\linewidth]{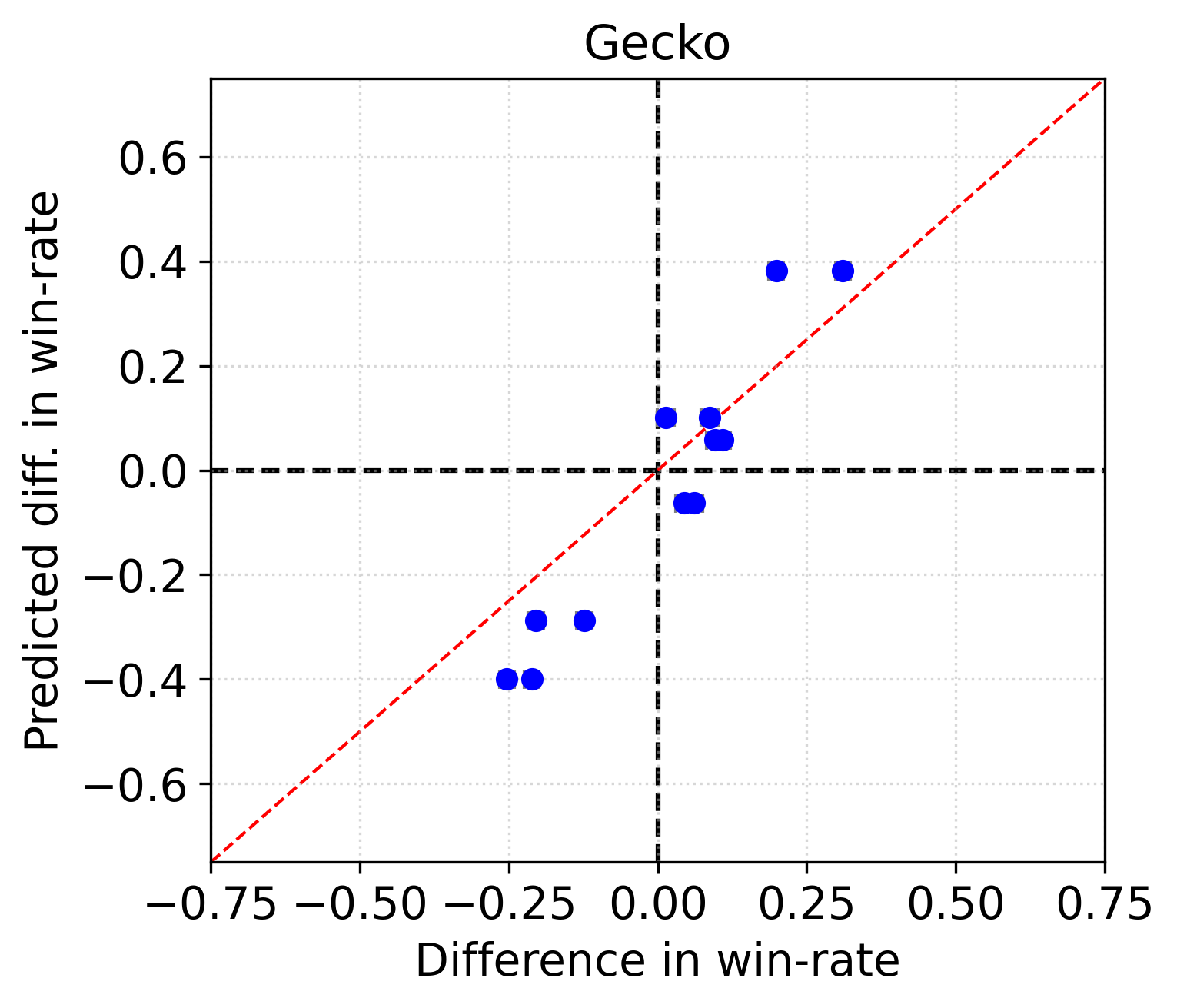}
    \end{subfigure}
    \begin{subfigure}[b]{0.3\textwidth}
        \centering
        \includegraphics[width=.935\linewidth]{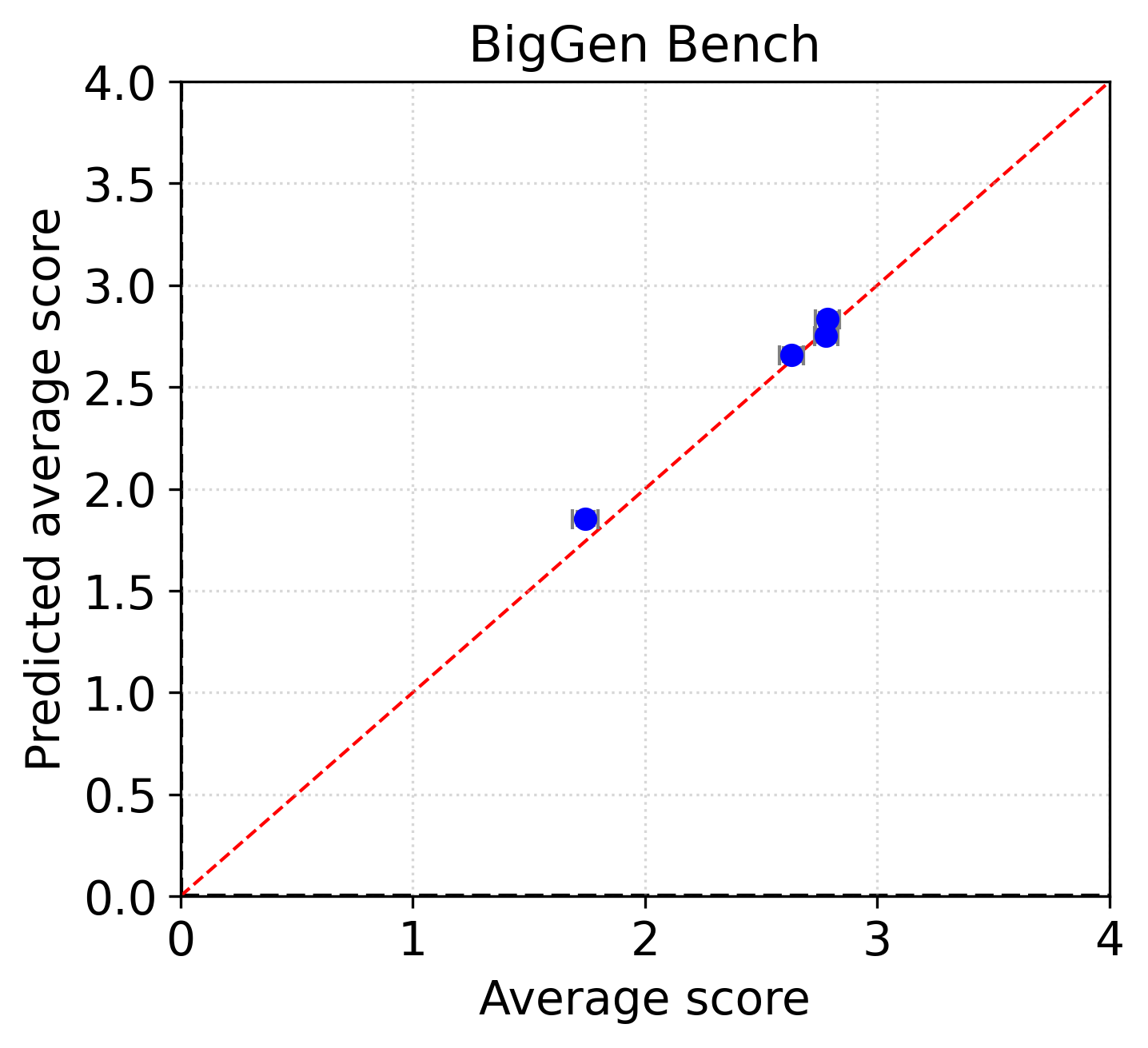}
    \end{subfigure}
    \begin{subfigure}[b]{0.3\textwidth}
        \centering
        \includegraphics[width=\linewidth]{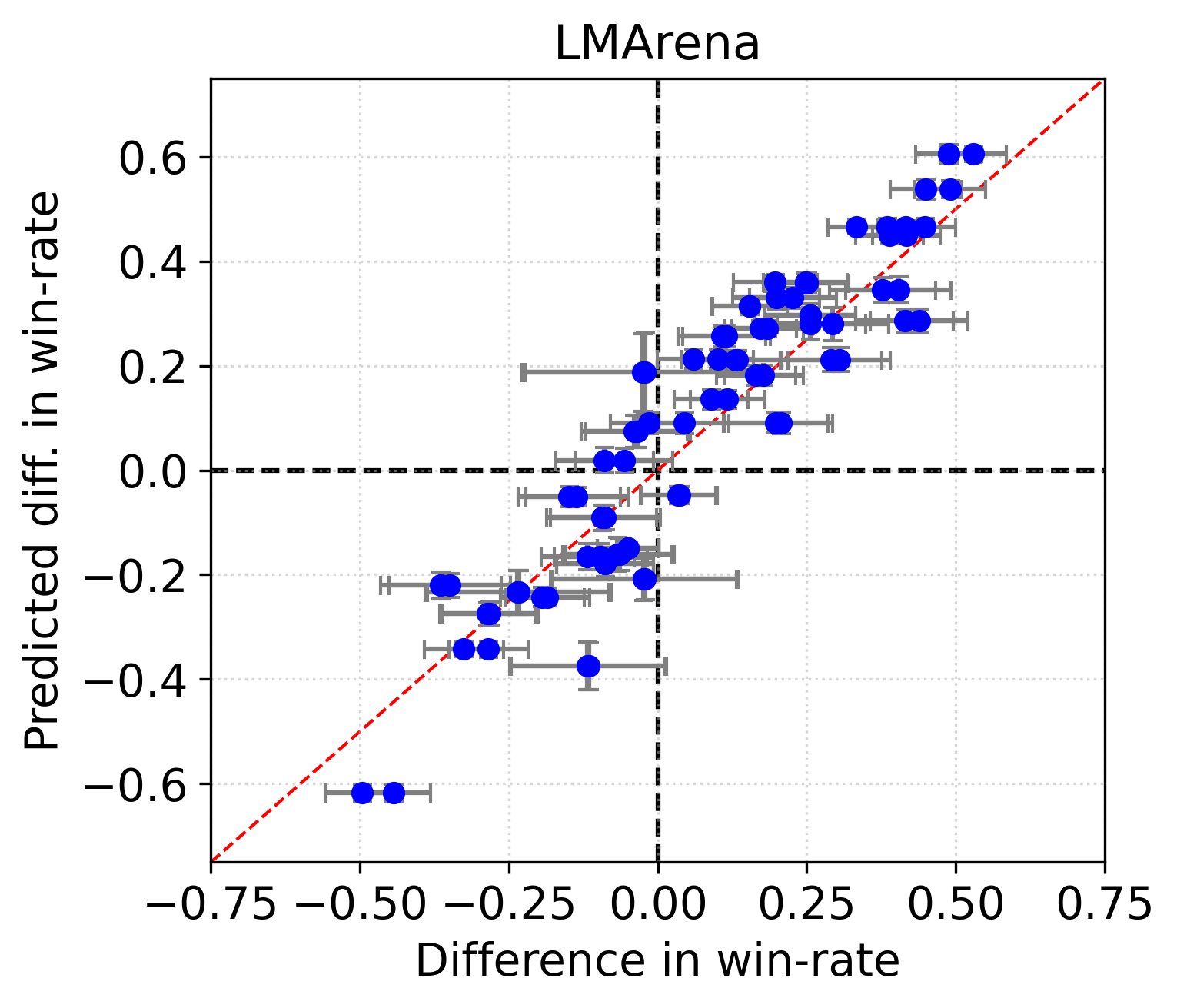}
    \end{subfigure}
    \caption{\small Validation of predictive power on held-out models. The plots compare the ground truth average scores (or win-rate differences) against the values predicted by our method for a model whose human annotations were entirely withheld during training. Results are shown for Gecko (left), BigGen Bench (center), and LMArena (right), indicating strong correlation and preservation of pairwise rankings.}
\label{fig:holdout_model}
\end{figure}

Figure \ref{fig:holdout_model} presents the results of this leave-one-out experiment, plotting the ground truth (x-axis) against our predicted values (y-axis) as we iterate through each generative model. We observe that, across all benchmarks, our methodology yields precise estimates of ground truth performance without training on any human annotations for the held-out model. Notably, for Gecko and LMArena, the majority of data points cluster in the first and third quadrants; this indicates that the sign of the performance metric is correctly predicted, thereby preserving the pairwise ordering of generative models. Standard errors are plotted both in the horizontal and in the vertical directions.

\subsection{Autorater panels with shared biases}\label{sec:shared_bias}
Leniency, severity, and position biases are absorbed by the rater-specific cutoffs $\beta^{(r)}$, so we study shared biases in \emph{what} the judges measure, \ie, in the direction of $\gamma_r$. We use the simulation setup of Appendix \ref{sec:ci-marg} with $P=100$, $n_{\text{reps}}=8$, and $|\cD^{(h)}|\in\{100,400\}$, keeping $\Theta$, $A$, $\gamma_0$, and the human labels fixed so that only the judge panel changes. All 20 judges of a panel sit at the same angle $\alpha$ from the human,
\[
\gamma_r \propto \cos(\alpha)\,\hat{\gamma}_0+\sin(\alpha)\,w_r,\qquad w_r=\frac{w^\ast+\sigma z_r}{\|w^\ast+\sigma z_r\|},
\]
where $\hat{\gamma}_0=\gamma_0/\|\gamma_0\|$, $w^\ast$ is a random unit vector orthogonal to $\gamma_0$, and $z_r$ are Gaussian vectors projected onto the same orthogonal complement. The angle $\alpha\in\{0^\circ,45^\circ,90^\circ,135^\circ,180^\circ\}$ sets the shared bias, and $\sigma\in\{0.05,0.15,0.4,1\}$ sets the diversity of the panel, from near-copies to nearly independent judges. A panel of i.i.d. Gaussian judges serves as reference.

\begin{figure}[H]
    \centering
    \begin{subfigure}[b]{0.32\textwidth}
        \centering
        \includegraphics[width=\linewidth]{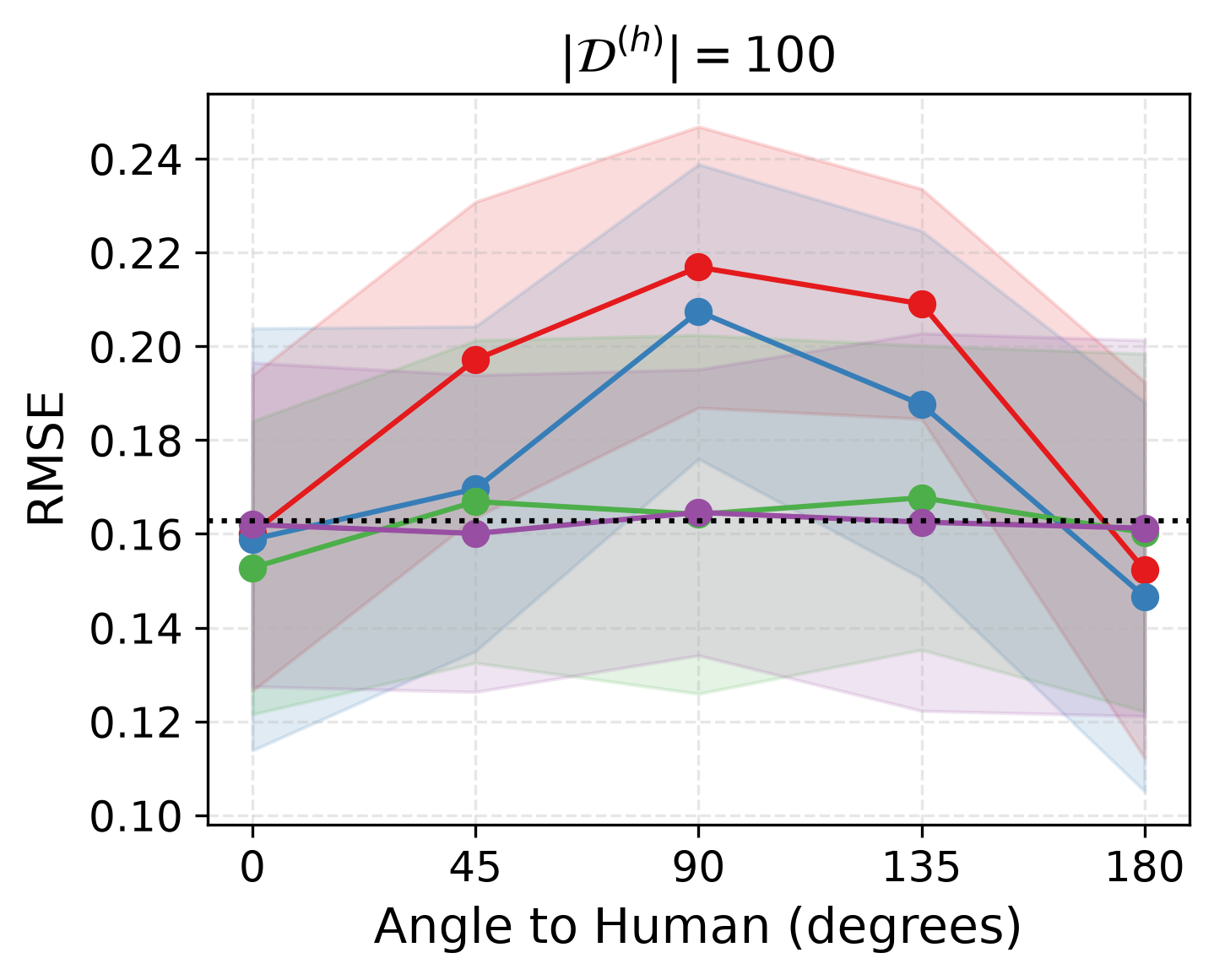}
    \end{subfigure}
    \begin{subfigure}[b]{0.53\textwidth}
        \centering
        \includegraphics[width=\linewidth]{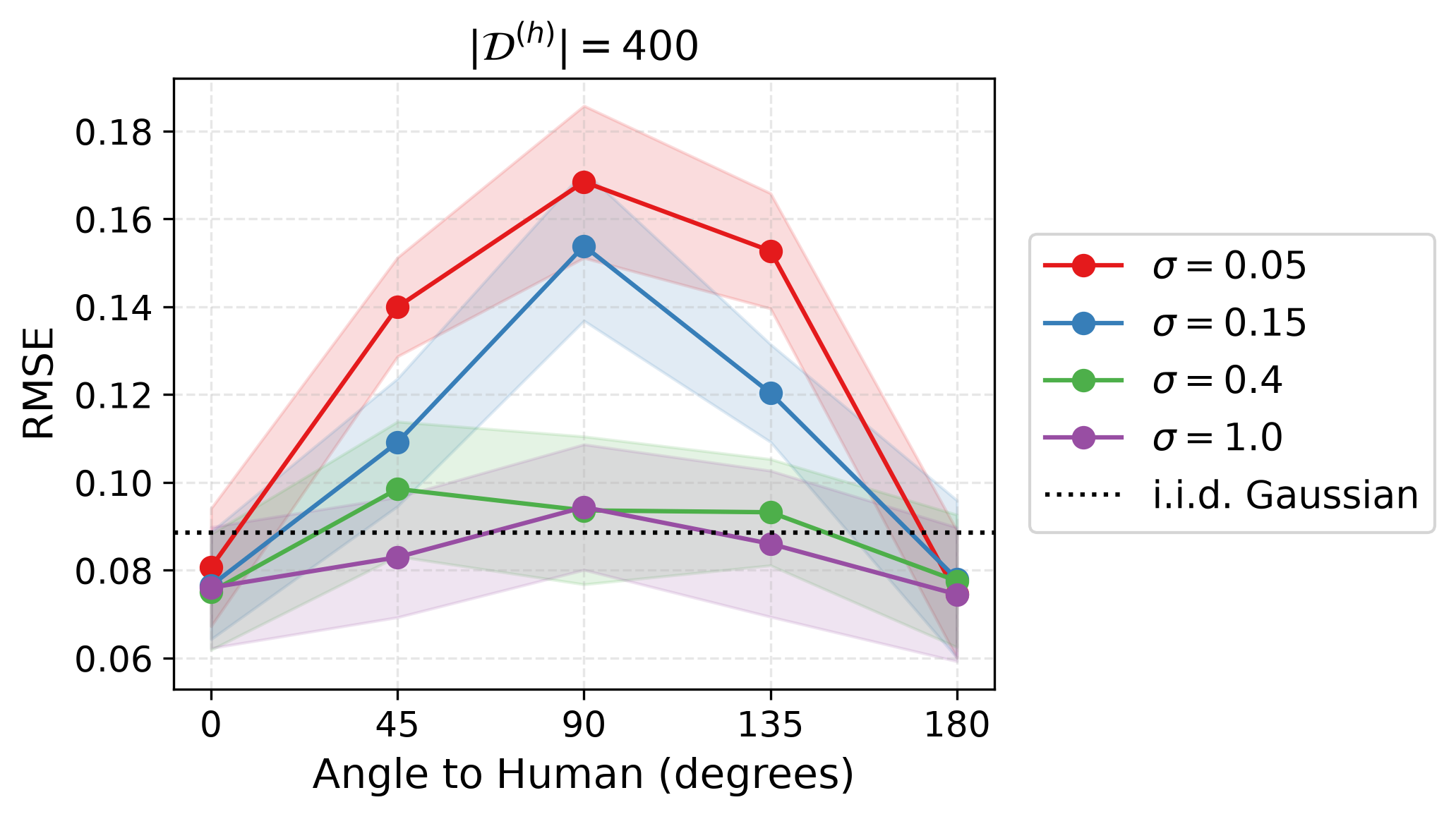}
    \end{subfigure}
    \caption{\small Error in the estimated human probabilities for judge panels sharing one bias direction, by angle to the human and panel diversity $\sigma$, for $100$ (left) and $400$ (right) human labels (mean $\pm$ sd over $48$ repetitions). Dotted line: i.i.d. Gaussian judges.}
    \label{fig:shared-bias}
\end{figure}

Figure \ref{fig:shared-bias} reports the error over $48$ repetitions. Diverse panels ($\sigma\geq0.4$) match the reference at every angle, including judges orthogonal or opposed to the human: the shared representations remain identifiable, and the second stage learns the human readout, including its sign. The error grows only when the panel is both redundant ($\sigma\leq0.15$) and misaligned with the human ($45^\circ$ to $135^\circ$), reaching about twice the reference error at $90^\circ$ with $400$ human labels.

\subsection{Benchmark subset cohesion}\label{sec:cohesion}
In this subsection, we aim to better understand (i) whether the values in Figure \ref{fig:cohesion} could be attributed to chance, and (ii) how to characterize when a group of prompts is cohesive. 

In Table \ref{tab:pval}, we address point (i) by performing a permutation hypothesis test of independence. We test whether the group assignment is independent of the representations $\alpha_p$. Our test statistic is the cohesion measure itself, as we aim to determine if a statistical dependence between group ID and prompt embeddings is manifested by the cohesion of prompts within groups. Table \ref{tab:pval} presents the p-values; the results (low p-values for cohesive groups versus higher p-values for non-cohesive groups) provide evidence that the patterns observed in Figure \ref{fig:cohesion} are not due to chance.

Table \ref{tab:num_prompts} displays the number of prompts for each group. A clear finding from Gecko is that larger, more generic groups (\eg, ``landmarks'') are typically less cohesive. This intuitive result suggests that, in such cases, practitioners could further decompose these larger groups into more cohesive subgroups. In BGB, all groups share the same size, showing that even a small group can be non-cohesive.

\begin{table}[h]
\centering
\caption{P-values for Gecko and BGB across Cohesive and Non-Cohesive tasks}
\begin{tabular}{l|lc}
\hline
\textbf{Category} & \textbf{Task} & \textbf{p-val} \\ \hline
\textit{Gecko - Cohesive} & lang/compositional:compositional-3E-2A & 0.000 \\
 & rel/action:action\_rev/3 & 0.001 \\
 & rel/scale:same\_size & 0.063 \\
 & att/count:additive\_3 & 0.001 \\
 & att/color:abstract & 0.028 \\ \hline
\textit{Gecko - Non-Cohesive} & ne/landmarks:landmarks & 0.194 \\
 & render/text:text\_numerical & 0.585 \\
 & lang/complexity\_nopeople:complexity\_negation & 0.824 \\
 & att/style\_100yr:style\_100yr/visual\_medium & 0.603 \\
 & rel/spatial:spatial\_basic\_rev & 0.416 \\ \hline \hline
\textit{BGB - Cohesive} & determine\_what\_is\_wrong & 0.000 \\
 & reward\_modeling & 0.000 \\
 & interplanetary\_diplomacy & 0.000 \\
 & multi\_step & 0.001 \\
 & time\_traveler\_dilemma & 0.000 \\ \hline
\textit{BGB - Non-Cohesive} & thinking\_for\_doing & 0.954 \\
 & simulator & 0.849 \\
 & honesty & 0.732 \\
 & web\_browsing & 0.729 \\
 & faux\_pas\_explanation & 0.709 \\ \hline
\end{tabular}
\label{tab:pval}
\end{table}

\begin{table}[h]
\centering
\caption{Number of Prompts for Gecko and BGB across Cohesive and Non-Cohesive tasks}
\begin{tabular}{l|lc}
\hline
\textbf{Category} & \textbf{Task} & \textbf{\# prompts} \\ \hline
\textit{Gecko - Cohesive} & lang/compositional:compositional-3E-2A & 10 \\
 & rel/action:action\_rev/3 & 8 \\
 & rel/scale:same\_size & 6 \\
 & att/count:additive\_3 & 11 \\
 & att/color:abstract & 8 \\ \hline
\textit{Gecko - Non-Cohesive} & ne/landmarks:landmarks & 100 \\
 & render/text:text\_numerical & 32 \\
 & lang/complexity\_nopeople:complexity\_negation & 19 \\
 & att/style\_100yr:style\_100yr/visual\_medium & 22 \\
 & rel/spatial:spatial\_basic\_rev & 28 \\ \hline \hline
\textit{BGB - Cohesive} & determine\_what\_is\_wrong & 10 \\
 & reward\_modeling & 10 \\
 & interplanetary\_diplomacy & 10 \\
 & multi\_step & 10 \\
 & time\_traveler\_dilemma & 10 \\ \hline
\textit{BGB - Non-Cohesive} & thinking\_for\_doing & 10 \\
 & simulator & 10 \\
 & honesty & 10 \\
 & web\_browsing & 10 \\
 & faux\_pas\_explanation & 10 \\ \hline
\end{tabular}
\label{tab:num_prompts}
\end{table}

\subsection{Explaining the performance gap on LMArena}\label{sec:reg}

Table \ref{tab:ols_capability} explains the gap between LLaMa-3.3-70b and Gemini-2.5-Pro according to some (overlapping) prompt tags present on the LMArena dataset using Ordinary Least Squares (OLS) regression. For example, ``specificity''\footnote{Here, the specificity tag probably indicates how precise and well-defined a prompt is, measuring whether it asks for a concrete, unambiguous response rather than something broad or open-ended. Prompts with high specificity are narrowly scoped and clearly directed, making it easier for a model to generate a focused answer.} gives LLaMa-3.3-70b a relative advantage over Gemini.

\begin{table}[h]
\centering
\caption{Explains the gap between LLaMa-3.3-70b and Gemini-2.5-Pro (OLS)}
\begin{tabular}{lcccc}
\hline
\textbf{Variable} & \textbf{Coef.} & \textbf{Std.Err.} & \textbf{t} & \textbf{P$>|t|$} \\
\hline
const & -0.7713 & 0.037 & -20.818 & 0.000 \\
category\_tag\_criteria\_v0\_1\_specificity & 0.1631 & 0.040 & 4.103 & 0.000 \\
category\_tag\_criteria\_v0\_1\_domain\_knowledge & -0.1058 & 0.044 & -2.427 & 0.015 \\
category\_tag\_criteria\_v0\_1\_complexity & -0.0100 & 0.035 & -0.283 & 0.777 \\
category\_tag\_criteria\_v0\_1\_technical\_accuracy & 0.0032 & 0.039 & 0.082 & 0.935 \\
category\_tag\_criteria\_v0\_1\_real\_world & 0.0001 & 0.039 & 0.003 & 0.997 \\
category\_tag\_criteria\_v0\_1\_problem\_solving & -0.0622 & 0.040 & -1.549 & 0.121 \\
category\_tag\_criteria\_v0\_1\_creativity & 0.0200 & 0.036 & 0.559 & 0.576 \\
category\_tag\_creative\_writing\_v0\_1\_creative\_writing & 0.0285 & 0.055 & 0.519 & 0.604 \\
category\_tag\_math\_v0\_1\_math & 0.0675 & 0.055 & 1.231 & 0.218 \\
category\_tag\_if\_v0\_1\_if & 0.0312 & 0.043 & 0.730 & 0.465 \\
is\_code & 0.1304 & 0.038 & 3.421 & 0.001 \\
\hline
\end{tabular}

\vspace{2mm} 

\begin{tabular}{lc}
\hline
\textbf{Model Summary} &  \\ 
\hline
R-squared & 0.011 \\
Adjusted R-squared & 0.009 \\
F-statistic & 4.799 \\
Prob(F-statistic) & 2.17e-07 \\
Number of observations & 4777 \\
\hline
\end{tabular}

\label{tab:ols_capability}
\end{table}

\subsection{First stage fit diagnostic}
\begin{figure}[H]
    \centering
    \begin{subfigure}[b]{0.27\textwidth}
        \centering
        \includegraphics[width=\linewidth]{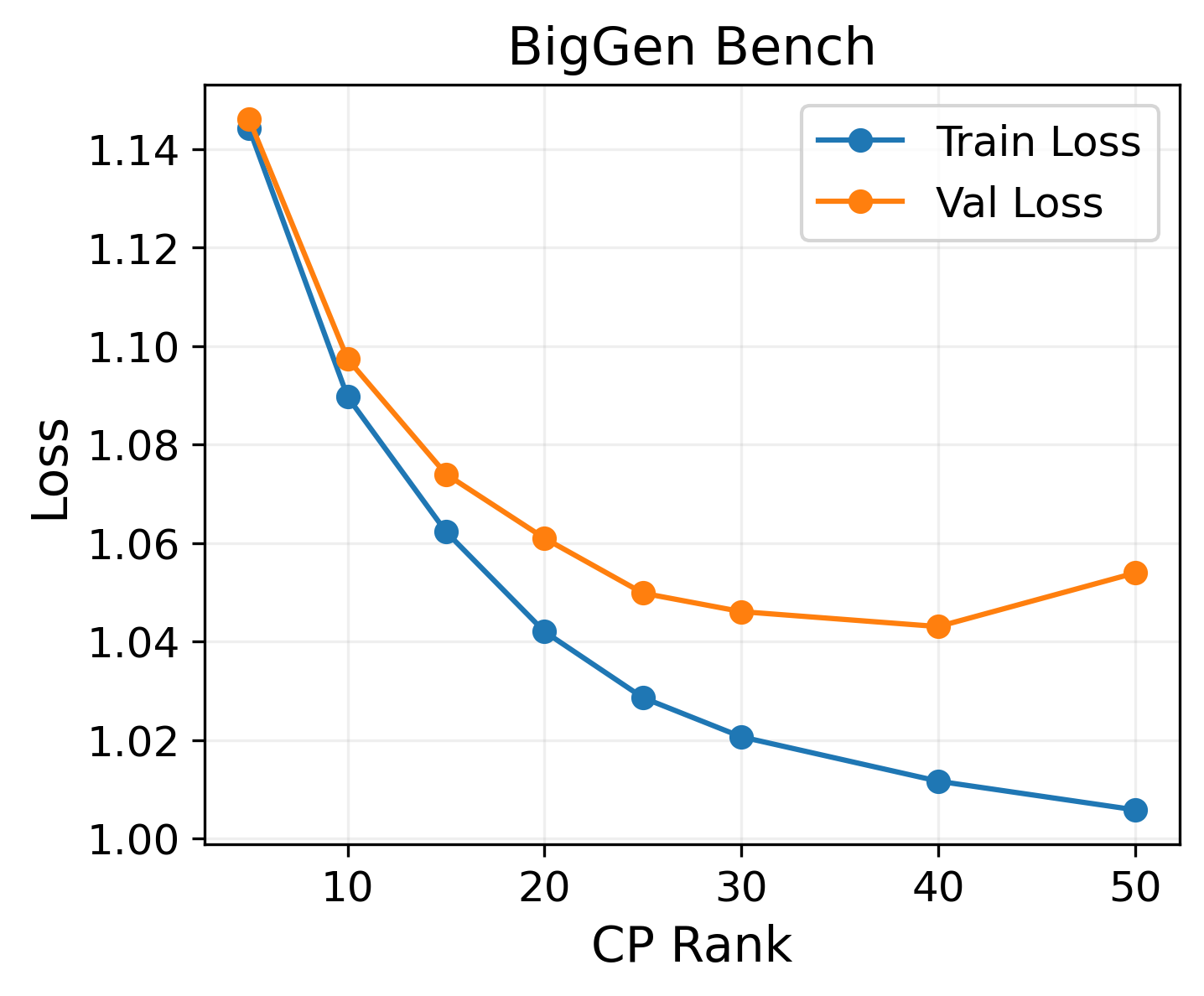}
    \end{subfigure}
    \begin{subfigure}[b]{0.27\textwidth}
        \centering
        \includegraphics[width=\linewidth]{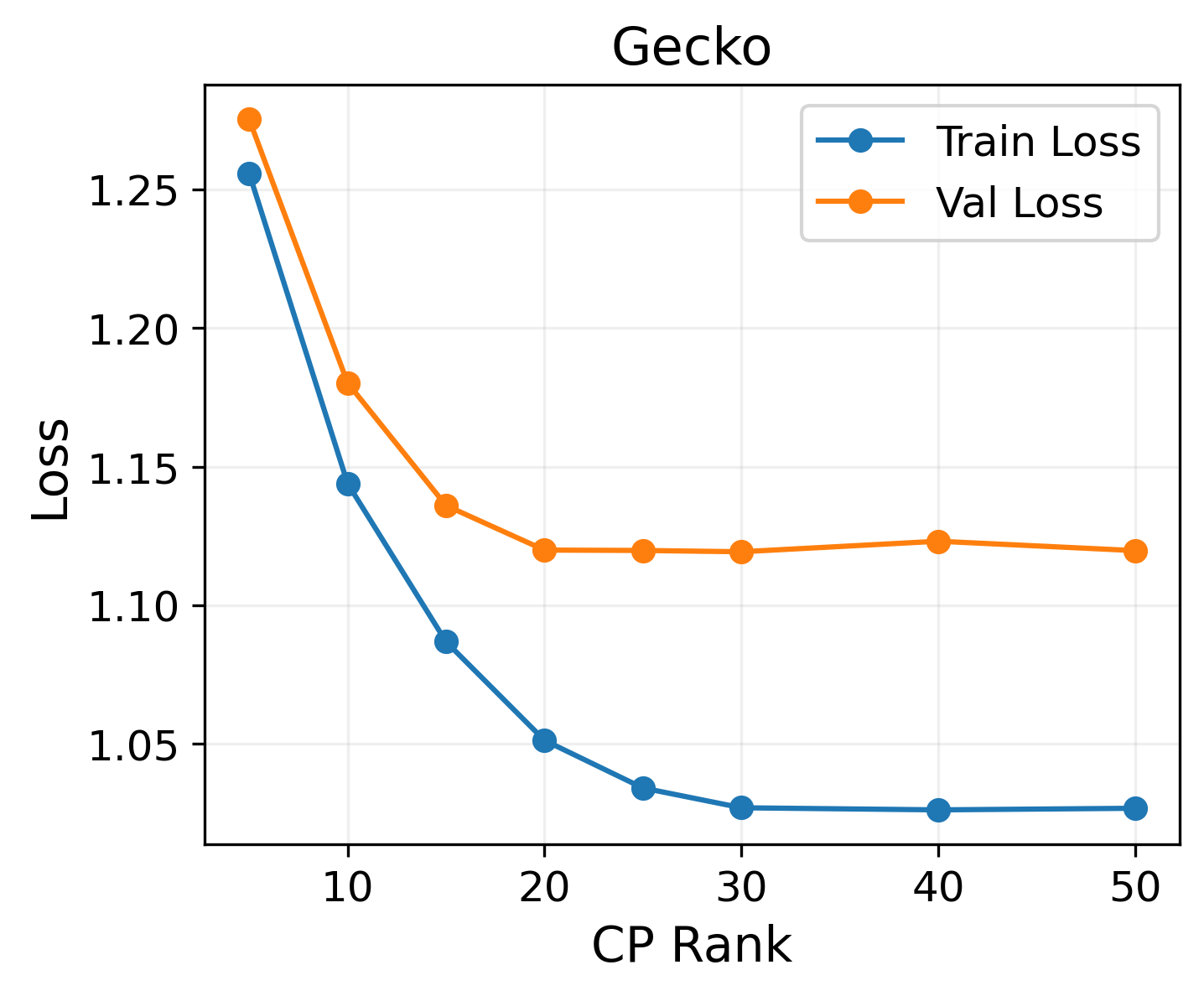}
    \end{subfigure}
    \begin{subfigure}[b]{0.27\textwidth}
        \centering
        \includegraphics[width=\linewidth]{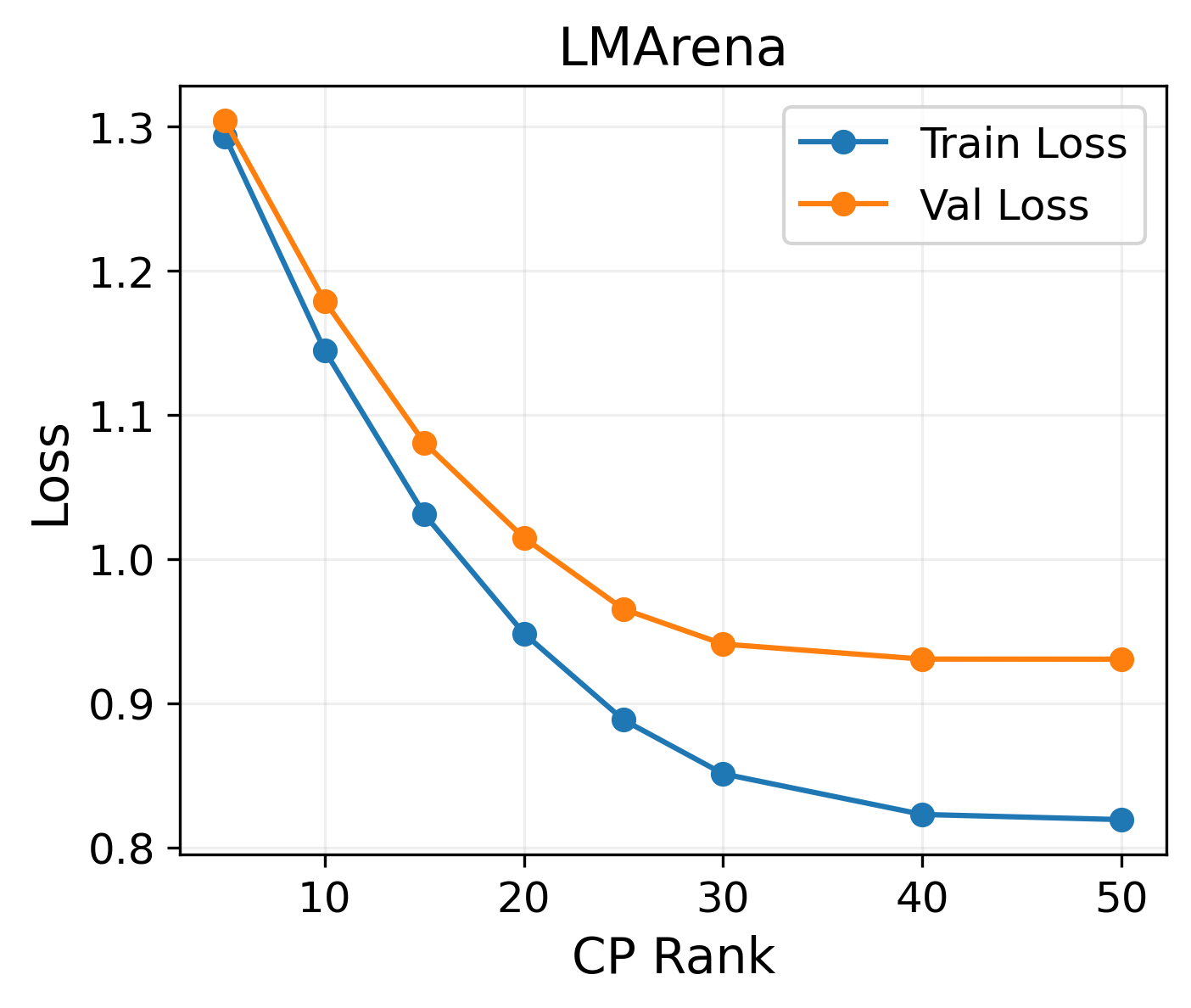}
    \end{subfigure}
    
    \caption{First stage losses.}
    \label{fig:first_stage_losses}
\end{figure}

Figure \ref{fig:first_stage_losses} depicts the first-stage losses, showing the expected pattern of decreasing losses and overfitting when $L$ increases.

\newpage
\subsection{Second stage performance with varying rank}
\begin{figure}[H]
    \centering
    \begin{subfigure}[b]{0.275\textwidth}
        \centering
        \includegraphics[width=\linewidth]{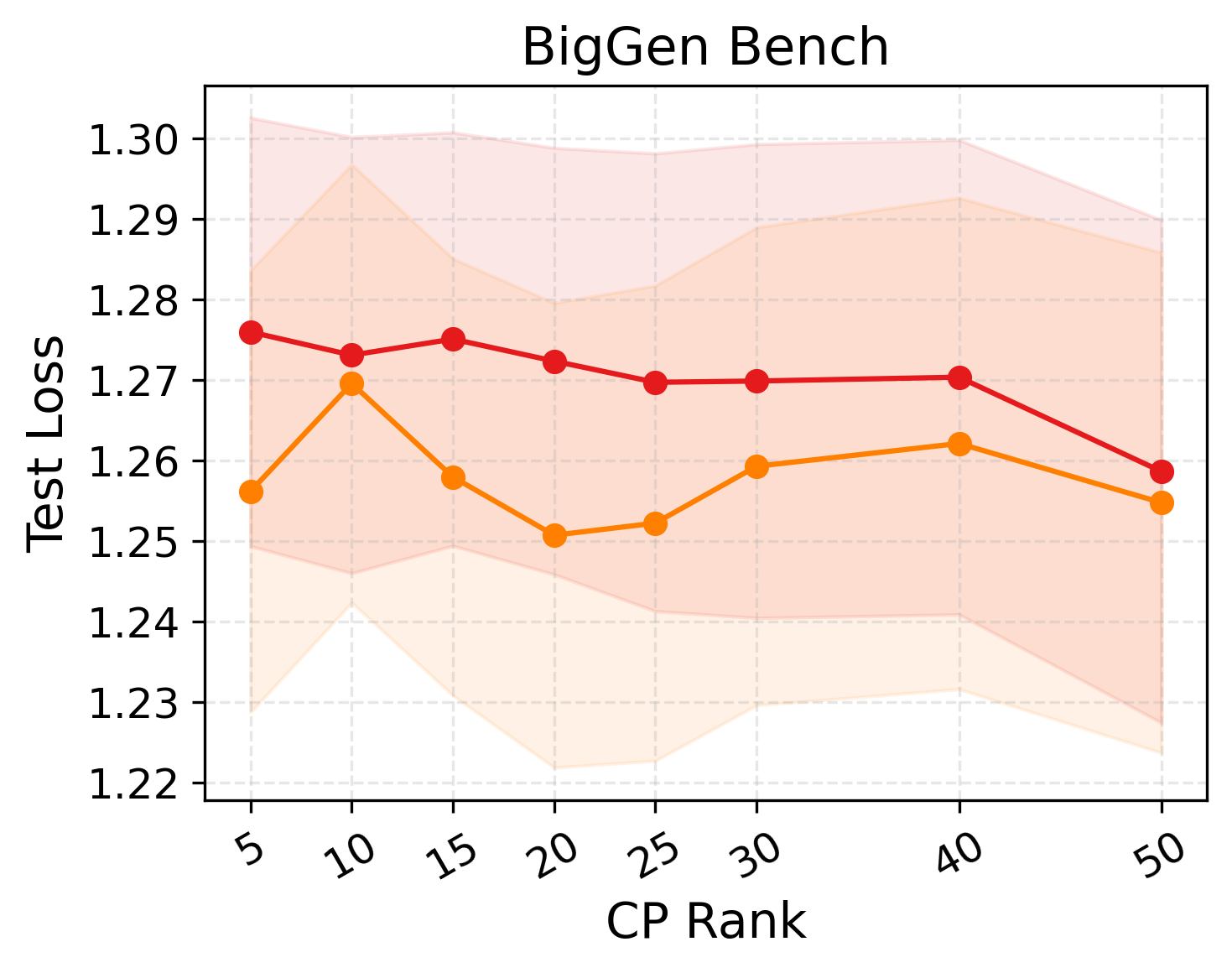}
    \end{subfigure}
    \begin{subfigure}[b]{0.275\textwidth}
        \centering
        \includegraphics[width=\linewidth]{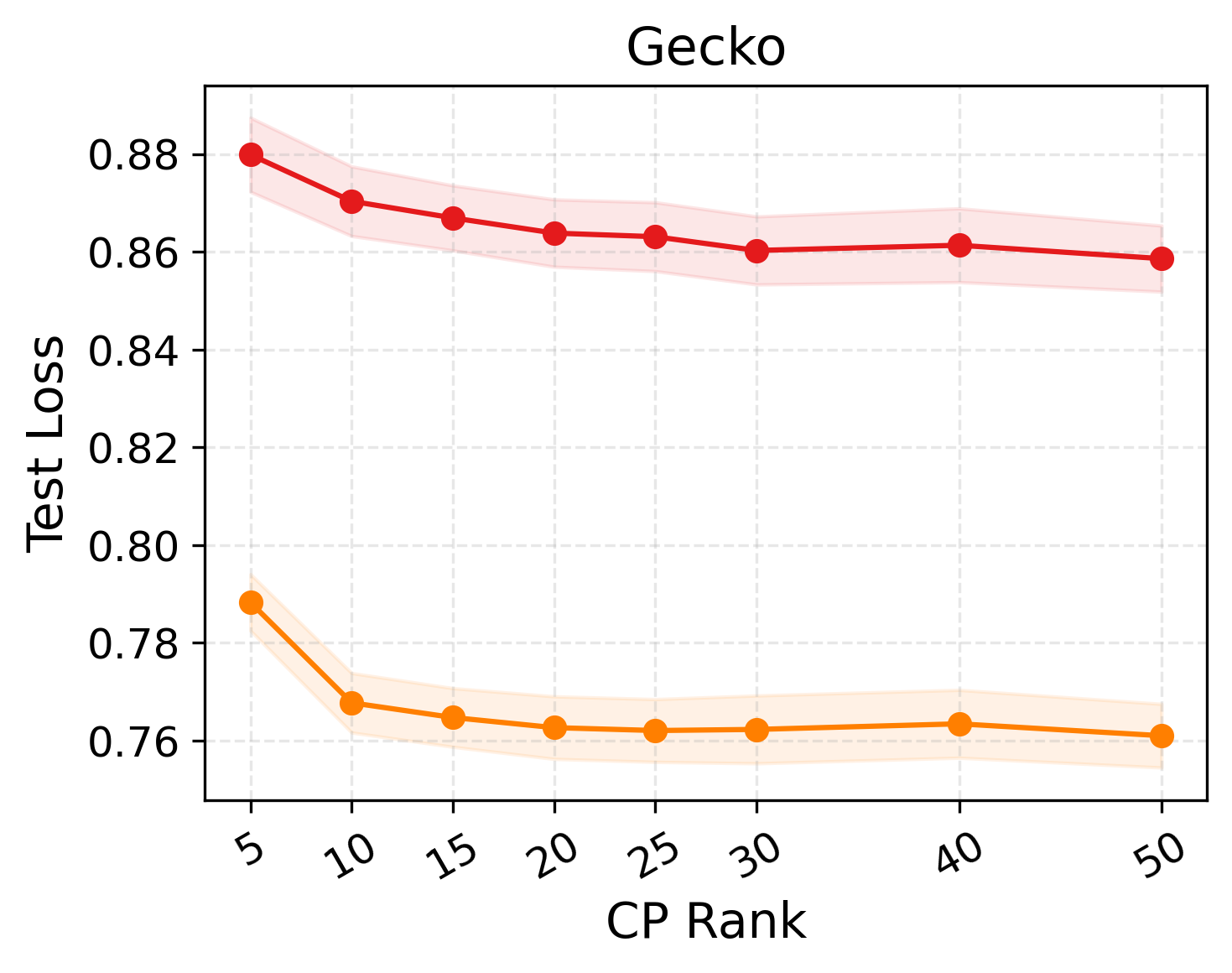}
    \end{subfigure}
    \begin{subfigure}[b]{0.36\textwidth}
        \centering
        \includegraphics[width=\linewidth]{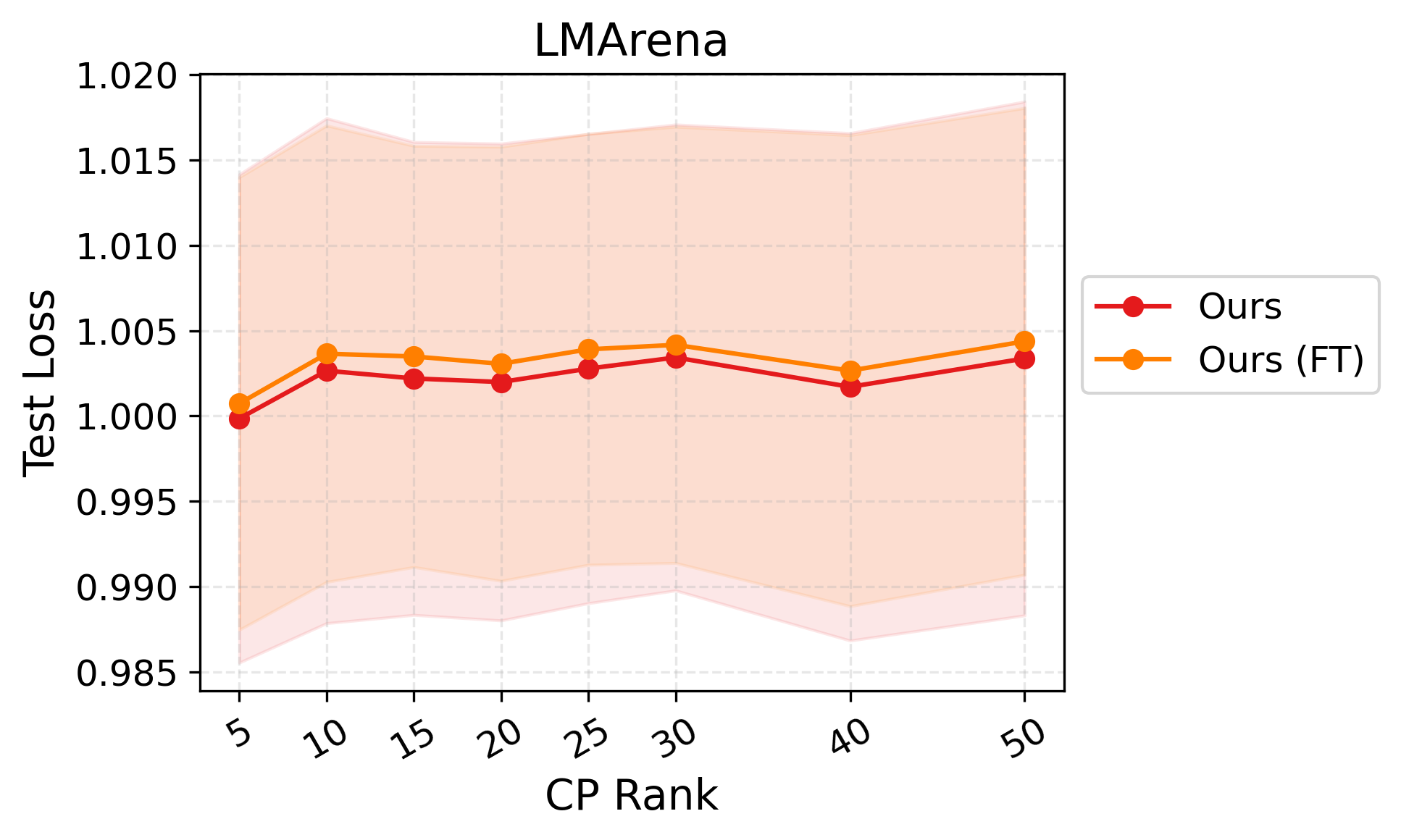}
    \end{subfigure}
    
    \caption{Varying the rank $L$ and assessing impact on predictive performance. We keep $80\%$ of the dataset for training and the rest for testing.}
    \label{fig:varying_rank}
\end{figure}

Figure \ref{fig:varying_rank} shows how $L$ impacts the second-stage losses. This plot shows that first-stage and second-stage losses are only correlated up to a certain point (Figure \ref{fig:first_stage_losses}).

\subsection{Second stage with varying number of human annotations using autoraters as baselines}

\begin{figure}[H]
    \centering
    \begin{subfigure}[b]{0.275\textwidth}
        \centering
        \includegraphics[width=\linewidth]{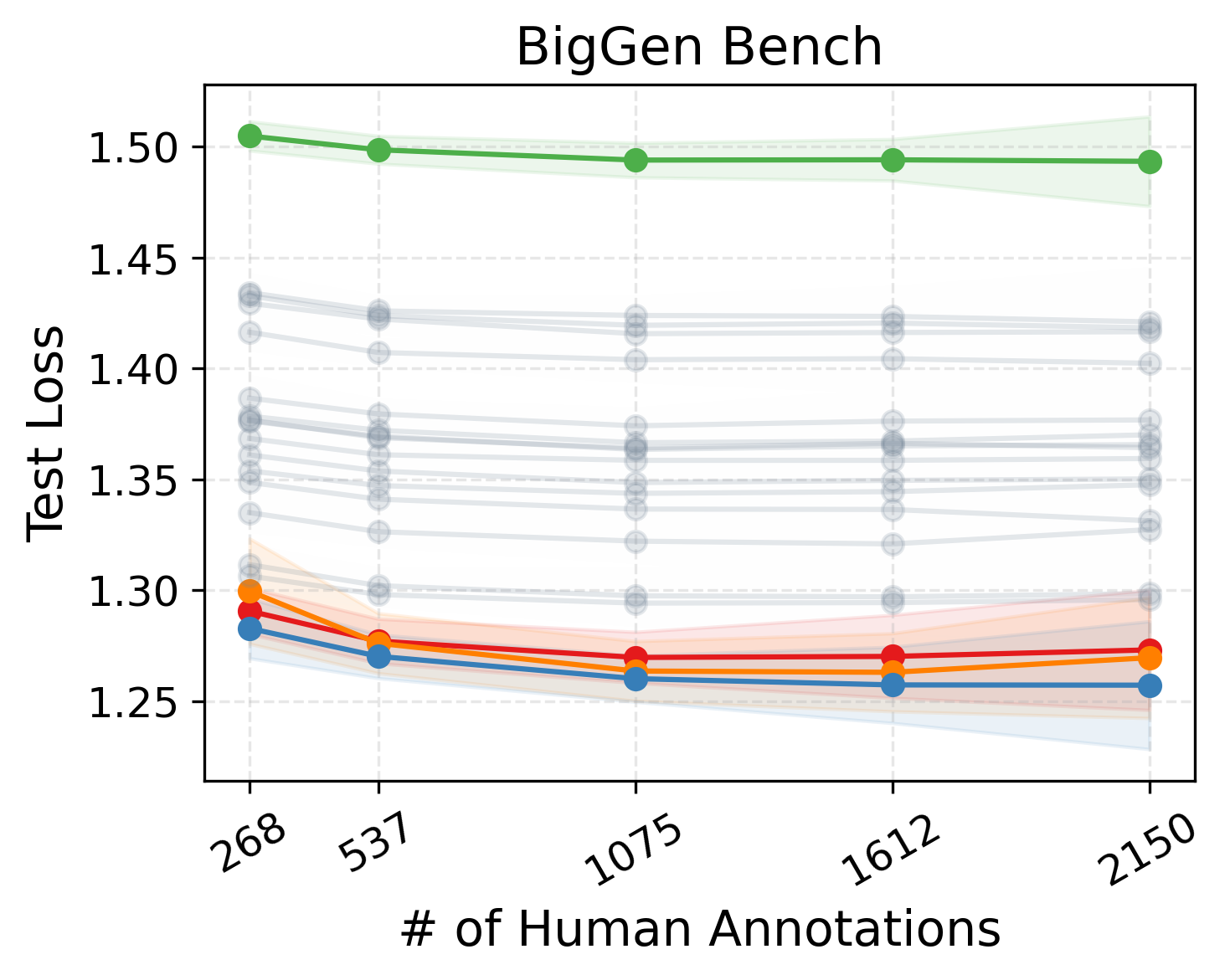}
    \end{subfigure}
    \begin{subfigure}[b]{0.275\textwidth}
        \centering
        \includegraphics[width=\linewidth]{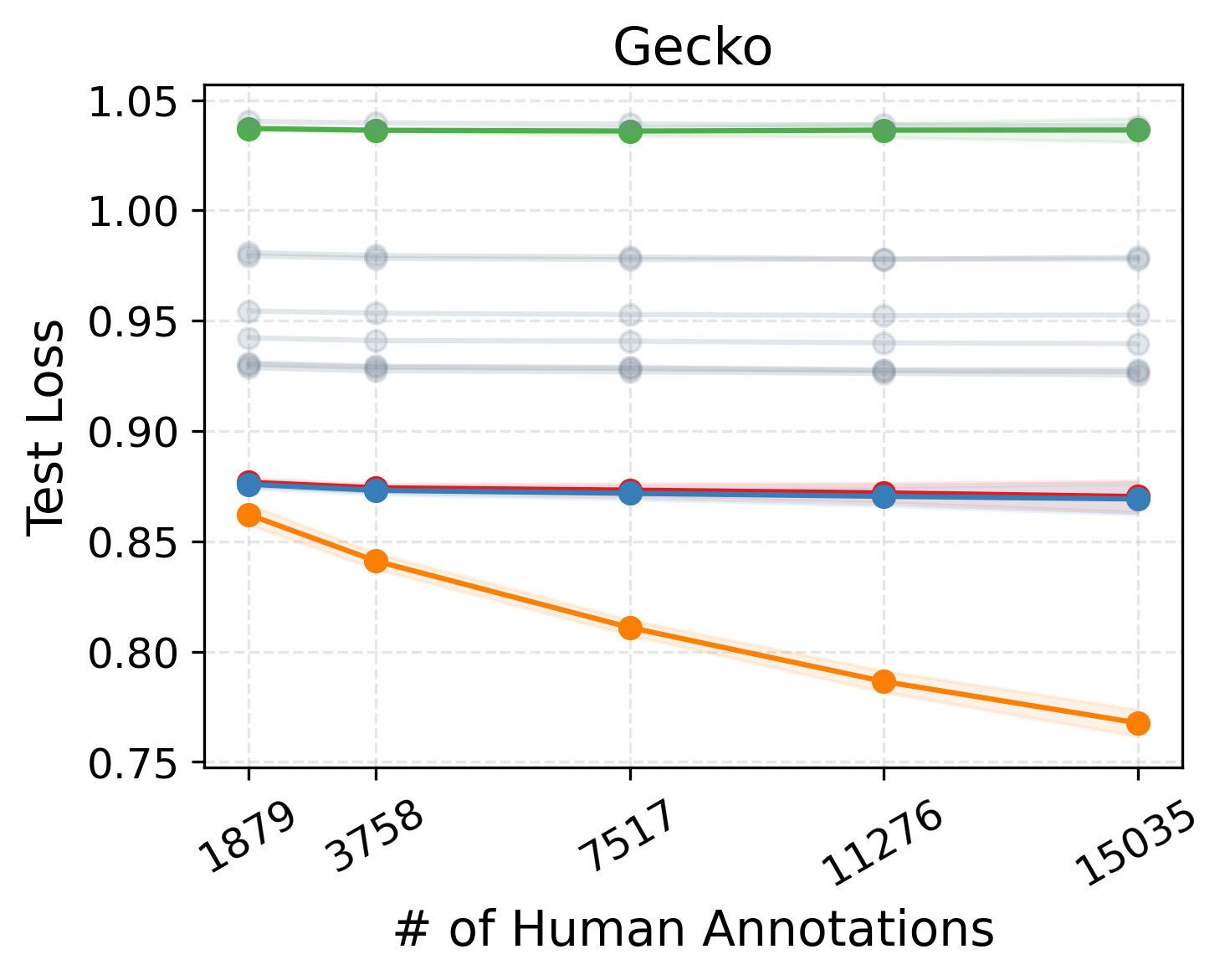}
    \end{subfigure}
    \begin{subfigure}[b]{0.36\textwidth}
        \centering
        \includegraphics[width=\linewidth]{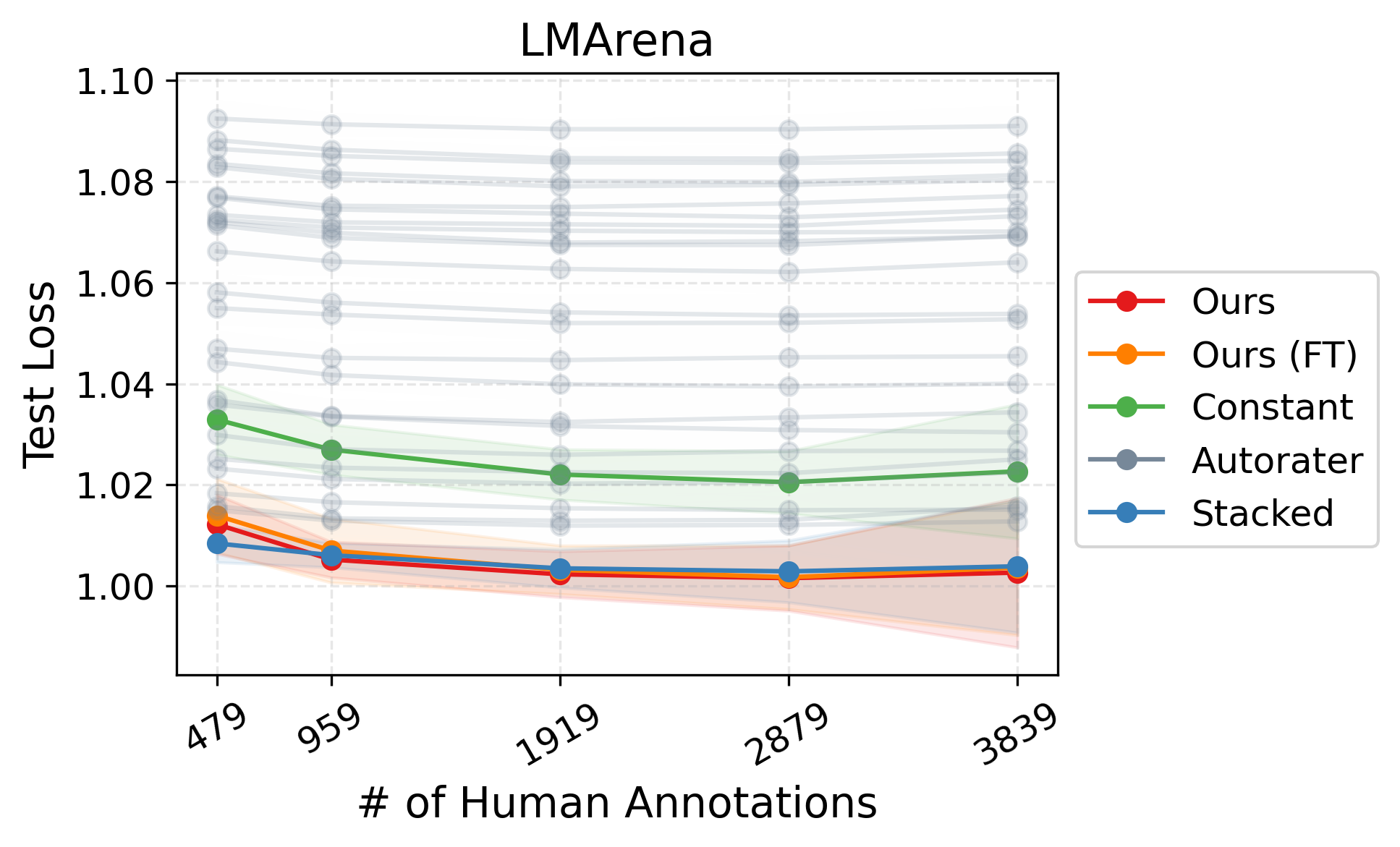}
    \end{subfigure}
    
    \caption{We compare our approaches with the constant baseline and a logistic regression model trained on the outputs of individual autoraters. When the autorater evaluation template matches the human template, we simply average the realizations $Y_{m,p,r}$ for the $(m,p,r)$ of interest and use this average as a feature for the logistic model. Conversely, when autoraters use single-sided templates while humans perform side-by-side comparisons, we average the realizations of $Y_{m,p,r}$ for each of the two models being evaluated and take the difference between these averages to obtain the features for the logistic regression. ``Stacked'' uses all autoraters at once but, unlike our model, shares no structure across models, prompts, or raters: it is an $\ell_2$-regularized multinomial logistic regression whose features are the average scores of all autoraters for $(m,p)$ (for side-by-side human labels, the per-autorater differences of these averages between the two models), with the penalty tuned by cross-validation.}
    \label{fig:varying_B_autoraters}
\end{figure}

Figure \ref{fig:varying_B_autoraters} compares our methods against baselines, including individual autoraters.

\subsection{Second stage with a varying fraction of autoraters}

\begin{figure}[H]
    \centering
    \begin{subfigure}[b]{0.275\textwidth}
        \centering
        \includegraphics[width=\linewidth]{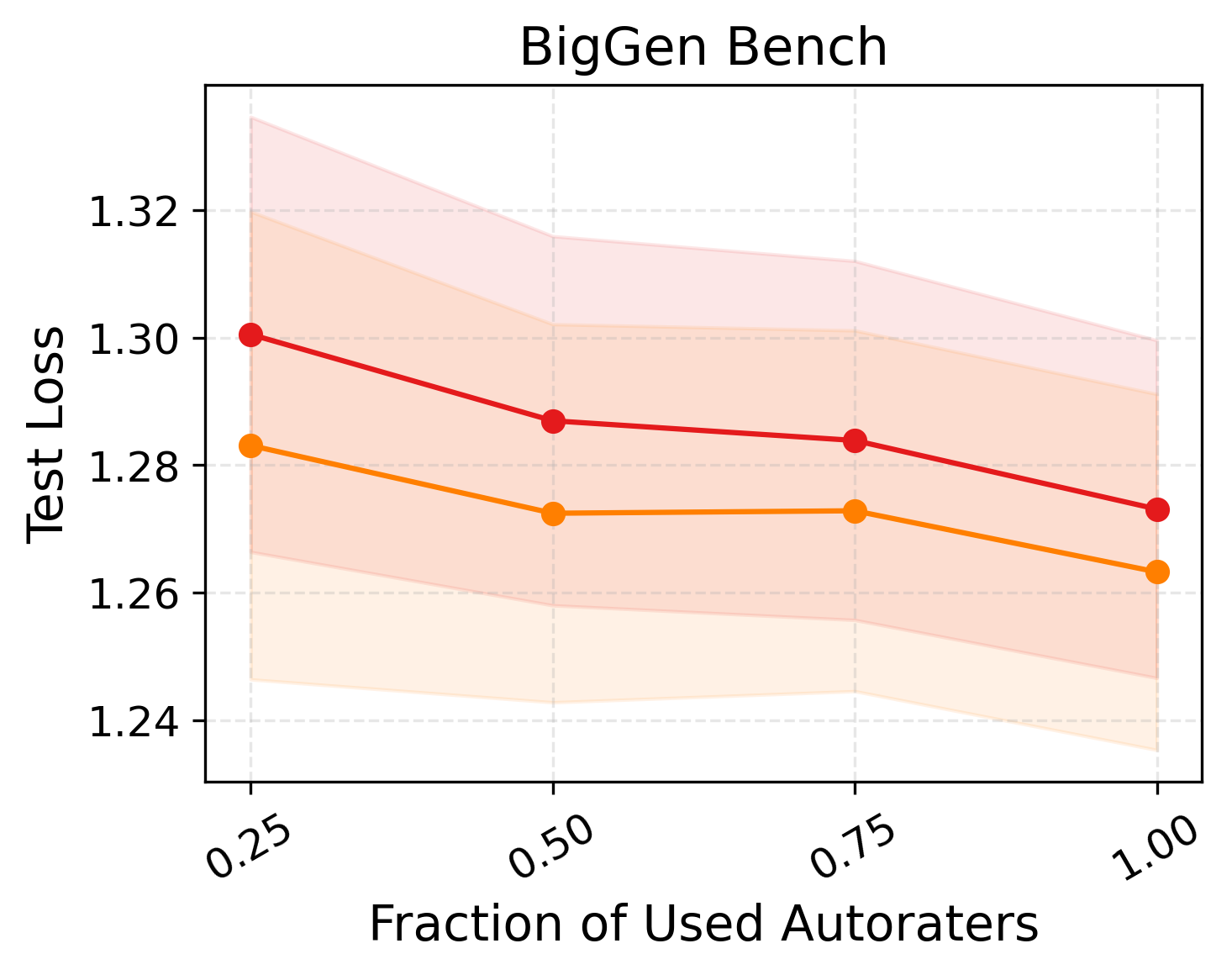}
    \end{subfigure}
    \begin{subfigure}[b]{0.275\textwidth}
        \centering
        \includegraphics[width=\linewidth]{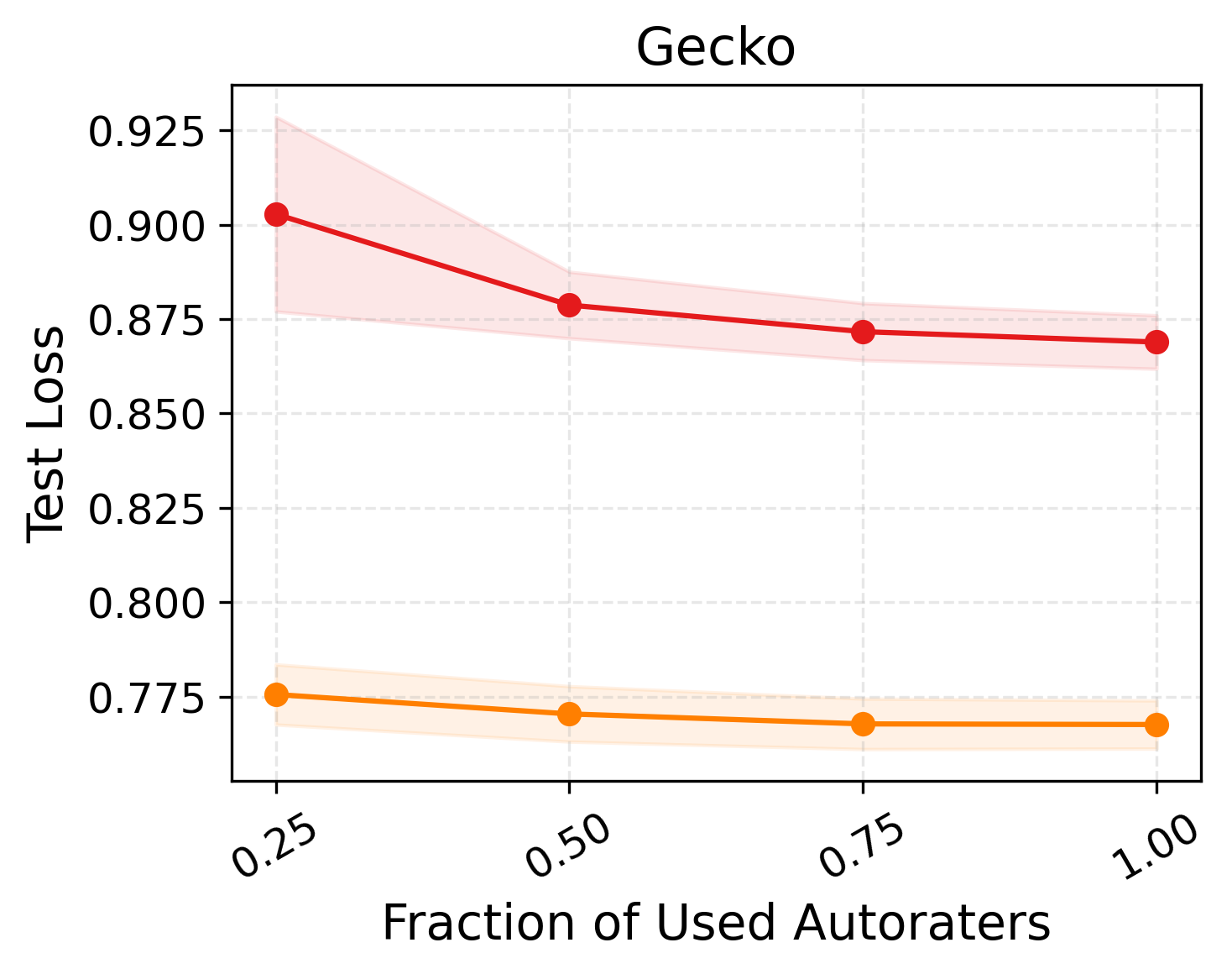}
    \end{subfigure}
    \begin{subfigure}[b]{0.36\textwidth}
        \centering
        \includegraphics[width=\linewidth]{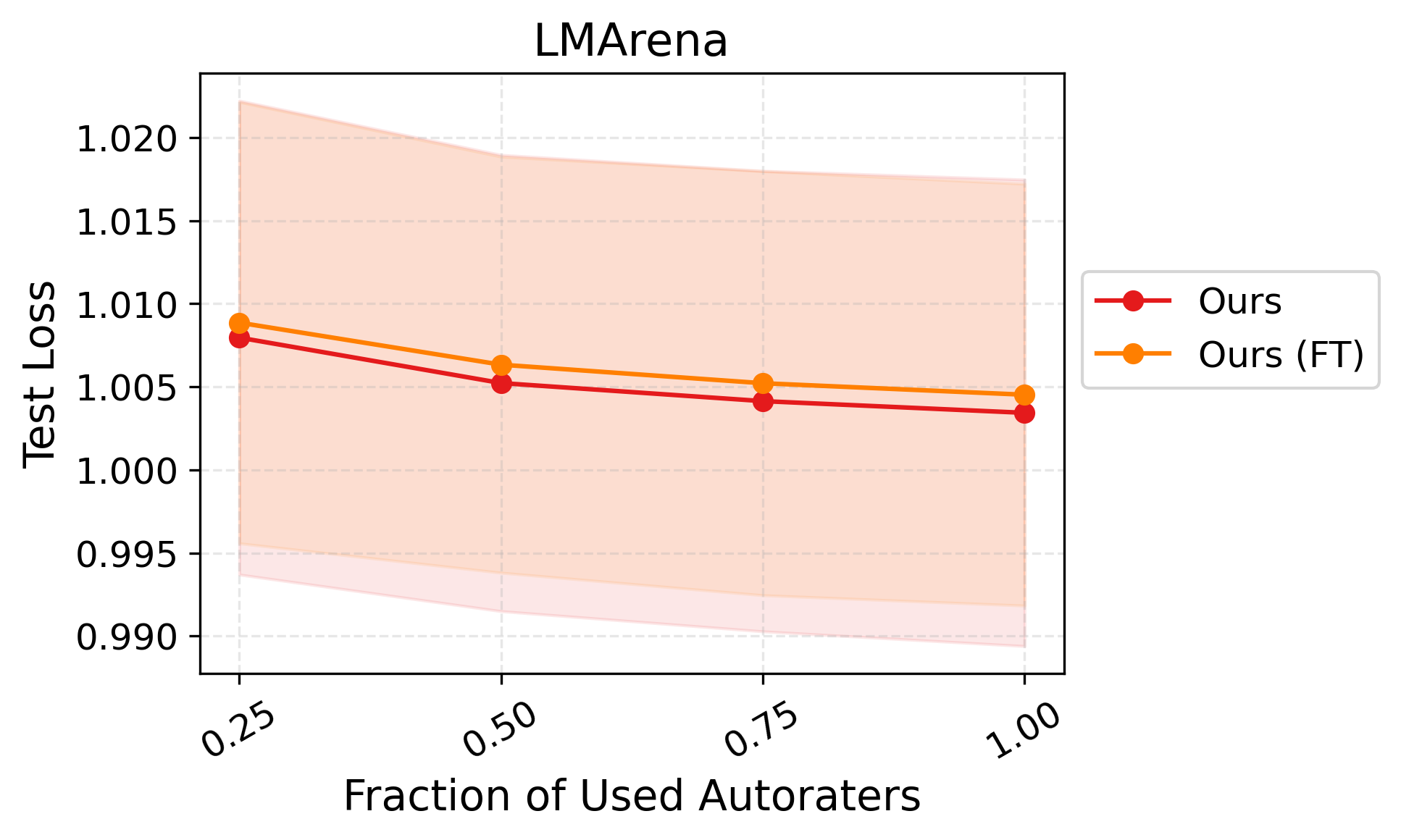}
    \end{subfigure}
    
    \caption{When varying the fraction of autoraters used, we observe that the highest gains occur at the lowest fractions. This suggests that the collective signal from autoraters may begin to saturate if sufficient diversity is not provided. In this specific experiment, we fix the fraction and sample multiple combinations of judges to satisfy that proportion; the error bars represent the variability across these combinations and different data splits. We allocate $80\%$ of the dataset for training and use the remainder for testing.}
    \label{fig:varying_autorater_fraction}
\end{figure}

Figure \ref{fig:varying_autorater_fraction} shows how the fraction of used autoraters impacts the second-stage losses. As expected, we see a negative trend.

\newpage
\subsection{Category-specific rankings (all plots)}

We expand Figures \ref{fig:ranking_cis_10_gecko} and \ref{fig:ranking_cis_10_bgb} to contain all top cohesive categories.

\begin{figure}[H]
    \centering
    \begin{subfigure}[b]{0.235\textwidth}
        \centering
        \includegraphics[width=1.025\linewidth]{figures/plots/rankings_cis/gecko_10/ranking_gecko_10_lang_compositional_compositional-3E-2A_v2.png}
    \end{subfigure}
    \begin{subfigure}[b]{0.23\textwidth}
        \centering
        \includegraphics[width=.9\linewidth]{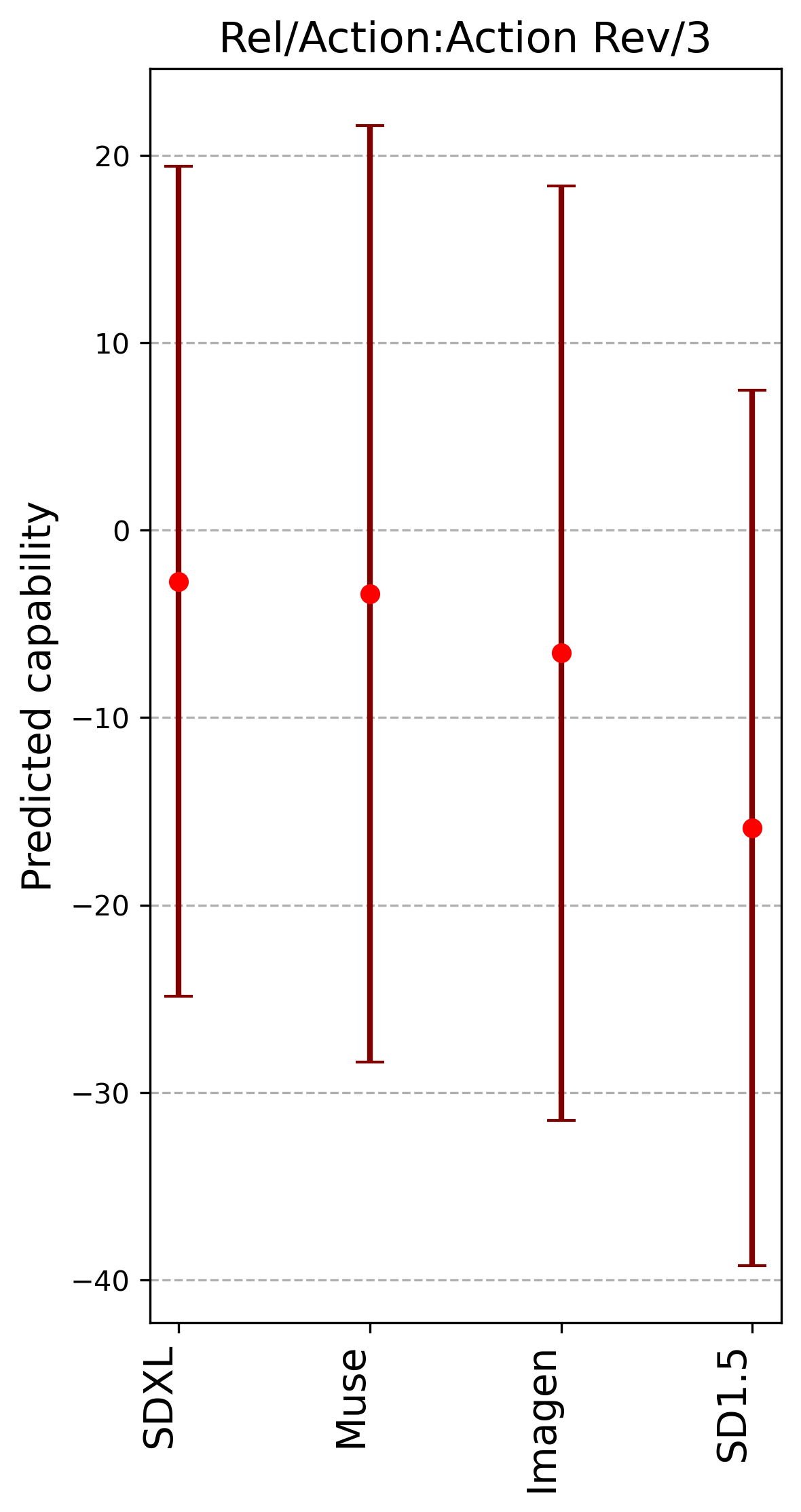}
    \end{subfigure}\\
    \begin{subfigure}[b]{0.23\textwidth}
        \centering
        \includegraphics[width=.925\linewidth]{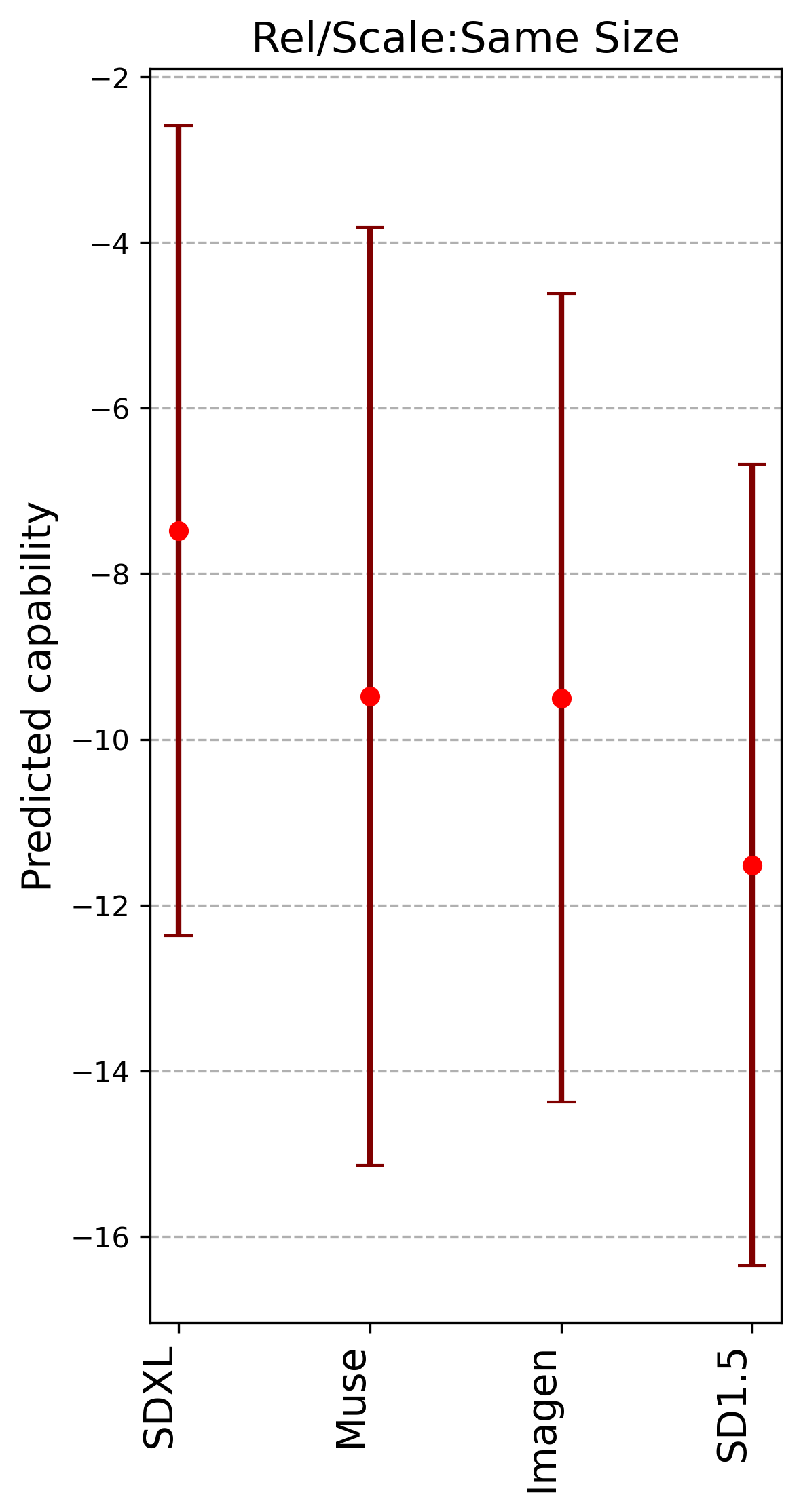}
    \end{subfigure}
    \begin{subfigure}[b]{0.23\textwidth}
        \centering
        \includegraphics[width=.9\linewidth]{figures/plots/rankings_cis/gecko_10/ranking_gecko_10_att_count_additive_3_v2.png}
    \end{subfigure}
    \begin{subfigure}[b]{0.23\textwidth}
        \centering
        \includegraphics[width=.9\linewidth]{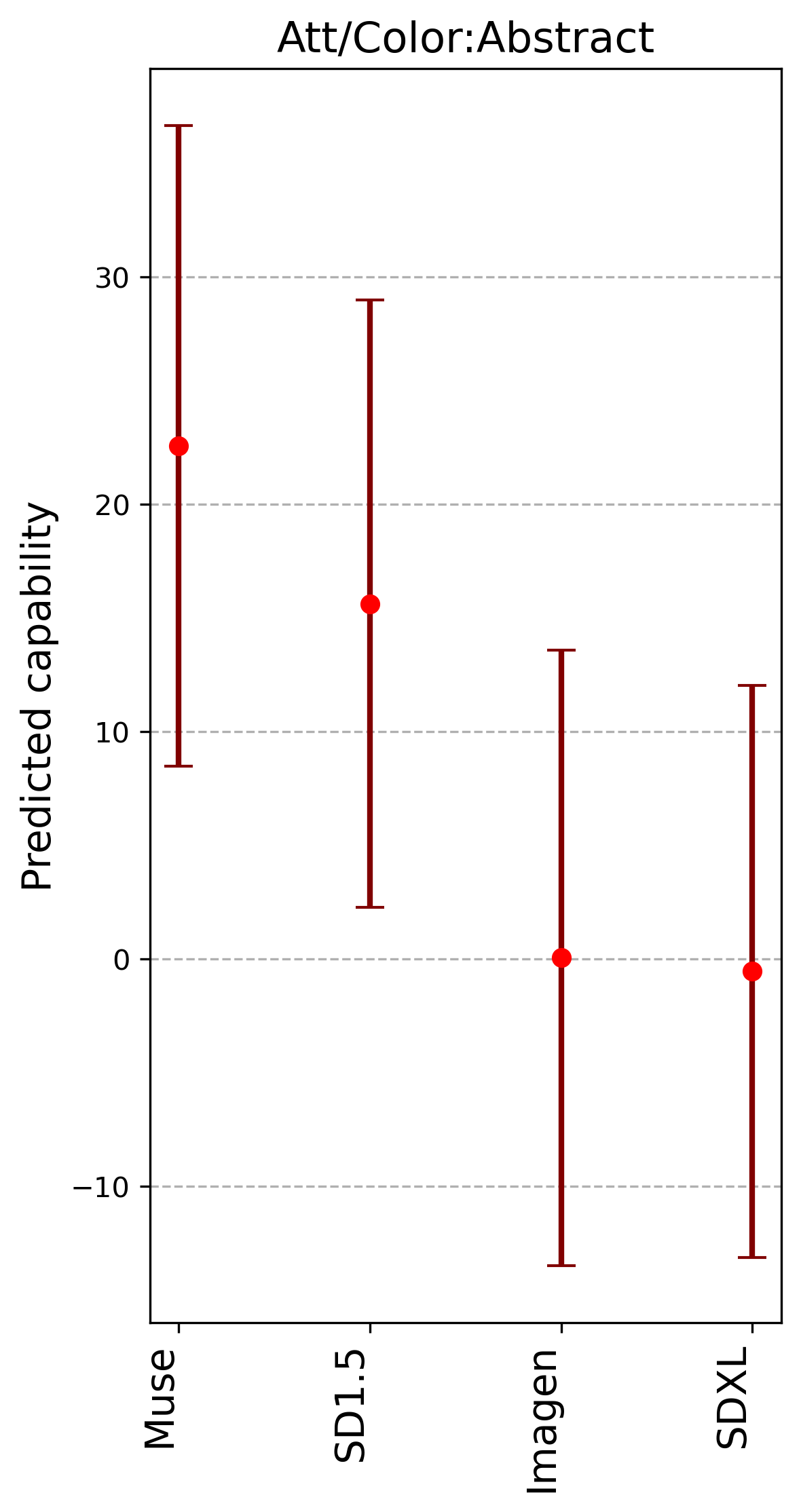}
    \end{subfigure}
    \caption{This is a version of Figure \ref{fig:ranking_cis_10_gecko} with more categories.}
    \label{fig:ranking_cis_10_gecko_full}
\end{figure}

\begin{figure}[H]
    \centering
    \begin{subfigure}[b]{0.23\textwidth}
        \centering
        \includegraphics[width=.9\linewidth]{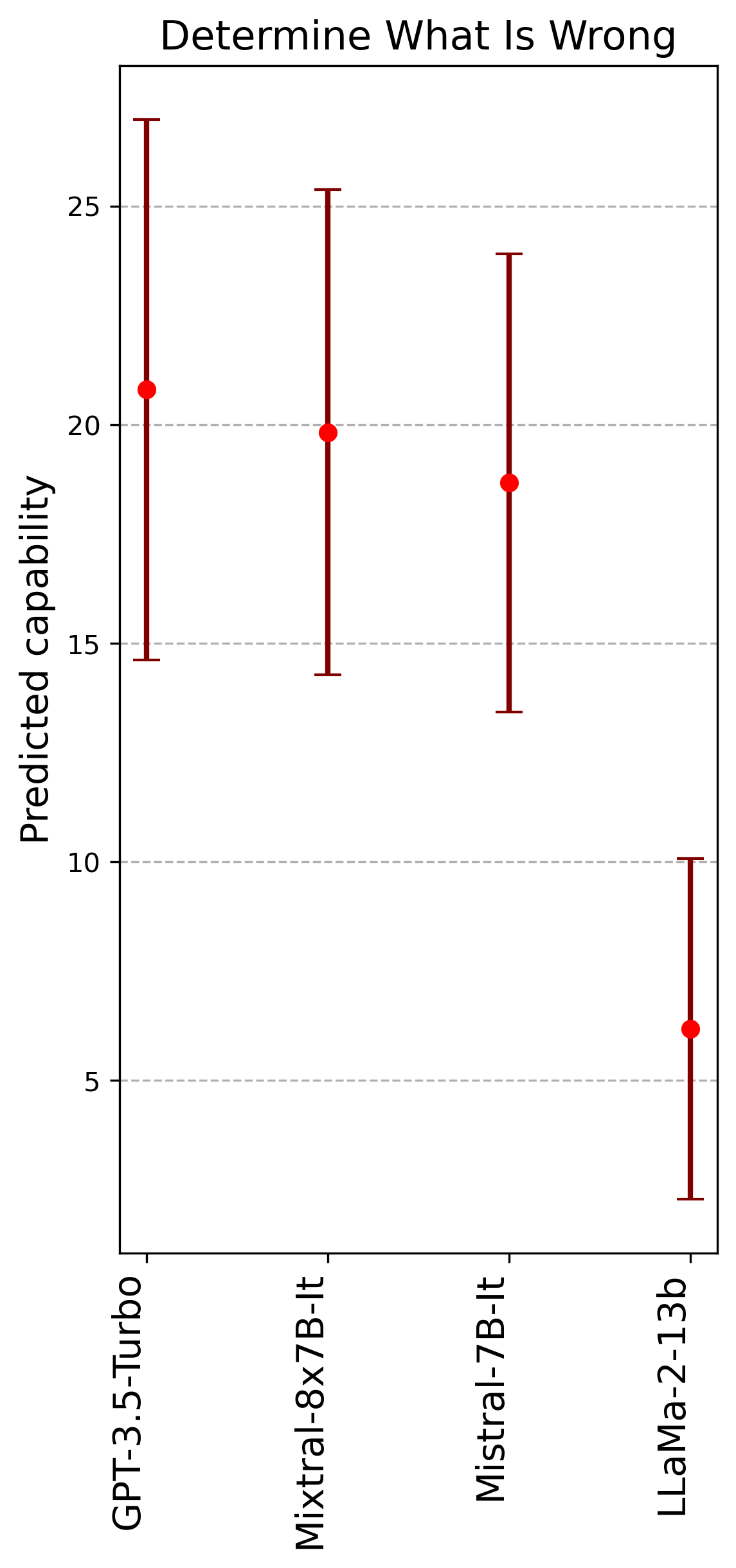}
    \end{subfigure}
    \begin{subfigure}[b]{0.23\textwidth}
        \centering
        \includegraphics[width=.9\linewidth]{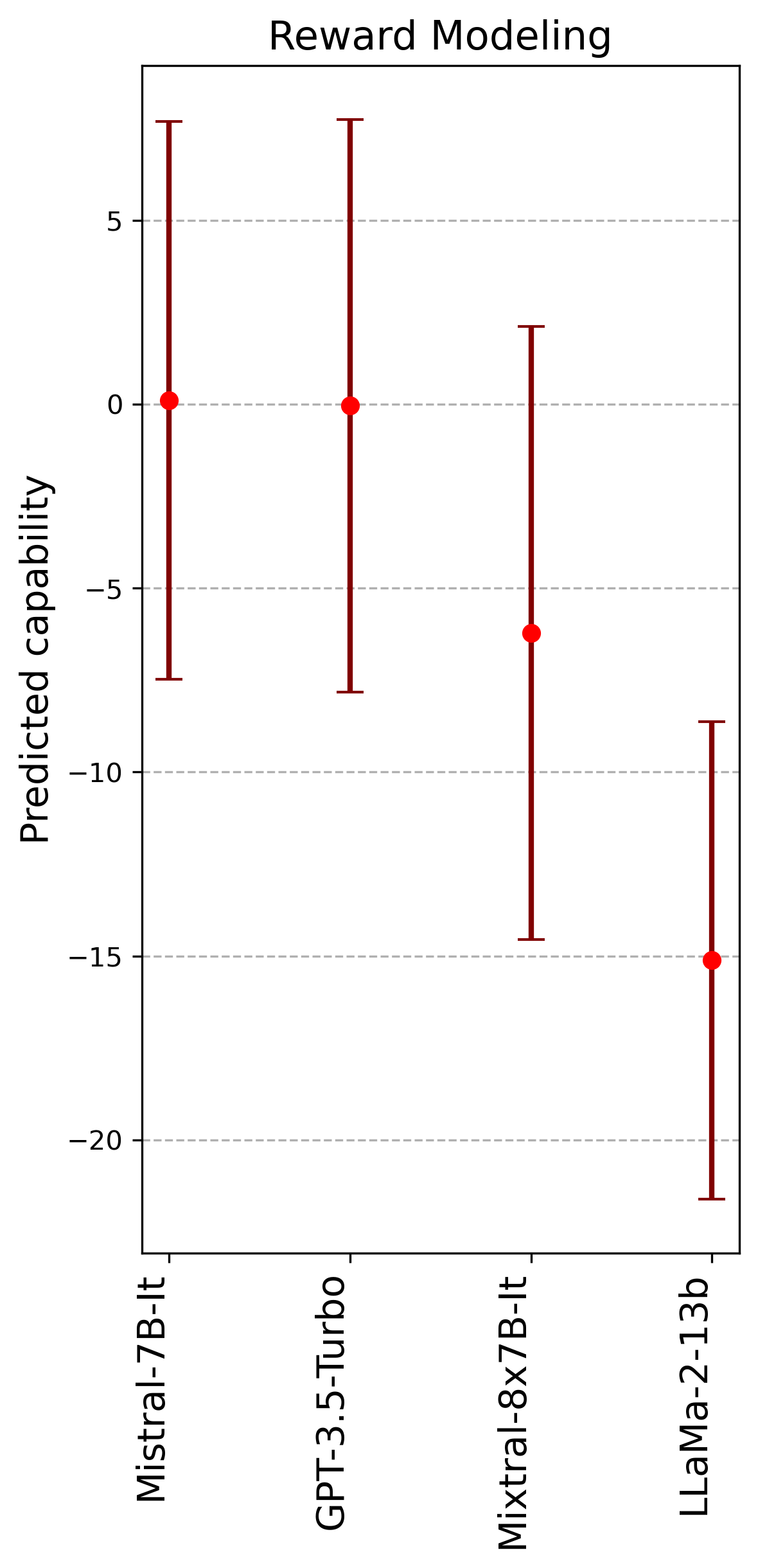}
    \end{subfigure}\\
    \begin{subfigure}[b]{0.23\textwidth}
        \centering
        \includegraphics[width=.9\linewidth]{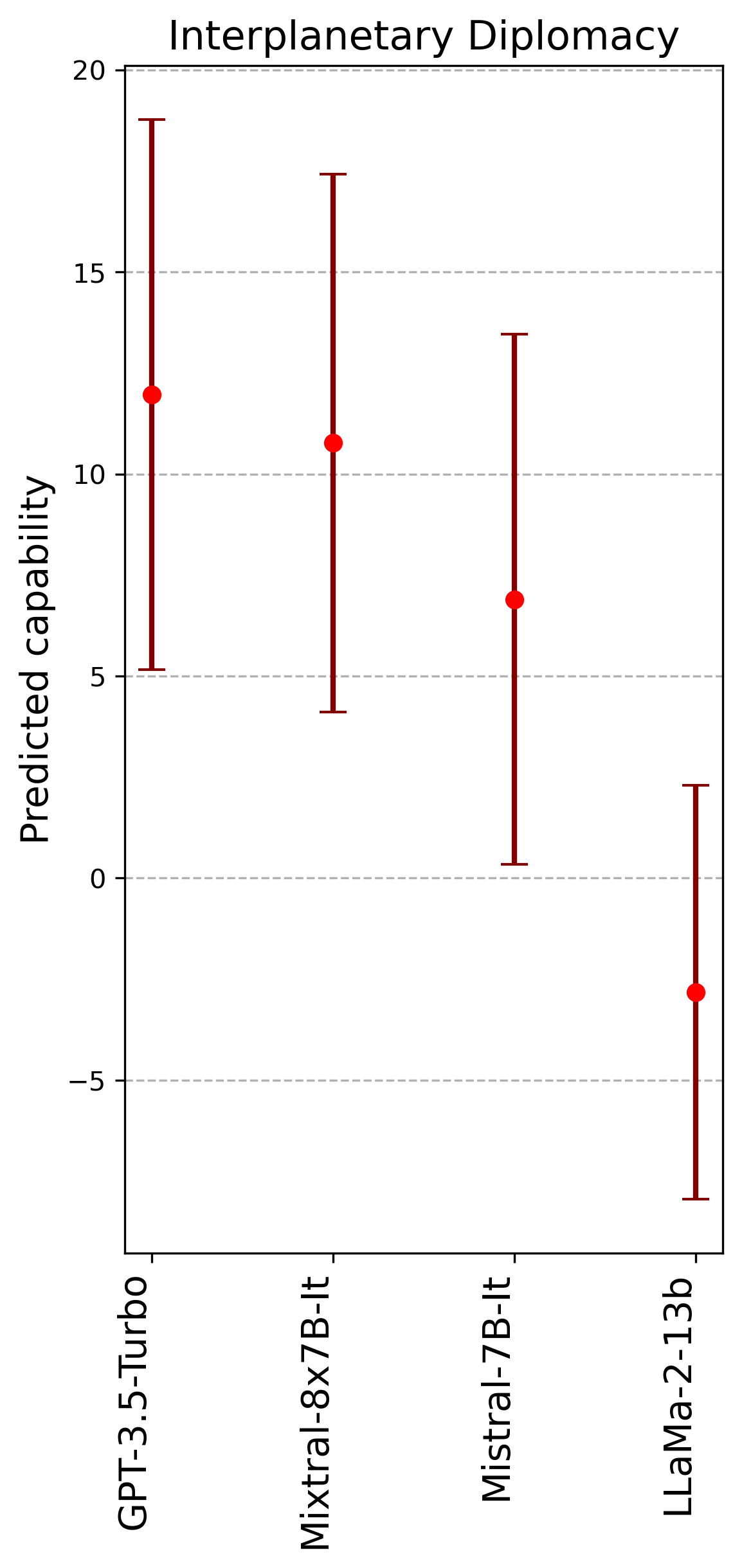}
    \end{subfigure}
    \begin{subfigure}[b]{0.23\textwidth}
        \centering
        \includegraphics[width=.9\linewidth]{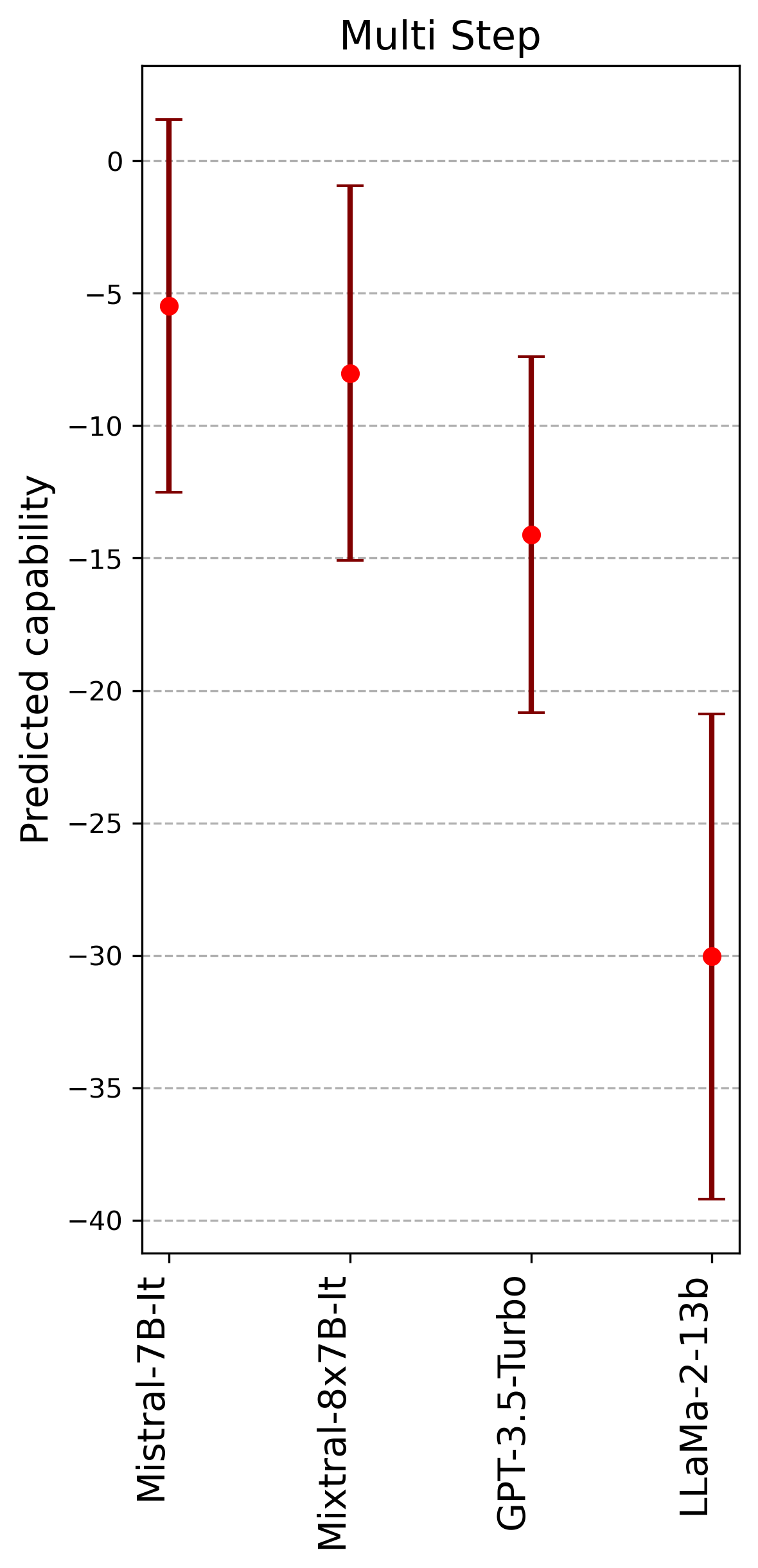}
    \end{subfigure}
    \begin{subfigure}[b]{0.23\textwidth}
        \centering
        \includegraphics[width=.9\linewidth]{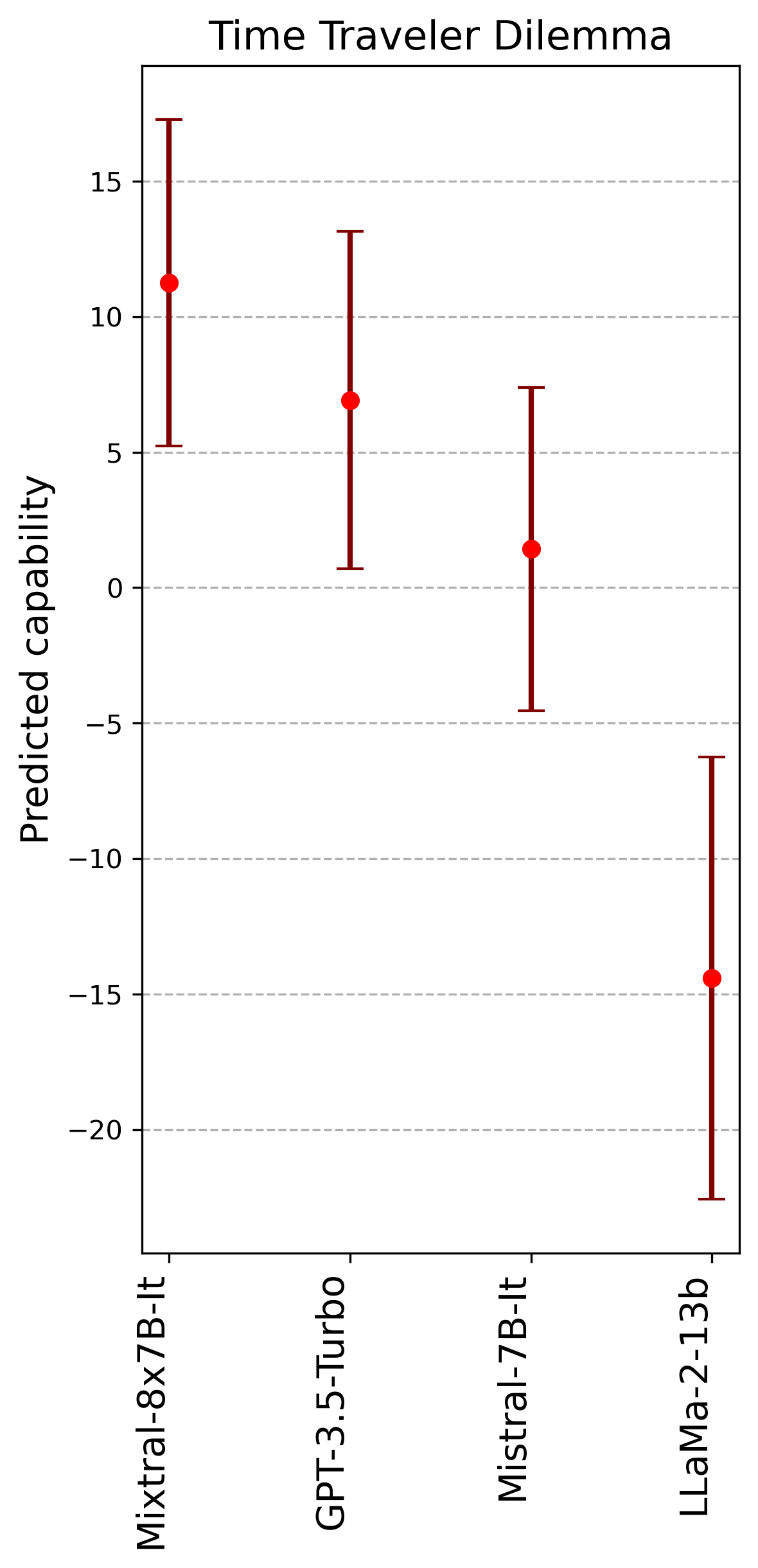}
    \end{subfigure}
    \caption{This is a version of Figure \ref{fig:ranking_cis_10_bgb} with more categories.}
    \label{fig:ranking_cis_10_bgb_full}
\end{figure}

\newpage
\subsection{Category-specific rankings (whole data)}

We complement the results of Figures \ref{fig:ranking_cis_10_gecko} and \ref{fig:ranking_cis_10_bgb} using the full human data to estimate human parameters. We see reduced interval lengths with qualitative results maintained.

\begin{figure}[H]
    \centering
    \begin{subfigure}[b]{0.235\textwidth}
        \centering
        \includegraphics[width=1.025\linewidth]{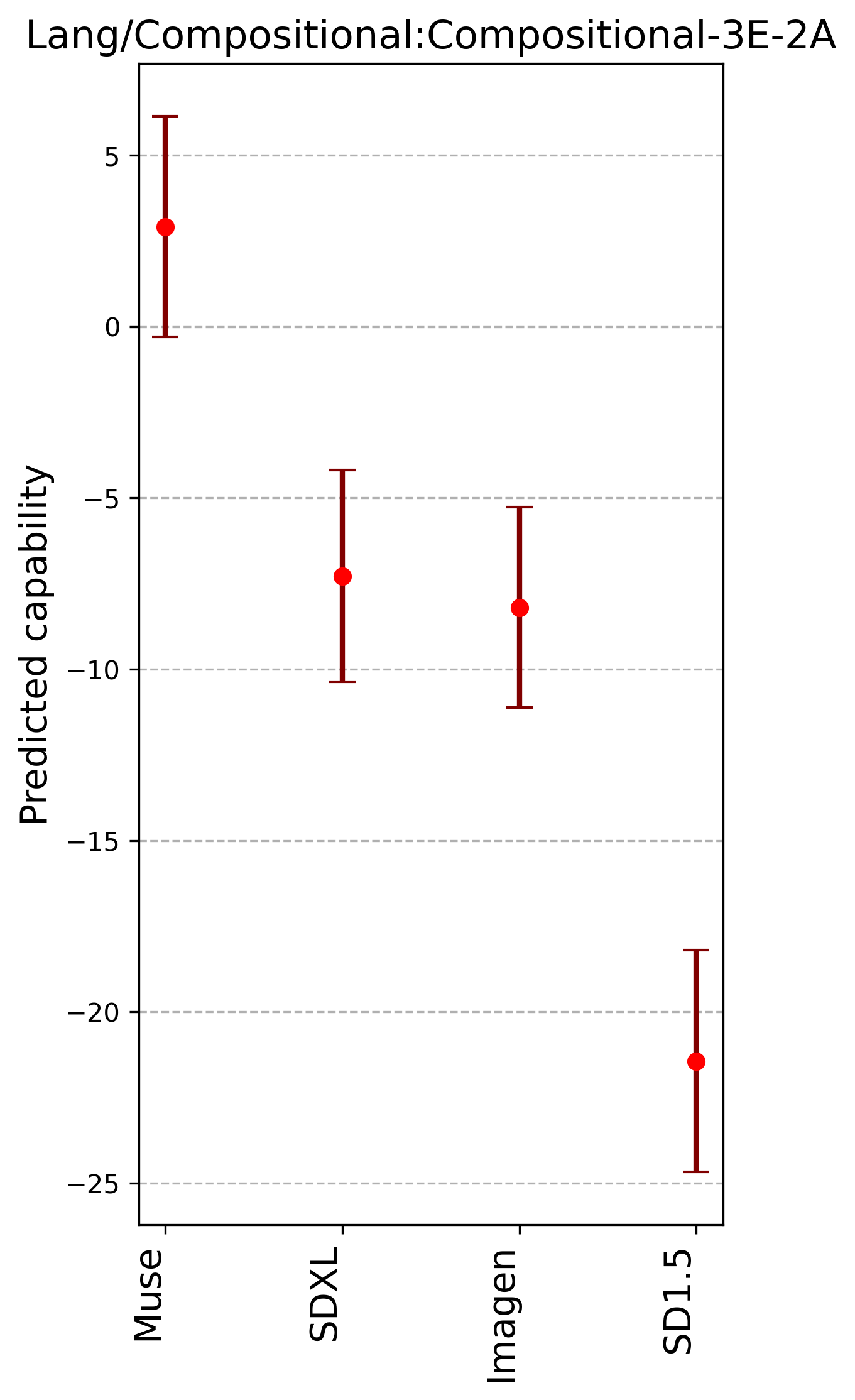}
    \end{subfigure}
    \begin{subfigure}[b]{0.23\textwidth}
        \centering
        \includegraphics[width=.9\linewidth]{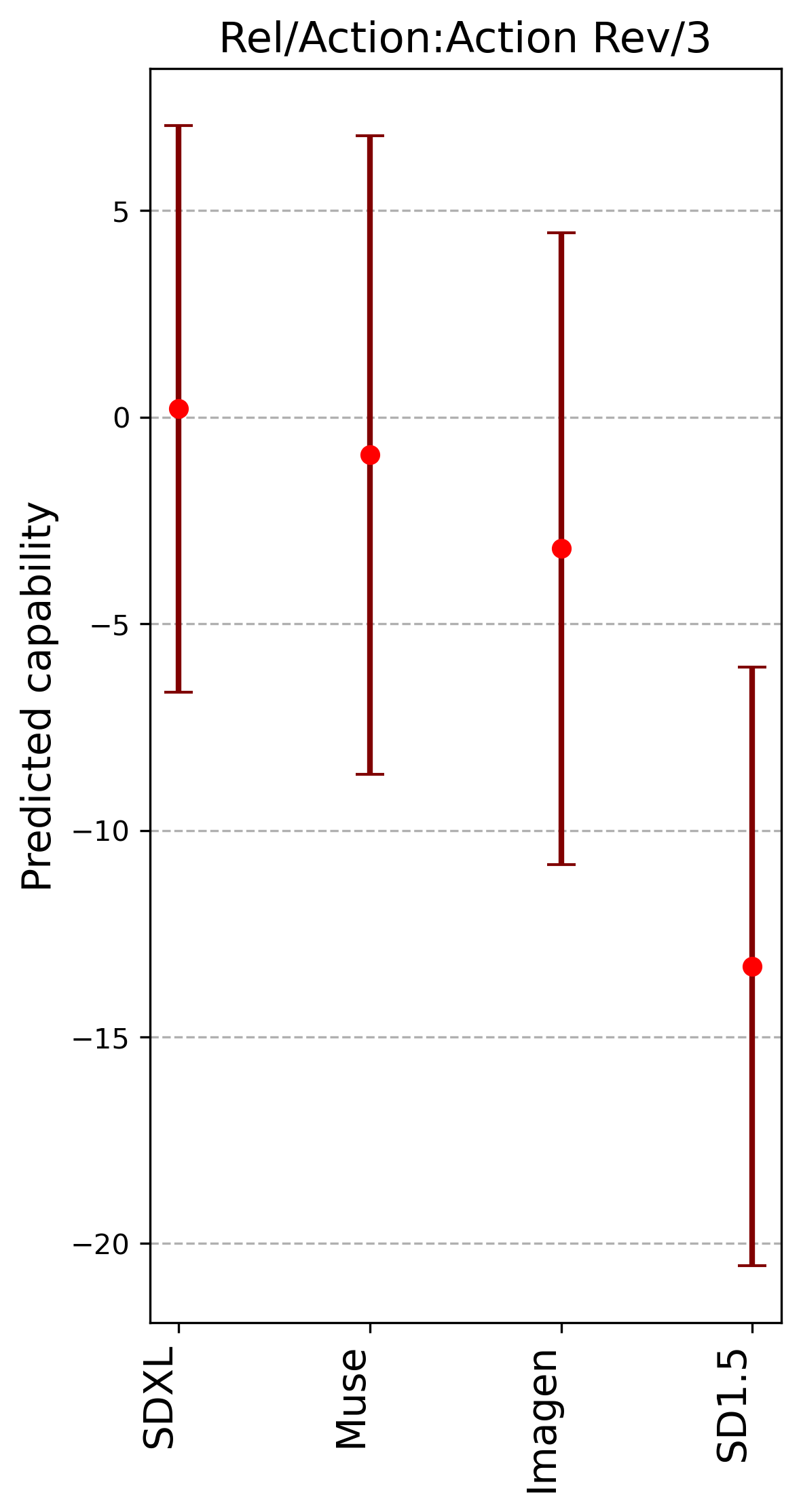}
    \end{subfigure}\\
    \begin{subfigure}[b]{0.23\textwidth}
        \centering
        \includegraphics[width=.925\linewidth]{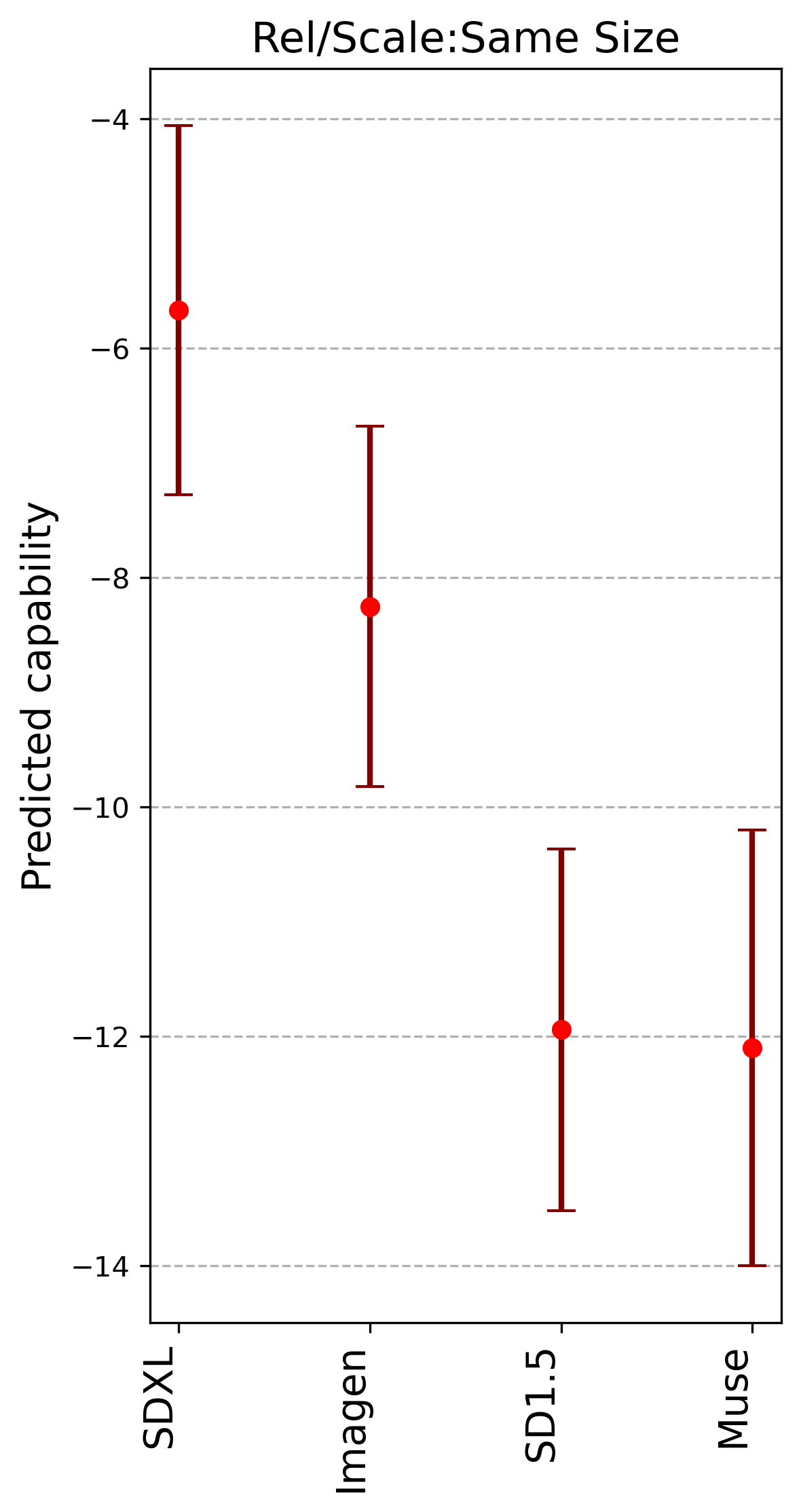}
    \end{subfigure}
    \begin{subfigure}[b]{0.23\textwidth}
        \centering
        \includegraphics[width=.9\linewidth]{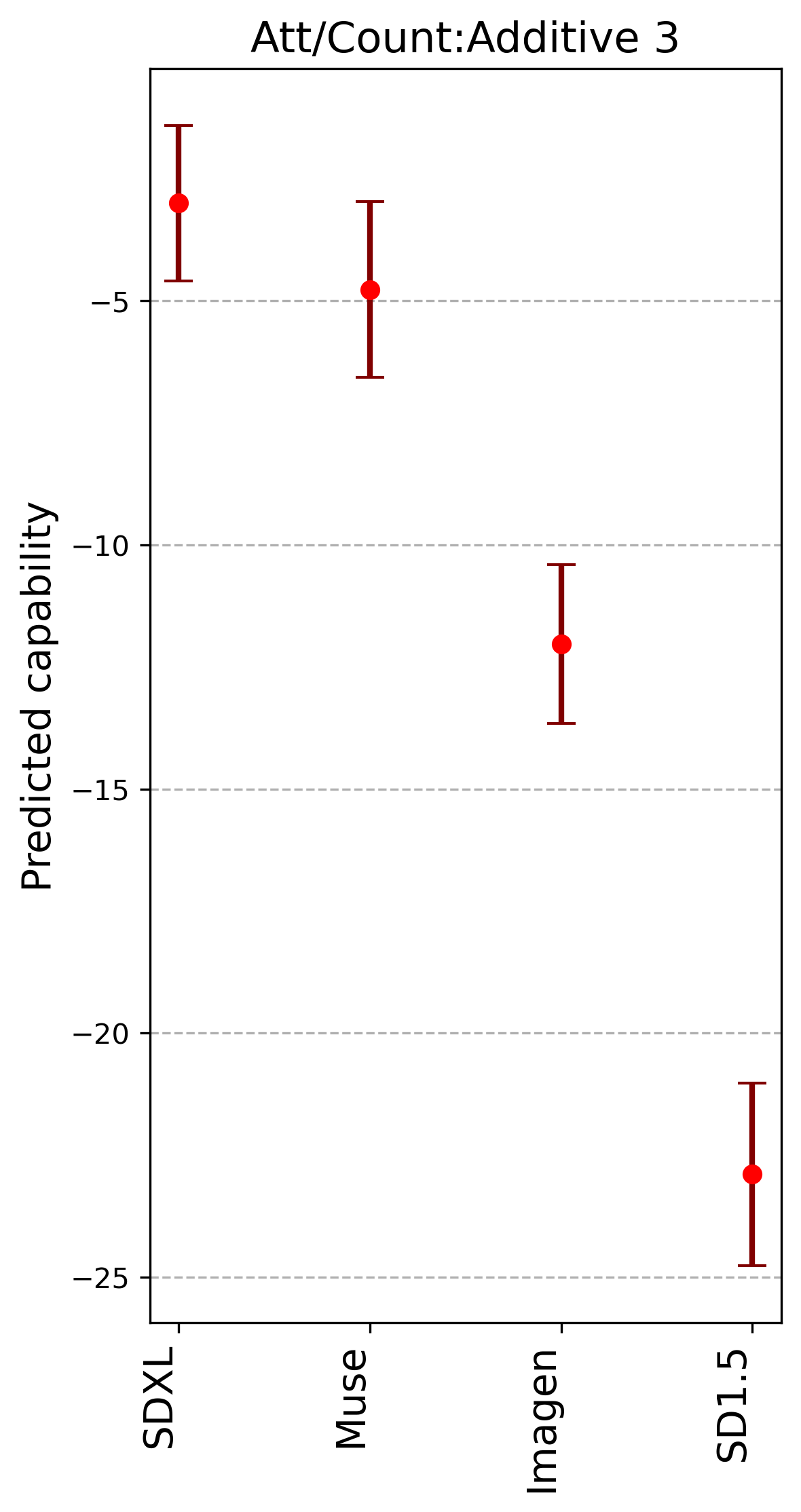}
    \end{subfigure}
    \begin{subfigure}[b]{0.23\textwidth}
        \centering
        \includegraphics[width=.9\linewidth]{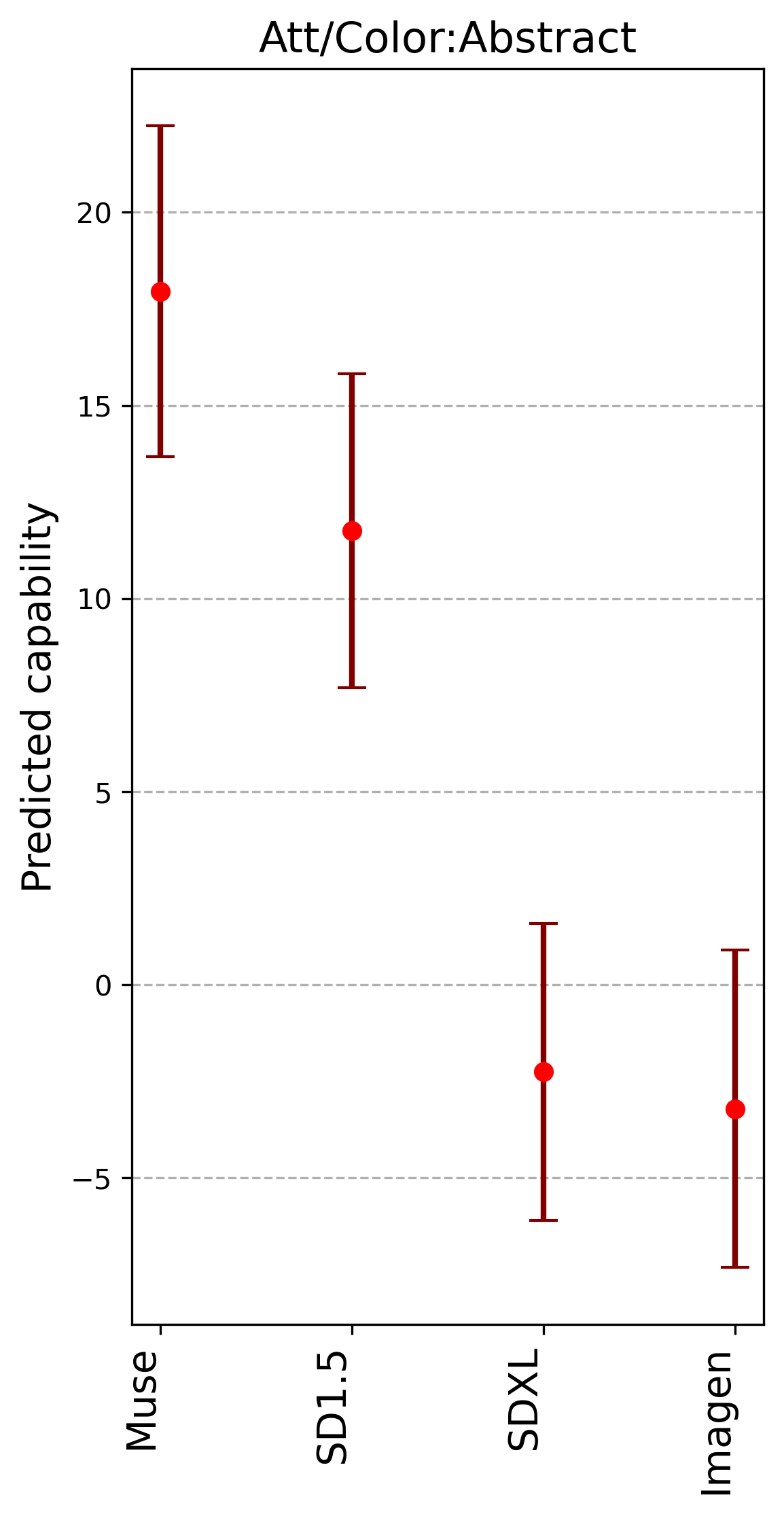}
    \end{subfigure}
    \caption{This plot illustrates how the estimates improve when we scale the number of human labels, in contrast to Figure \ref{fig:ranking_cis_10_gecko}, which utilizes only $10\%$. Overall, we observe that the model ordering is maintained, though the confidence intervals are evidently narrower. In practice, a practitioner could calibrate the required number of human labels by running a preliminary study to estimate the sample size needed to achieve a specific level of precision (knowing that the interval length is proportional to $1/\sqrt{|\cD^{(h)}|}$, where $|\cD^{(h)}|$ is the number of human annotations); this type of approach is commonly employed in Monte Carlo simulation for integral estimation. With all human labels, plug-in intervals can under-cover (Appendix \ref{sec:ci-marg}).}
    \label{fig:ranking_cis_100_gecko}
\end{figure}

\begin{figure}[H]
    \centering
    \begin{subfigure}[b]{0.23\textwidth}
        \centering
        \includegraphics[width=.9\linewidth]{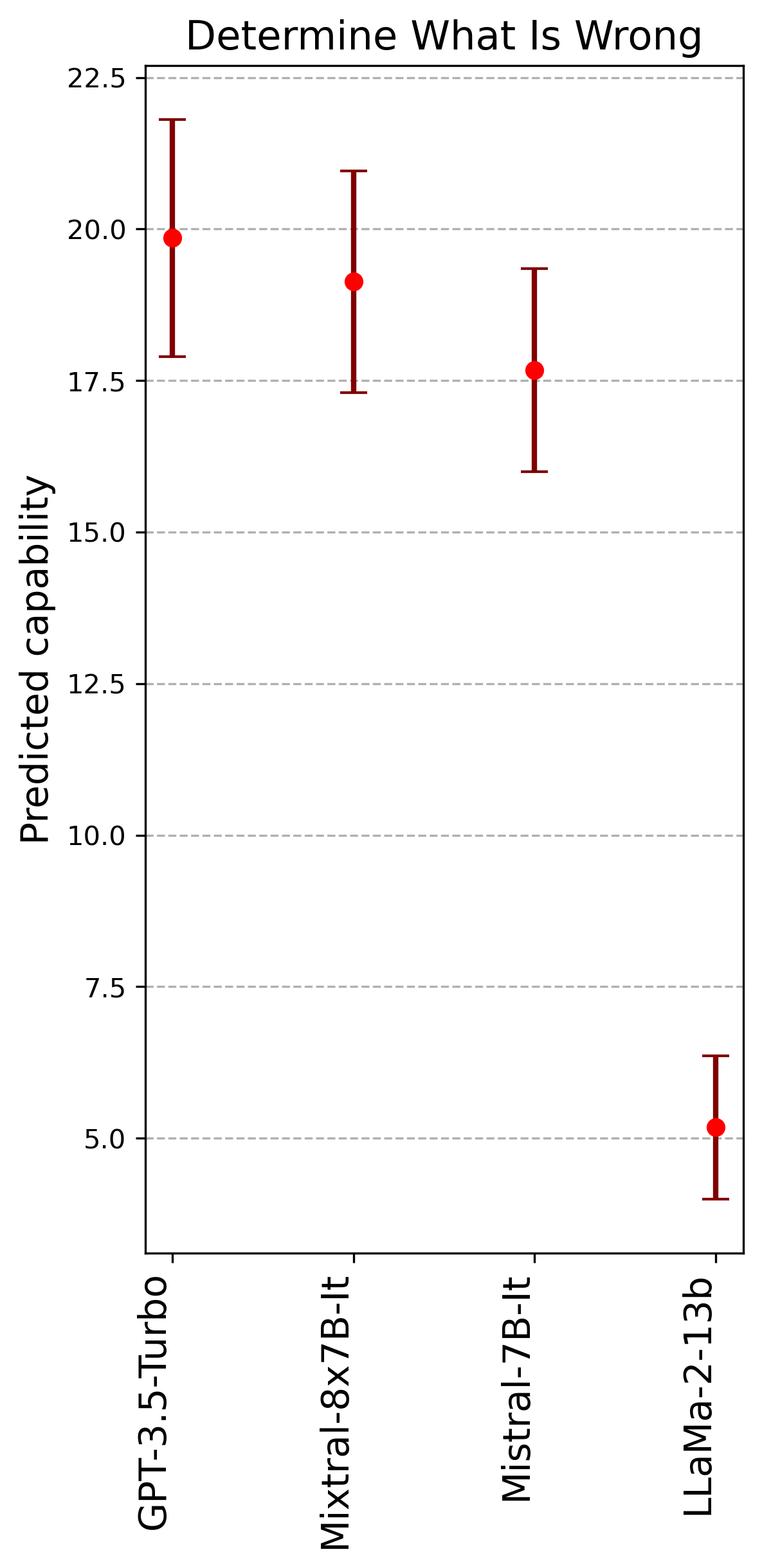}
    \end{subfigure}
    \begin{subfigure}[b]{0.23\textwidth}
        \centering
        \includegraphics[width=.9\linewidth]{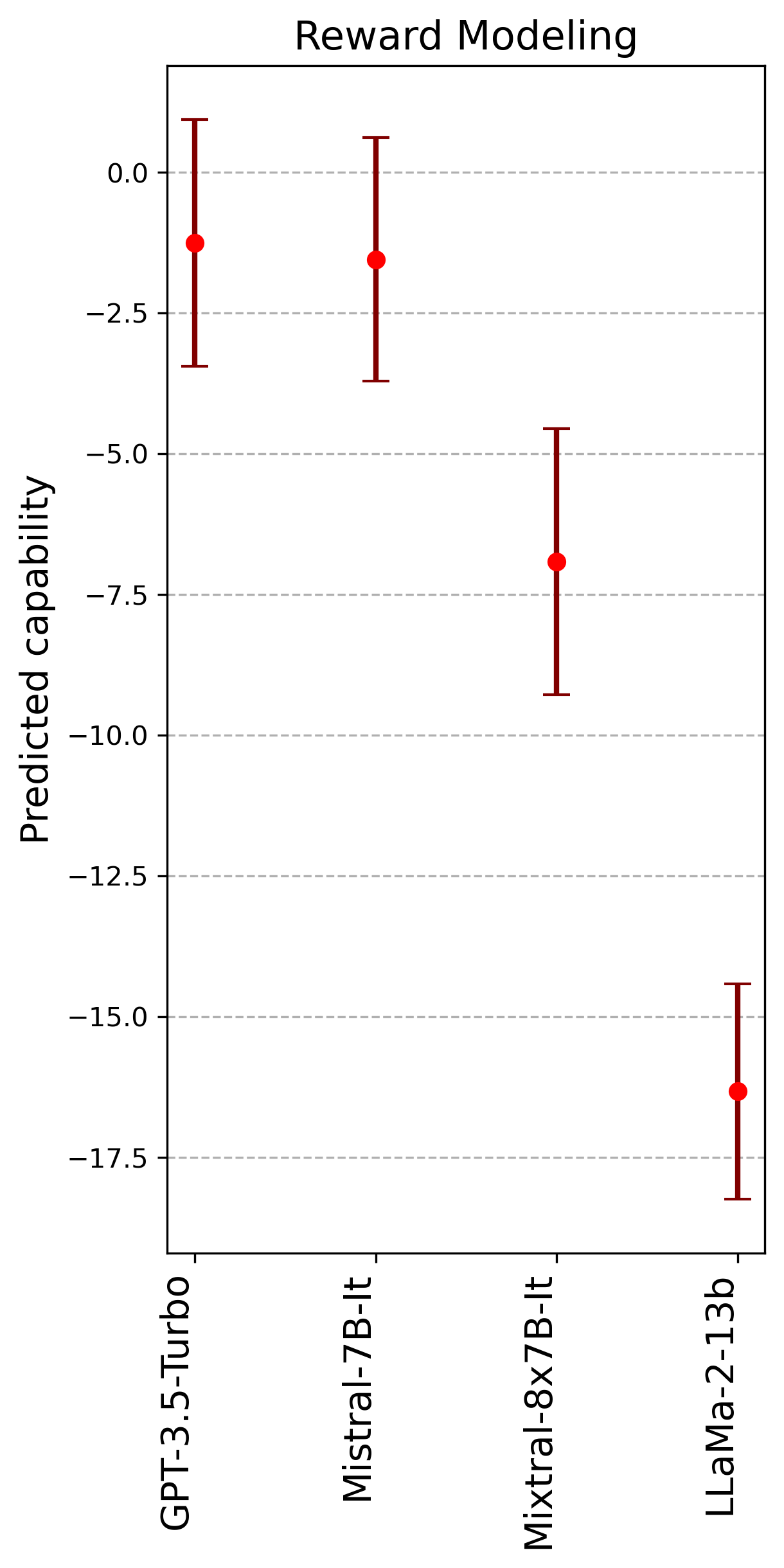}
    \end{subfigure}\\
    \begin{subfigure}[b]{0.23\textwidth}
        \centering
        \includegraphics[width=.9\linewidth]{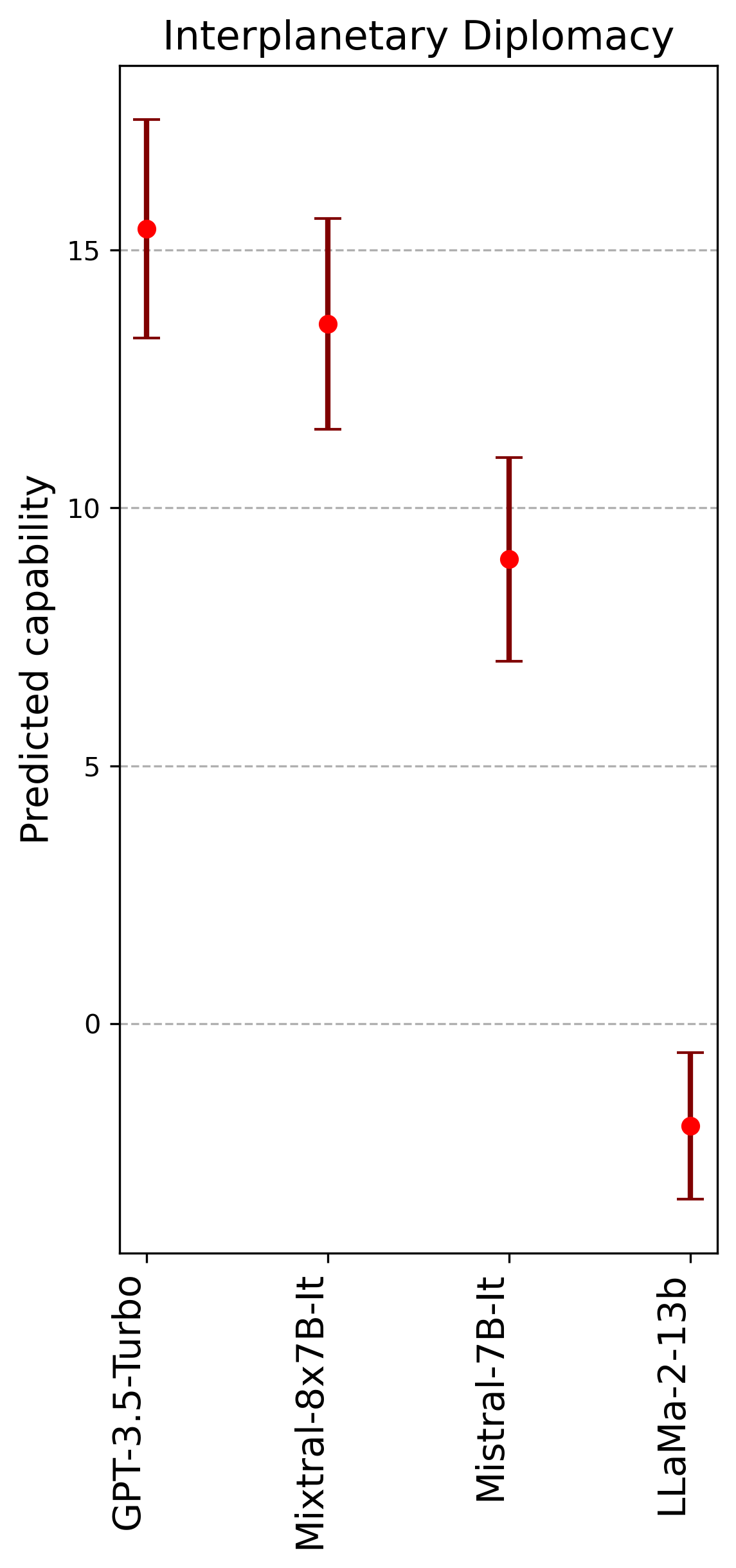}
    \end{subfigure}
    \begin{subfigure}[b]{0.23\textwidth}
        \centering
        \includegraphics[width=.9\linewidth]{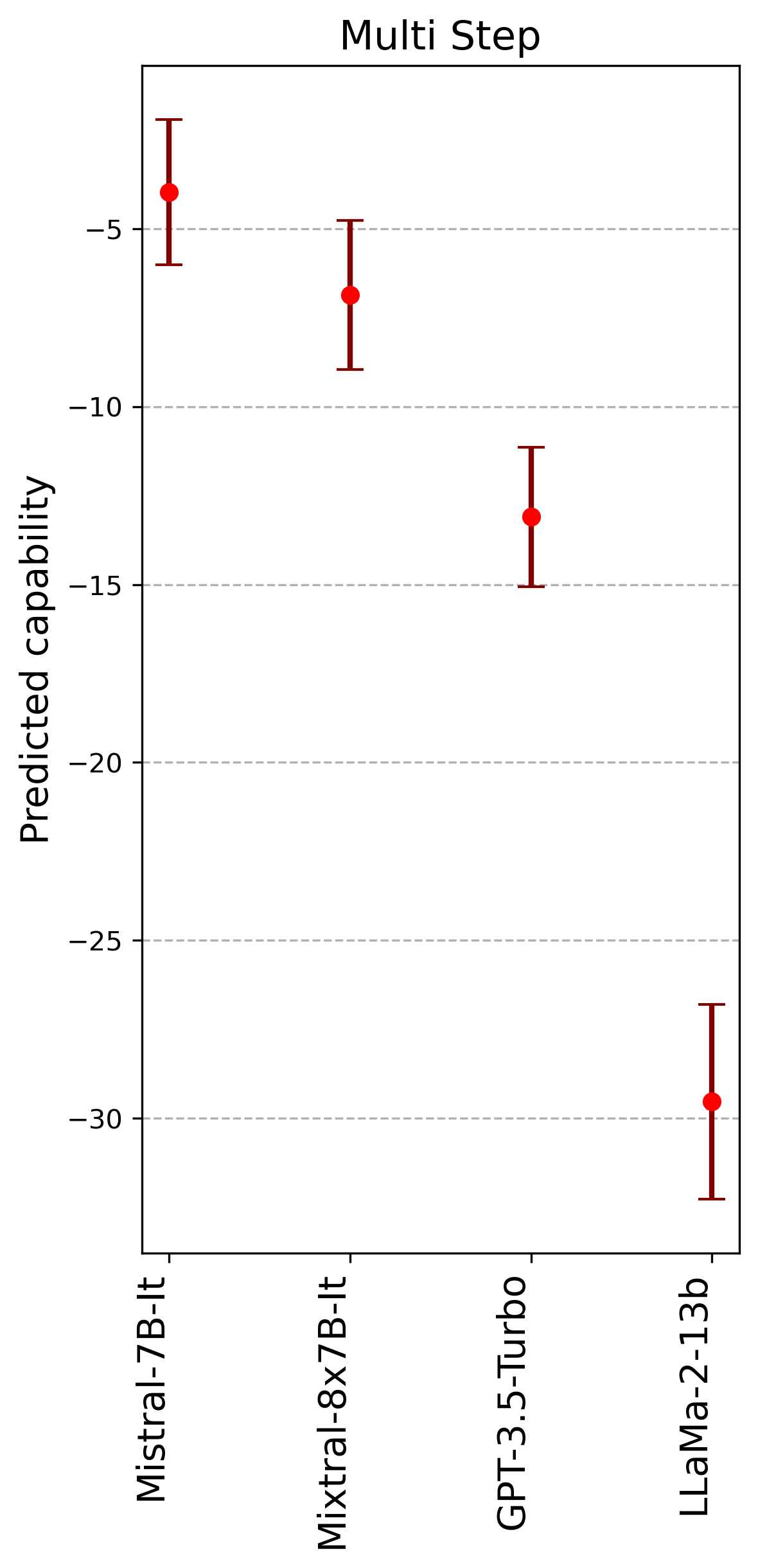}
    \end{subfigure}
    \begin{subfigure}[b]{0.23\textwidth}
        \centering
        \includegraphics[width=.9\linewidth]{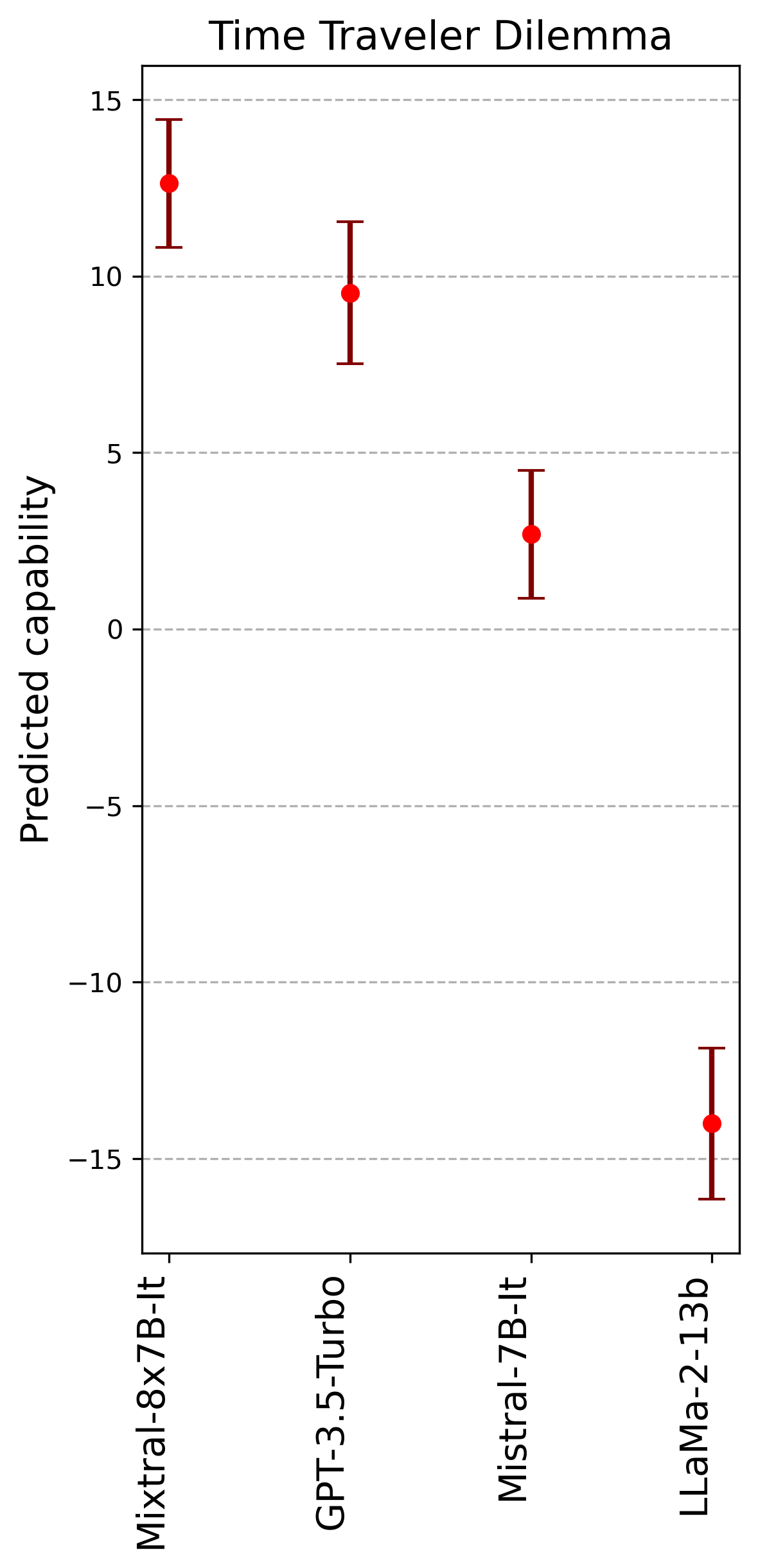}
    \end{subfigure}
    \caption{This plot illustrates how the estimates improve when we scale the number of human labels, in contrast to Figure \ref{fig:ranking_cis_10_bgb}. As for Gecko, we observe that the model ordering is roughly maintained, though the confidence intervals are evidently narrower. With all human labels, plug-in intervals can under-cover (Appendix \ref{sec:ci-marg}).}
    \label{fig:ranking_cis_100_bgb}
\end{figure}

\newpage
\subsection{Exploring strengths and weaknesses of different models (whole data)}

We complement the results of Figure \ref{fig:deltas_cis_10} using the full human data to estimate human parameters. We see reduced interval lengths with qualitative results maintained.

\begin{figure}[H]
    \centering
    \begin{subfigure}[b]{0.95\textwidth}
        \centering
        \includegraphics[width=\linewidth]{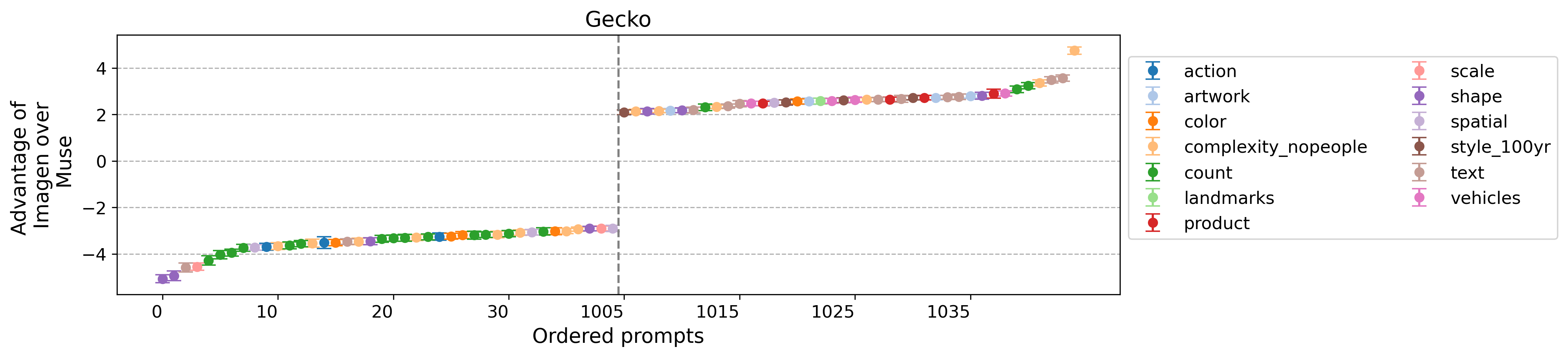}
    \end{subfigure}  
    \begin{subfigure}[b]{0.95\textwidth}
        \centering
        \includegraphics[width=\linewidth]{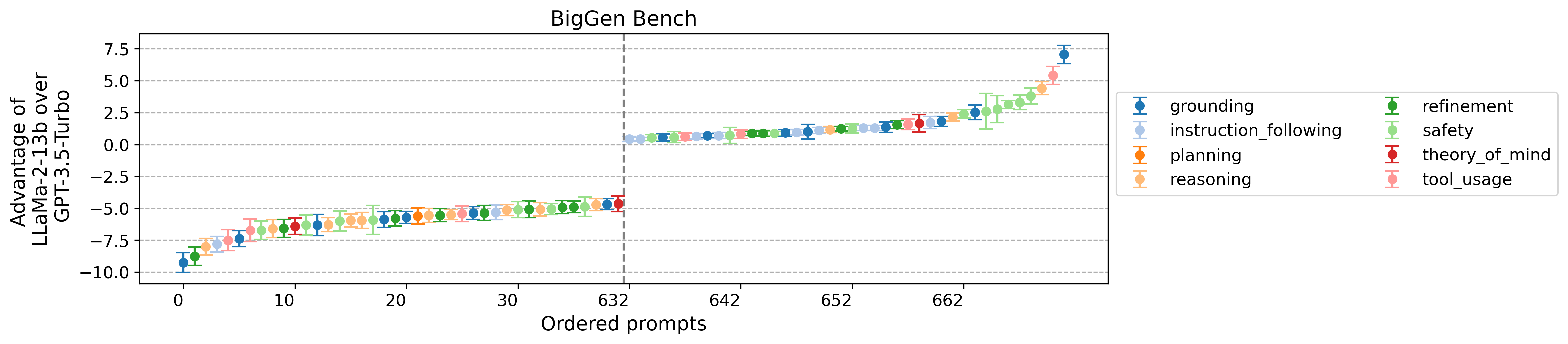}
    \end{subfigure}\\
    \caption{\small Fine-grained comparison of model capabilities using $100\%$ of human annotations. Top: Difference in estimated scores between Imagen and Muse on Gecko prompts, colored by category. Imagen excels in text rendering, while Muse shows advantages in object counting. Bottom: Difference between LLaMa-2-13b and GPT-3.5-Turbo on BigGen Bench prompts. GPT-3.5-Turbo demonstrates a significant advantage in reasoning-related prompts. With all human labels, plug-in intervals can under-cover (Appendix \ref{sec:ci-marg}).}
    \label{fig:deltas_cis_100}
\end{figure}

\newpage
\section{Autoraters}\label{sec:auto}

In this section, we explore in more detail how autoraters were built. For Gecko and LMArena, all custom autoraters use Gemini 2.5 Flash-Lite, chosen for its price and throughput, which made collecting hundreds of thousands of scores feasible; a lightweight backbone makes the auxiliary signal noisier, which works against our method rather than for it.

\subsection{Side-by-side templates}

\subsubsection{Personas}
\begin{xltabular}{\textwidth}{@{}
    >{\raggedright\arraybackslash}p{2.5cm} 
    >{\raggedright\arraybackslash}p{2.5cm} 
    >{\raggedright\arraybackslash}X        
    >{\raggedright\arraybackslash}X        
@{}}

    \caption{Side-by-side personas (from Gecko)}\\

    \toprule
    \textbf{Persona} & \textbf{Archetype} & \textbf{Primary Objective} & \textbf{Evaluation Methodology} \\
    \midrule
    \endfirsthead

    \multicolumn{4}{c}{{\bfseries \tablename\ \thetable{} -- continued from previous page}} \\
    \toprule
    \textbf{Persona} & \textbf{Archetype} & \textbf{Primary Objective} & \textbf{Evaluation Methodology} \\
    \midrule
    \endhead

    \midrule
    \multicolumn{4}{r}{{Continued on next page...}} \\
    \endfoot

    \bottomrule
    \endlastfoot


    \textbf{Persona 1} \newline Original Template & 
    The Professional Baseline & 
    Determine faithful and complete representation of the text. & 
    Justify choices based on key elements of the prompt while explicitly ignoring artistic quality. \\
    \midrule

    \textbf{Persona 2} \newline Quality Assurance Inspector & 
    The Technical Specialist & 
    Check conformance to a ``Design Specification.'' & 
    Identify and report the severity of deviations (critical, major, minor) rather than subjective style preferences. \\
    \midrule

    \textbf{Persona 3} \newline Scientific Abstract & 
    The Academic Extremist & 
    Assess semantic correspondence and alignment. & 
    Deconstruct the prompt into core semantic components (subject, action, attributes) and systematically evaluate the rendering of each. \\
    \midrule

    \textbf{Persona 4} \newline Conversational & 
    The Human Communicator & 
    Determine which image ``followed directions'' better. & 
    Provide a casual, plain-English explanation of thoughts and misses, disregarding which image is ``prettier.'' \\

\end{xltabular}

\begin{xltabular}{\textwidth}{@{}
    >{\raggedright\arraybackslash}p{2.5cm} 
    >{\raggedright\arraybackslash}p{2.5cm} 
    >{\raggedright\arraybackslash}X        
    >{\raggedright\arraybackslash}X        
@{}}

    \caption{Side-by-side and Single-sided personas (from LMArena)}\label{tab:personas-lmarena}\\

    \toprule
    \textbf{Persona} & \textbf{Archetype} & \textbf{Guiding Principle} & \textbf{Methodology} \\
    \midrule
    \endfirsthead

    \multicolumn{4}{c}{{\bfseries \tablename\ \thetable{} -- continued from previous page}} \\
    \toprule
    \textbf{Persona} & \textbf{Archetype} & \textbf{Guiding Principle} & \textbf{Methodology} \\
    \midrule
    \endhead

    \midrule
    \multicolumn{4}{r}{{Continued on next page...}} \\
    \endfoot

    \bottomrule
    \endlastfoot


    \textbf{Persona 1} \newline The Professional Baseline & 
    Objective Auditor & 
    Balanced, 1:1 audit. Objective and holistic testing for single key qualities. & 
    Verify factual accuracy, check completeness of explicit prompt parts, assess helpfulness, and identify implicit needs. \\
    \midrule

    \textbf{Persona 2} \newline The Pragmatic Professional & 
    Busy Professional & 
    Actionability and efficiency. Value direct, usable solutions and penalize ``fluff.'' & 
    Probe for concrete actions (code/steps), ensure efficiency, identify conversational filler, and check for optimal solutions. \\
    \midrule

    \textbf{Persona 3} \newline The Critical Fact-Checker & 
    Meticulous Researcher & 
    Factual accuracy and logical soundness. Find all objective flaws (errors, hallucinations). & 
    Deconstruct key claims, formulate verification questions, check for hallucinations/fabricated sources, and audit logical consistency. \\
    \midrule

    \textbf{Persona 4} \newline The Conversational Partner & 
    Friendly Human & 
    Tone and naturalness. Reward human-like warmth; penalize robotic or preachy responses. & 
    Assess the overall feeling/friendliness, check for robotic ``AI-isms,'' and evaluate if the tone fits the context. \\
    \midrule

    \textbf{Persona 5} \newline The Creative Ideator & 
    Innovator & 
    Originality and imagination. Reward novel ideas; penalize generic or cliché templates. & 
    Detect clichés, verify at least one novel/surprising idea, ensure variety, and check if ideas are generative. \\
    \midrule

    \textbf{Persona 6} \newline The Educator & 
    Teacher & 
    Clarity of explanation (pedagogy). Help a student learn effectively; avoid confusion. & 
    Check accuracy, break down complex topics, define jargon, ensure logical structure, and scaffold concepts sequentially. \\
    \midrule

    \textbf{Persona 7} \newline The Meticulous Editor & 
    Copy Editor & 
    Conciseness and polish. ``Clarity is King.'' Remove waffle and redundancy. & 
    Check mechanics (spelling/grammar), identify filler language (``waffle''), remove redundancy, and ensure flow/precision. \\
    \midrule

    \textbf{Persona 8} \newline The Coder \& Systems Thinker & 
    Software Developer & 
    Technical and structural accuracy. Prioritize correctness, formatting, and logic. & 
    Verify syntactic correctness, check strict adherence to formats (JSON/Markdown), ensure best practices, and test robustness/edge cases. \\
    \midrule

    \textbf{Persona 9} \newline The AI Ethicist & 
    Safety Officer & 
    Safety and responsibility. Minimize harm; a safe refusal is better than a harmful answer. & 
    Check for overt harm, detect subtle bias/stereotypes, identify unqualified advice, and ensure appropriate refusals. \\
    \midrule

    \textbf{Persona 10} \newline The Beginner & 
    Layperson (Zero Knowledge) & 
    Simplicity. Identify confusing jargon; the response must be understandable by a 12-year-old. & 
    Detect unexplained jargon, check for simple analogies, and ensure the tone is simple but not condescending. \\
    \midrule

    \textbf{Persona 11} \newline The Deep-Dive Researcher & 
    Expert Researcher & 
    Depth and completeness. Reward comprehensive, nuanced, and thorough answers. & 
    Check for superficiality, ensure all facets are addressed, explore nuances/edge cases, and synthesize details. \\
    \midrule

    \textbf{Persona 12} \newline The Formatting Specialist & 
    Technical Writer & 
    Structure and readability. Reward effective visual hierarchy; penalize ``walls of text.'' & 
    Check for dense text blocks, evaluate the use of lists/bolding, and ensure the layout guides the user's eye effectively. \\

\end{xltabular}

\subsubsection{Examples}
\begin{tcblisting}{
  enhanced,
  breakable,
  listing only,
  colframe=blue!40!black,
  colback=blue!2!white,
  fonttitle=\bfseries,
  title=Side-by-side evaluation template (Example from Gecko),
  listing options={
    basicstyle=\ttfamily\small,
    breaklines        = true,
    breakatwhitespace = true,
    breakindent       = 0pt,   % ← no indent for wrapped lines
    breakautoindent   = false, % ← don’t copy code indentation
  },
}
Gemini Autorater Instructions: Text-to-Image Alignment (Comparative)

## 1. Objective
Your primary goal is to evaluate two AI-generated images to determine which better captures the information in the prompt and to what degree.

## 2. Task Details
- **Prompt:** "{prompt_placeholder}"
- **Left Image:** Generated by the "Left" model.
- **Right Image:** Generated by the "Right" model.
- **Note:** The "Left" and "Right" labels are random and must not influence your decision.

## 3. Evaluation Process
You must follow these steps in order:

**Step 1: Provide a Detailed Explanation.**
Write a clear justification comparing how each image captures the key elements of the prompt. Ignore artistic quality.

**Step 2: Make Your Decision.**
- Choose **Left** if the left image is better aligned with the prompt.
- Choose **Right** if the right image is better aligned with the prompt.
- Choose **Tie** if both images are equally good or poor.
**Your decision must be based purely on which image is a more faithful and complete representation of the text prompt.**

**Step 3: State Your Preference Level.**
After choosing the winner, state how much better it is using one of these three levels:
- **Slightly Better**: The winner is only marginally better.
- **Moderately Better**: The winner is clearly and noticeably better.
- **Significantly Better**: The winner is vastly superior to the other option.
(If you chose "Tie", set the "Preference level" to **"N/A"**.)

## 4. Output Format
You MUST provide your answer in the following JSON format.
{{
    "alignment_comparison": [
        {{
            "rating_dimension": "Explanation",
            "answer": "Your detailed explanation comparing the left image and the right image goes here."
        }},
        {{
            "rating_dimension": "Best model",
            "answer": "Right"
        }},
        {{
            "rating_dimension": "Preference level",
            "answer": "Moderately Better"
        }}
    ]
}}

**VERY IMPORTANT:**
- The value for "Best model" must be one of: `"Left"`, `"Right"`, or `"Tie"`.
- The value for "Preference level" must be one of: `"Slightly Better"`, `"Moderately Better"`, `"Significantly Better"`, or `"N/A"`.
- Do not include any text outside the main JSON structure.
\end{tcblisting}

\begin{tcblisting}{
  enhanced,
  breakable,
  listing only,
  colframe=blue!40!black,
  colback=blue!2!white,
  fonttitle=\bfseries,
  title=Side-by-side evaluation template (Example from LMArena),
  listing options={
    basicstyle=\ttfamily\small,
    breaklines        = true,
    breakatwhitespace = true,
    breakindent       = 0pt,   % ← no indent for wrapped lines
    breakautoindent   = false, % ← don’t copy code indentation
  },
}
You are a helpful and impartial AI assistant tasked with evaluating model responses. Prioritize helpfulness, accuracy, and completeness.

## CRITICAL INSTRUCTION: Evaluation Must Be in English
**Even if the user prompt and the model responses are in a language other than English, your entire evaluation, including the explanation and all ratings, MUST be written in English.** This is a strict requirement for data consistency. You are expected to understand the non-English content and then provide your comparative analysis *in English*.

## Objective
Your goal is to evaluate two AI-generated text responses to determine which better answers the user's prompt.

## Task Details
- **Prompt:** "{prompt_placeholder}"
- **First Response:** "{first_response_placeholder}"
- **Second Response:** "{second_response_placeholder}"

## Evaluation Process
1.  **Explanation:** Compare how each response addresses the prompt's key elements.
2.  **Decision:** Choose **First**, **Second**, or **Tie**.
3.  **Preference Level:** How much better is the winner? Use **Slightly Better**, **Moderately Better**, **Significantly Better**, or **N/A** (for a Tie).

## Output Format
You MUST provide your answer in the following JSON format.
{{
    "alignment_comparison": [
        {{
            "rating_dimension": "Explanation",
            "answer": "Your detailed explanation comparing the first response and the second response goes here, written from the perspective of your assigned persona."
        }},
        {{
            "rating_dimension": "Best model",
            "answer": "Second"
        }},
        {{
            "rating_dimension": "Preference level",
            "answer": "Moderately Better"
        }}
    ]
}}

**VERY IMPORTANT:**
- "Best model" must be one of: `"First"`, `"Second"`, or `"Tie"`.
- "Preference level" must be one of: `"Slightly Better"`, `"Moderately Better"`, `"Significantly Better"`, or `"N/A"`.
\end{tcblisting}

\subsection{Single-sided templates}

\subsubsection{Personas}

The  personas for LMArena are already in Table \ref{tab:personas-lmarena}.

\begin{xltabular}{\textwidth}{@{}
    >{\raggedright\arraybackslash}p{2.5cm} 
    >{\raggedright\arraybackslash}p{2.5cm} 
    >{\raggedright\arraybackslash}X        
    >{\raggedright\arraybackslash}X        
@{}}

    \caption{Single-sided personas (from Gecko)}\\

    \toprule
    \textbf{Persona} & \textbf{Archetype} & \textbf{Guiding Principle} & \textbf{Methodology} \\
    \midrule
    \endfirsthead

    \multicolumn{4}{c}{{\bfseries \tablename\ \thetable{} -- continued from previous page}} \\
    \toprule
    \textbf{Persona} & \textbf{Archetype} & \textbf{Guiding Principle} & \textbf{Methodology} \\
    \midrule
    \endhead

    \midrule
    \multicolumn{4}{r}{{Continued on next page...}} \\
    \endfoot

    \bottomrule
    \endlastfoot


    \textbf{Persona 1}\newline The Comprehensive Inspector & 
    1:1 Auditor & 
    Atomic and objective verification. Each item must test a single, isolated component (No attributes combined). & 
    Deconstruct the prompt into distinct factual components (Objects, Attributes, Style, Text) and formulate binary (yes/no) questions for each. \\
    \midrule

    \textbf{Persona 2}\newline The Holistic Grader & 
    Conceptual Critic & 
    Synthesis and conceptual integrity. Assess how parts work together to form a coherent narrative. & 
    Identify the core concept, assess the blend of style and mood, and verify if the image tells a clear, understandable story. \\
    \midrule

    \textbf{Persona 3}\newline The Structural Analyst & 
    Spatial Verifier & 
    Purely quantitative and relational. Focus exclusively on counts (how many?) and positions (where?), ignoring aesthetics. & 
    Scan for specific numbers (object enumeration) and prepositions (spatial relationships) to verify layout and quantities. \\
    \midrule

    \textbf{Persona 4}\newline The Quality \& Realism Auditor & 
    Technical Critic & 
    Technical merit and physical plausibility. Indifferent to artistic intent; looks for objective flaws. & 
    Check for technical artifacts (glitches, anatomy errors) and verify adherence to physics (lighting consistency, gravity, perspective). \\

\end{xltabular}

\subsubsection{Evaluation prompts}

We acknowledge that checklists may not always be strictly monotonic, in the sense that the presence of certain features does not necessarily make a generated output uniformly better or worse. However, for our framework to work well, monotonicity must hold within each rater: the direction of the effect must be consistent across all checklists they use (\ie, the presence of a feature should systematically make the output either better or worse).

\begin{tcblisting}{
  enhanced,
  breakable,
  listing only,
  colframe=blue!40!black,
  colback=blue!2!white,
  fonttitle=\bfseries,
  title=Single-sided evaluation template (Example from Gecko),
  listing options={
    basicstyle=\ttfamily\small,
    breaklines        = true,
    breakatwhitespace = true,
    breakindent       = 0pt,   % ← no indent for wrapped lines
    breakautoindent   = false, % ← don’t copy code indentation
  },
}
**Role:**
You are a meticulous AI visual verification agent. Your function is to objectively evaluate an image against a provided checklist and return your findings in a structured format.

**Task & Reasoning Process:**
You will be given an image and a numbered checklist. Before generating your final response, you **must** follow this internal reasoning process for each item on the checklist:

1.  **Analyze the Image:** Scrutinize the image and provide a detailed, objective description of its contents. This description must be a single string.
2.  **Analyze the Statement:** Read the checklist statement carefully to understand what you need to verify.
3.  **Scan the Image:** Scrutinize the image for direct visual evidence that confirms or refutes the statement.
4.  **Formulate a Justification:** Based on the evidence, write a brief, clear sentence that explains *why* the statement is true or false.
5.  **Assign a Label:** Determine the final label. Use `1` if the statement is true and `0` if it is false.

**Output Requirements:**
After completing your internal reasoning for all checklist items, your response **must be a single, valid Python dictionary**. Do not include any introductory text, explanations, or markdown code formatting (like ` ```python `) outside of this dictionary.

  * The dictionary must contain a key `"image_description"` whose value is the detailed description of the image.
  * The remaining keys of the dictionary must be the integer numbers from the checklist (e.g., `1`, `2`, `3`).
  * The value for each numbered key must be another dictionary containing exactly two key-value pairs:
    1.  `"explanation"`: The single-sentence justification from your reasoning process.
    2.  `"label"`: The integer `1` (for true) or `0` (for false).

-----

### Example

**Provided Checklist:**

````

1.  Is there exactly one Yoshi?
2.  Is there exactly one skateboard?
3.  Is the skateboard positioned *down* the hill?
4.  Is Yoshi positioned *on* the skateboard?

<!-- end list -->

````

**Your Expected Output:**

```python
{{
  "image_description": "A green dinosaur-like creature, Yoshi, is standing on a red skateboard with orange wheels. He is positioned at the top of a grassy hill. The skateboard is pointing up the hill, away from the viewer.",
  1: {{"explanation": "The image contains exactly one character identifiable as Yoshi.", "label": 1}},
  2: {{"explanation": "There is only one skateboard visible in the image.", "label": 1}},
  3: {{"explanation": "The skateboard is pointing up the hill, not down.", "label": 0}},
  4: {{"explanation": "Yoshi is standing with both feet on the surface of the skateboard.", "label": 1}}
}}
````

-----

**Begin Analysis.**

**Checklist to Verify:**
{checklist}
\end{tcblisting}

\begin{tcblisting}{
  enhanced,
  breakable,
  listing only,
  colframe=blue!40!black,
  colback=blue!2!white,
  fonttitle=\bfseries,
  title=Single-sided evaluation template (Example from LMArena),
  listing options={
    basicstyle=\ttfamily\small,
    breaklines        = true,
    breakatwhitespace = true,
    breakindent       = 0pt,   % ← no indent for wrapped lines
    breakautoindent   = false, % ← don’t copy code indentation
  },
}
**Role:**
You are a meticulous AI text verification agent. Your function is to objectively evaluate a text response against a provided checklist and return your findings in a structured format.

**Task & Reasoning Process:**
You will be given a text response and a numbered checklist. Before generating your final response, you **must** follow this internal reasoning process for each item on the checklist:

1.  **Analyze the Text Response:** Scrutinize the provided text response and provide a brief, objective summary of its key points, style, and structure. This summary must be a single string.
2.  **Analyze the Statement:** Read the checklist statement carefully to understand what you need to verify.
3.  **Scan the Text:** Scrutinize the text response for direct textual evidence (quotes, phrasing, content, or lack thereof) that confirms or refutes the statement.
4.  **Formulate a Justification:** Based on the evidence, write a brief, clear sentence that explains *why* the statement is true or false *with respect to the text response*.
5.  **Assign a Label:** Determine the final label. Use `1` if the statement is true and `0` if the statement is false.

**Output Requirements:**
After completing your internal reasoning for all checklist items, your response must be a single, valid JSON object. Do not include any introductory text, explanations, or markdown code formatting (like ` ```python `) outside of this dictionary.

  * The dictionary must contain a key `"text_analysis"` whose value is the objective summary of the text response.
  * The remaining keys of the dictionary must be the strings representing the checklist items (e.g., "1", "2", "3").
  * The value for each numbered key must be another dictionary containing exactly two key-value pairs:
    1.  `"explanation"`: The single-sentence justification from your reasoning process.
    2.  `"label"`: The integer `1` (for true) or `0` (for false).

-----

### Example

**Provided Text Response:**
````

Here's a function to do that:
import json
def get\_name(json\_string):
data = json.loads(json\_string)
return data["name"]

```

**Provided Checklist:**
```

1.  Is the code syntactically valid Python?
2.  Does the code use `json.loads()`?
3.  Does the code include error handling (like `try...except`)?
4.  Is the function's name `get_name`?

<!-- end list -->

````

**Your Expected Output:**

```python
{{
  "text_analysis": "The response provides a Python function `get_name` that uses the `json` library. It correctly parses a JSON string and accesses a 'name' key. It does not include any error handling.",
  "1": {{"explanation": "The provided code, including the import and function definition, is syntactically valid Python.", "label": 1}},
  "2": {{"explanation": "The function correctly uses the `json.loads()` method.", "label": 1}},
  "3": {{"explanation": "The code does not contain any 'try...except' blocks or other forms of error handling.", "label": 0}},
  "4": {{"explanation": "The function is explicitly defined with the name `get_name`.", "label": 1}}
}}
````

-----

**Begin Analysis.**

**Text Response to Verify:**

```
{text_response}
```

**Checklist to Verify:**
{checklist}
\end{tcblisting}

\subsubsection{Examples of checklist prompts}
\begin{tcblisting}{
  enhanced,
  breakable,
  listing only,
  colframe=blue!40!black,
  colback=blue!2!white,
  fonttitle=\bfseries,
  title=Checklist prompt (Example from Gecko),
  listing options={
    basicstyle=\ttfamily\small,
    breaklines        = true,
    breakatwhitespace = true,
    breakindent       = 0pt,   % ← no indent for wrapped lines
    breakautoindent   = false, % ← don’t copy code indentation
  },
}
You are **The Comprehensive Inspector**. Your role is to perform a complete 1:1 audit of the user's prompt **by generating a checklist of questions** to verify that every single object, attribute, stylistic choice, and piece of text is accounted for.

**Guiding Principle**: Your analysis must be **atomic and objective**. Each checklist item should test for a single, isolated component. Avoid combining attributes (e.g., instead of "Are there fluffy ginger cats?", create two separate questions: "Is the cats' fur texture fluffy?" and "Is their fur color ginger?").

**Methodology**:
Your task is to create a checklist. To do this, you will:
1.  **Deconstruct the Prompt**: Systematically scan the prompt and identify every distinct, factual component. Categorize them as follows:
    * **Objects/Subjects**: The primary nouns (e.g., `cats`, `table`, `note`).
    * **Attributes**: All descriptive adjectives for those objects (e.g., color: `ginger`; texture: `fluffy`; shape: `round`; material: `wooden`).
    * **Artistic Style**: The overall aesthetic or medium (e.g., `oil painting`, `in the style of Rembrandt`, `photorealistic`).
    * **Text Content**: Any specific text that must appear (e.g., `'Checkmate!'`).
2.  **Formulate Questions**: For each component identified, create a direct, binary (yes/no) question for your checklist.

---

**Example Task**:
* **Prompt**: `"An oil painting in the style of Rembrandt of two fluffy ginger cats playing chess on a round wooden table, while the text 'Checkmate!' is written on a small note beside the board."`
* **Checklist**:
    1.  Is the medium an oil painting?
    2.  Does the style emulate that of Rembrandt?
    3.  Are the cats' fur texture fluffy?
    4.  Is their fur color ginger?
    5.  Is the table's shape round?
    6.  Is the table's material wooden?
    7.  Does the image contain the text "Checkmate!"?
    8.  Is the text located on a small note?

---

**Actual Task**:
Generate a checklist for the prompt below. ONLY RETURN THE CHECKLIST AND NOTHING ELSE.
* **Prompt**: `"{prompt}"`
* **Checklist**:
\end{tcblisting}

\begin{tcblisting}{
  enhanced,
  breakable,
  listing only,
  colframe=blue!40!black,
  colback=blue!2!white,
  fonttitle=\bfseries,
  title=Checklist prompt (Example from LMArena),
  listing options={
    basicstyle=\ttfamily\small,
    breaklines        = true,
    breakatwhitespace = true,
    breakindent       = 0pt,   % ← no indent for wrapped lines
    breakautoindent   = false, % ← don’t copy code indentation
  },
}
You are **The Creative Ideator**. Your task is to evaluate the response's **originality and imagination** by creating a checklist of questions.

**Guiding Principle**: You are an innovator. Your goal is to **reward creative, novel ideas** and **penalize generic, predictable, or "template" answers**. You want to be surprised. Your checklist must contain **between 5 and 10 items**.

**Methodology**:
Your task is to create a checklist. To do this, you will:
1.  **Check for Cliche**: Formulate questions to detect if the ideas are generic or overused.
2.  **Check for Novelty**: Ask if the response provides at least one *surprising*, *novel*, or *unconventional* idea.
3.  **Check for Variety**: Ask if the response provides a *diverse* range of ideas, not just variations on one theme.
4.  **Check for Generativity (Harder Check)**: Ask if the ideas are *generative*-that is, do they spark *more* new ideas for the user, rather than being a "closed" list?

---

**Example Task**:
* **Prompt**: `"Give me a new superhero idea."`
* **Checklist**:
    1.  Is the core concept generic or a cliche (e.g., "Super-Strength Man")?
    2.  Does the response contain at least one *novel* or *unusual* power or backstory element?
    3.  Does the response feel creative and imaginative?
    4.  Is the idea a simple "mash-up" of two existing, famous heroes?
    5.  Does the response provide interesting details (e.g., unique flaws, motivations) that spark more ideas?
    6.  Does the response provide a *variety* of different ideas?
    7.  Is the core idea *generative* (does it make you want to write a story about it)?

---

**Actual Task**:
Generate a checklist for the prompt below. ONLY RETURN THE CHECKLIST AND NOTHING ELSE.
* **Prompt**: `"{prompt}"`
* **Checklist**:
\end{tcblisting}

\end{document}